%% file: funnel_library_ijrr.tex
\title{Funnel Libraries for Real-Time Robust Feedback Motion Planning}

\author{
        Anirudha Majumdar and Russ Tedrake \\
        Computer Science and Artificial Intelligence Lab \\
        Massachusetts Institute of Technology \\
        Cambridge, MA, USA\\
        Email: \{anirudha,russt\}@mit.edu \\
}

\date{\today}

\documentclass[11pt]{article}
\usepackage{fullpage}
\usepackage{breakcites}

\usepackage{graphicx}
\usepackage{amsmath}
\usepackage{amssymb}
\usepackage[pdfborder={0 0 0},colorlinks=false,citecolor=blue,urlcolor=blue]{hyperref}
\usepackage{pdfpages}
\usepackage[export]{adjustbox}
\usepackage[titletoc]{appendix}

\usepackage{mathtools}
\usepackage{rotating}

\usepackage{algorithm2e}

\usepackage{enumitem}
\usepackage{collect}

\usepackage{longtable}
\usepackage{boldline}

\usepackage{array}
\newcolumntype{L}[1]{>{\raggedright\let\newline\\\arraybackslash\hspace{0pt}}m{#1}}
\newcolumntype{C}[1]{>{\centering\let\newline\\\arraybackslash\hspace{0pt}}m{#1}}
\newcolumntype{R}[1]{>{\raggedleft\let\newline\\\arraybackslash\hspace{0pt}}m{#1}}

\newcommand{\RR}{\mathbb{R}}

\usepackage{stackengine}
\usepackage{scalefnt}
\usepackage{float}
\usepackage{makeidx}         
\usepackage[bottom,hang,flushmargin]{footmisc}
\usepackage{amsmath}
\usepackage{amsfonts}
\usepackage{amsthm}
\usepackage[hang,tight]{subfigure}
\usepackage[font=footnotesize]{caption}
\usepackage{algorithmic}
\usepackage{color}
\usepackage{framed}
\usepackage{mathtools}

\newtheoremstyle{nospace}
{2pt}   				
{2pt}   				
{\itshape}  			
{} 		  		    	
{\bfseries} 			
{.}         			
{5pt plus 1pt minus 1pt}
{}          			

\theoremstyle{nospace} 
\theoremstyle{nospace} 
\theoremstyle{nospace} 
\theoremstyle{nospace} \newtheorem{remark}{Remark}
\theoremstyle{nospace} 
\theoremstyle{nospace} \newtheorem{definition}{Definition}
\theoremstyle{nospace} 
\theoremstyle{nospace}

\newcommand{\GF}{\mathcal{G}(\mathcal{F})}
\newcommand{\F}{\mathcal{F}}

\newcommand{\T}{\mathcal{T}}

\newcommand{\R}{\mathbb{R}}

\newcommand{\OO}{\mathcal{O}}

\newcommand{\revision}[1]{\normalsize{{#1}}}

\begin{document}
\maketitle

\begin{abstract}
We consider the problem of generating motion plans for a robot that are guaranteed to succeed despite uncertainty in the environment, parametric model uncertainty, and disturbances. Furthermore, we consider scenarios where these plans must be generated in real-time, because constraints such as obstacles in the environment may not be known until they are perceived (with a noisy sensor) at runtime. Our approach is to pre-compute a library of ``funnels" along different maneuvers of the system that the state is guaranteed to remain within (despite bounded disturbances) when the feedback controller corresponding to the maneuver is executed. We leverage powerful computational machinery from convex optimization (\emph{sums-of-squares programming} in particular) to compute these funnels. The resulting \emph{funnel library} is then used to sequentially compose motion plans at runtime while ensuring the safety of the robot. A major advantage of the work presented here is that by explicitly taking into account the effect of uncertainty, the robot can evaluate motion plans based on how vulnerable they are to disturbances. 

We demonstrate and validate our method using extensive hardware experiments on a small fixed-wing airplane avoiding obstacles at high speed ($\sim$12 mph), along with thorough simulation experiments of ground vehicle and quadrotor models navigating through cluttered environments. To our knowledge, these demonstrations constitute one of the first examples of provably safe and robust control for robotic systems with complex nonlinear dynamics that need to plan in real-time in environments with complex geometric constraints. 

\end{abstract}

\input{introduction}

\input{related}

\input{background}

\input{funnels}

\input{funnel_libraries}

\input{planning}

\input{examples}

\input{hardware}
\input{conclusion}

\input{acknowledgements}


\bibliographystyle{apalike}

\bibliography{elib}

\end{document}

%% file: introduction.tex
\section{Introduction}
\label{sec:intro}

Imagine an unmanned aerial vehicle (UAV) flying at high speed through a cluttered environment in the presence of wind gusts, a legged robot traversing rough terrain, or a mobile robot grasping and manipulating previously unlocalized objects in the environment. These applications demand that the robot move
through (and in certain cases interact with) its environment with a very high degree of agility while still being in close proximity to obstacles. Such systems today lack guarantees on their safety and can fail dramatically in
the face of uncertainty in their environment and dynamics. 

The tasks mentioned above are characterized by three main challenges. First, the dynamics of the system are nonlinear, underactuated, and subject to constraints on the input (e.g. torque limits). Second, there is a significant amount of uncertainty in the dynamics of the system due to disturbances and modeling error. Finally, the geometry of the environment that the robot is operating in is unknown until runtime, thus forcing the robot to plan in \emph{real-time}.

\subsection{Contributions}

\begin{figure*}[t!]\label{fig:robustness}
\centering
  \subfigure[A plane deviating from its nominal planned trajectory due to a heavy cross-wind.]{
  \includegraphics[width=.47\textwidth]{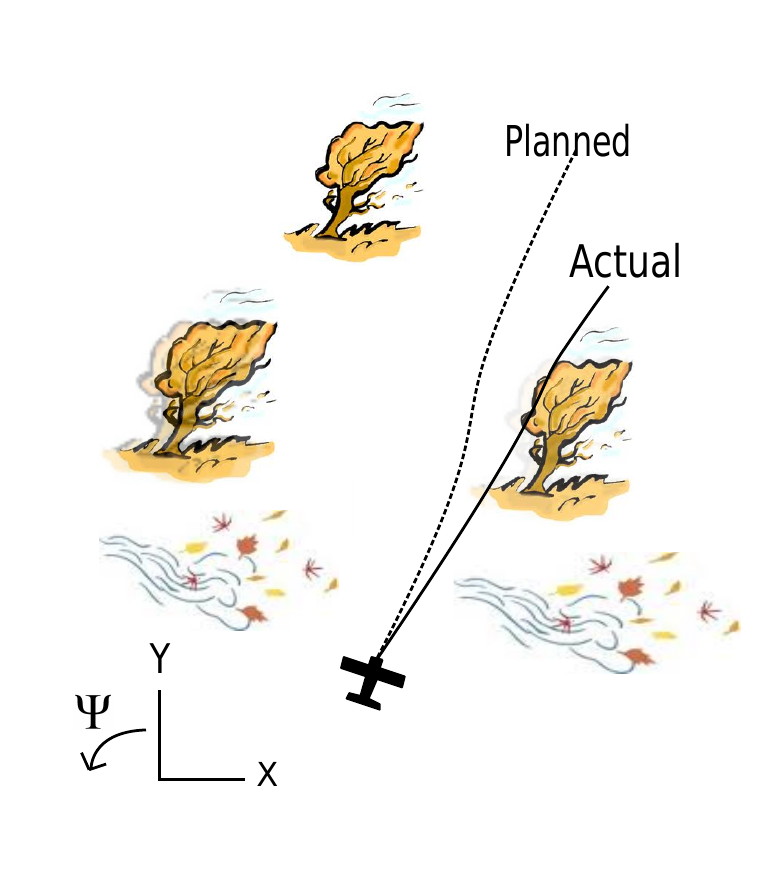} 
  \label{fig:robustness_a}
  }
 \subfigure[The ``funnel" of possible trajectories.]{
  \includegraphics[width=.47\textwidth]{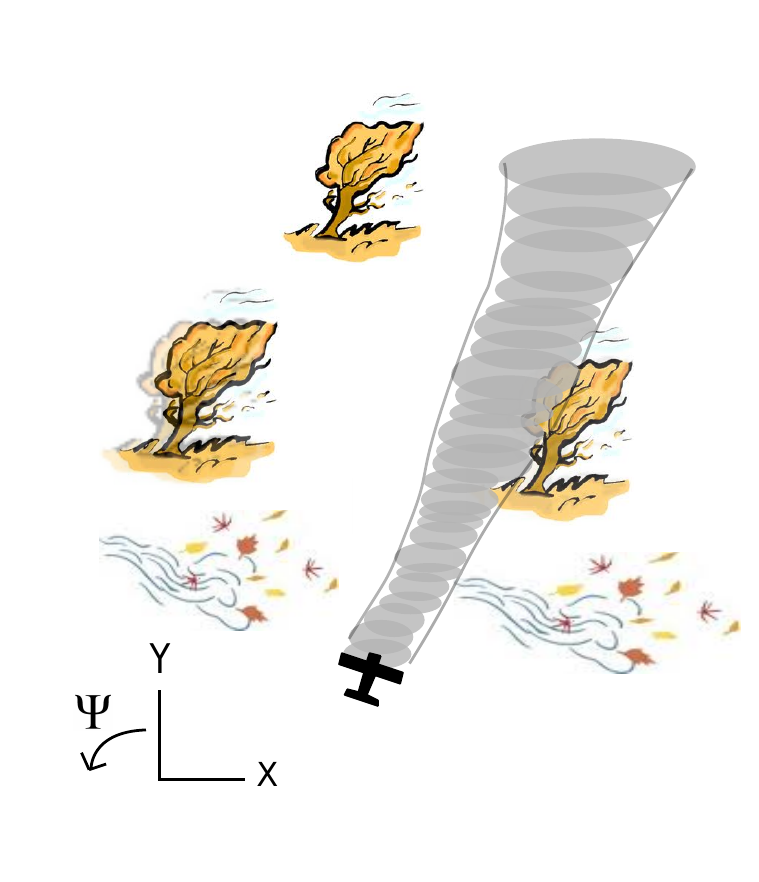} 
 \label{fig:robustness_b}
  }
\caption[The need to account for uncertainty during planning]{Not accounting for uncertainty while planning motions can lead to disastrous consequences.}
\end{figure*}

In this paper, we address these challenges by combining approaches from motion planning, feedback control, and tools from Lyapunov theory and convex optimization in order to perform robust real-time motion planning in the face of uncertainty. In particular, in an offline computation stage, we first design a finite library of open loop trajectories. For each trajectory in this library, we use tools from convex optimization (\emph{sums-of-squares (SOS) programming} in particular) to design a controller that explicitly attempts to minimize the size of the worst case reachable set of the system given a description of the uncertainty in the dynamics and bounded external disturbances. This control design procedure yields an outer approximation of the reachable set, which can be visualized as a ``funnel'' around the trajectory, that the closed-loop system is guaranteed to remain within. A cartoon of such a funnel is shown in Figure \ref{fig:robustness_b}. Finally, we provide a way of \emph{sequentially composing} these robust motion plans at runtime in order to operate in a provably safe manner in previously unseen environments.

One of the most important advantages that our approach affords us is the ability to choose between the motion primitives in our library in a way that takes into account the dynamic effects of uncertainty. Imagine a UAV flying through a forest that has to choose between two motion primitives: a highly dynamic roll maneuver that avoids the trees in front of the UAV by a large margin or a maneuver that involves flying straight while avoiding the trees only by a small distance. An approach that neglects the effects of disturbances and uncertainty may prefer the former maneuver since it avoids the trees by a large margin and is therefore ``safer". However, a more careful consideration of the two maneuvers could lead to a different conclusion: the dynamic roll maneuver is far more susceptible to wind gusts and perturbations to the initial state than the second one. Thus, it may in fact be more robust to execute the second motion primitive. Further, it may be possible that neither maneuver is guaranteed to succeed and it is safer to abort the mission and simply transition to a hover mode. Our approach allows robots to make these critical decisions, which are essential if robots are to move out of labs and operate in real-world environments.

We demonstrate and validate our approach using thorough simulation experiments of ground vehicle and quadrotor models navigating through cluttered environments, along with extensive hardware experiments on a small fixed-wing airplane avoiding obstacles at high speed ($\sim$12 miles per hour) in an indoor motion capture environment. To the best of our knowledge, the resulting demonstrations constitute one of the first examples of provably safe and robust control for robotic systems with complex nonlinear dynamics that need to plan in real-time in cluttered environments.

\subsection{Outline}

The outline of the paper is as follows. Section \ref{sec:related} discusses prior work; Section \ref{sec:background} provides a brief background on semidefinite and sums-of-squares (SOS) programming, which are used heavily throughout the paper; Section \ref{sec:funnels} shows how to use SOS programming to compute funnels; Section \ref{sec:funnel libraries} introduces the notion of a funnel library; Section \ref{sec:planning} describes our algorithm for using funnel libraries for real-time robust planning in environments that have not been seen by the robot before; Section \ref{sec:examples} presents extensive simulation results on a ground vehicle model and compares our approach with an approach based on trajectory libraries; Section \ref{sec:examples} also considers a quadrotor model and shows how one can use our approach to guarantee collision-free flight in certain environments;  Section \ref{sec:hardware} presents hardware experiments on a small fixed-wing airplane in order to demonstrate and validate our approach; Section \ref{sec:conclusion} concludes the paper with a discussion of challenges and open problems.

%% file: related.tex
\section{Relevant Work}
\label{sec:related}

\subsection{Motion Planning}

Motion planning has been the subject of significant research in the last few decades and has enjoyed a large degree of success in recent years. Planning algorithms like the Rapidly-exploring Randomized Tree (RRT) \cite{Kuffner00} (along with variants that attempt to find optimal motion plans \cite{Karaman11} \cite{Kobilarov12}) and related trajectory library approaches \cite{Liu09a} \cite{Frazzoli05} \cite{Stolle06} can handle large state-space dimensions and complex differential constraints. These algorithms have been successfully demonstrated on a wide variety of hardware platforms  \cite{Kuwata08} \cite{Shkolnik10b} \cite{Sermanet08} \cite{Satzinger16}. However, an important challenge is their inability to explicitly reason about  uncertainty and feedback. Modeling errors, state uncertainty and disturbances can lead to failure if the system deviates from the planned nominal trajectories.

The motion planning aspect of our approach draws inspiration from the vast body of work that is focused on computing motion primitives in the form of trajectory libraries. For example, trajectory libraries have been used in diverse applications such as humanoid balance control \cite{Liu09a}, autonomous ground vehicle navigation \cite{Sermanet08}, grasping \cite{Berenson07} \cite{Dey11}, and UAV navigation \cite{Dey14} \cite{Barry16}. The Maneuver Automaton \cite{Frazzoli05} attempts to capture the formal properties of trajectory libraries as a hybrid automaton, thus providing a unifying theoretical framework. Maneuver Automata have also been used for real-time motion planning with static and dynamic obstacles \cite{Frazzoli02}. Further theoretical investigations have focused on the offline generation of diverse but sparse trajectories that ensure the robot's ability to perform the necessary tasks online in an efficient manner \cite{Green07}. More recently, tools from sub-modular sequence optimization have been leveraged in the optimization of the sequence and content of trajectories evaluated online \cite{Dey11, Dey15a}. Prior work has also been aimed at \emph{learning} maneuver libraries in the form of ``skill trees" from expert demonstrations \cite{konidaris2011robot}.

Robust motion planning has also been a very active area of research in the robotics community. Early work \cite{Brooks82} \cite{Lozano-Perez84} \cite{Jacobs90} \cite{Latombe91a} focused on settings where the dynamics of the system are not dominant and one can treat the system as a kinematic one. The problem is then one of planning paths through configuration space that are robust to uncertainty in the motion of the robot and geometry of the workspace. Our work with funnels takes inspiration from the early work on fine-motion planning, where the notions of \emph{funnels} \cite{Mason85} and \emph{preimage backchaining} \cite{Lozano-Perez84} (also known as \emph{goal regression} or \emph{sequential composition}) were first introduced. The theme of robust kinematic motion planning has persisted in recent work \cite{Missiuro06} \cite{Guibas09} \cite{Malone13}, which deals with uncertainty in the geometry of the environment and obstacles.

\subsection{Planning under Uncertainty}

In settings where the dynamics of the system must be taken into account (e.g., for underactuated systems), the work on \emph{planning under uncertainty} attempts to address the challenges associated with uncertainty in the dynamics, state, and geometry of the environment. In particular, chance-constrained programming \cite{Charnes59} provides an elegant mathematical framework for reasoning about stochastic uncertainty in the dynamics, initial conditions, and obstacle geometry \cite{Blackmore06} \cite{Oldewurtel08} \cite{DuToit11} and allows one to generate motion plans with bounds on the probability of collision with obstacles. In settings where the state of the system is partially observable and there is significant uncertainty in the observations, one can extend this framework to plan in the \emph{belief space} of the system (i.e., the space of distributions over states) \cite{Vitus11} \cite{Vitus12}. While these approaches allow one to explicitly reason about uncertainty in the system, they are typically restricted to handling linear dynamical systems with Gaussian uncertainty. This is due to the computational burden of solving chance constrained problems for nonlinear and non-Gaussian systems.

Similarly, other approaches for belief space planning \cite{Bry11} \cite{Platt10} \cite{Prentice07} \cite{VanDenBerg12} \cite{Agha-Mohammadi14} \cite{Patil15} also approximate the belief state as a Gaussian distribution over state space (however, see \cite{Platt12a} for an exception to this) for computational efficiency and hence the true belief state is not tracked. Thus, in general, one does not have robustness guarantees. The approach we take in this work is to assume that disturbances/uncertainty are \emph{bounded} and provide explicit bounds on the reachable set to facilitate safe operation of the \emph{nonlinear} system.

More generally, the rich literature on Partially Observable Markov Decision Processes (POMDPs) \cite{Kaelbling98} is also relevant here. POMDPs present an elegant mathematical framework for reasoning about uncertainty in state and dynamics. However, we note that our focus in this work is on dynamical systems with continuous state and action spaces, whereas the POMDP literature typically focuses on discretized state/action spaces for the most part.    

\subsection{Reachable Sets}

Reachability analysis for linear and nonlinear control systems has a long history in the controls community. For \emph{linear} systems subject to bounded disturbances, there exist a number of techniques for efficiently computing (approximations of) backwards and forwards reachable sets \cite{Kurzhanski00a} \cite{Girard05} \cite{Yazarel04}. One can apply techniques from linear reachability analysis to conservatively approximate reachable sets of nonlinear systems by treating nonlinearities as bounded disturbances \cite{Althoff08}. This idea has been used in \cite{Althoff14} for performing online safety verification for ground vehicles. A similar idea was used in \cite{Althoff15} to perform online safety verification for UAVs and to decide when the UAV should switch to an emergency maneuver (a ``loiter circle"). In this paper we will also compute outer approximations of reachable sets (which we refer to as ``funnels"). However, the approach we present here is not based on a linearization of the system and thus has the potential to be less conservative for highly nonlinear systems. Further, the scope of our work extends beyond verification; the emphasis here is on safe real-time \emph{planning} with funnels. 

Approximations of reachable sets for nonlinear systems can be computed via Hamilton-Jacobi-Bellman (HJB) differential game formulations \cite{Mitchell05a}.
This method was used in \cite{Gillula10} for designing motion primitives for making a quadrotor perform an autonomous backflip in a 2D plane. While this approach handles unsafe sets that the system is not allowed to enter, it is assumed that these sets are specified \emph{a priori}. In this paper, we are concerned with scenarios in which unsafe sets (such as obstacles) are not specified until runtime and must thus be reasoned about \emph{online}. Further, techniques for computing reachable sets based on the HJB equation have historically suffered from the curse of dimensionality since they rely on discretizations of the state space of the system. Hence, unless the system under consideration has special structure (e.g., decoupled systems \cite{Chen15}), these methods have difficulty scaling up beyond approximately 5-6 dimensional state spaces.     

An approach that is closely related to our work is the work presented in \cite{Ny12}. The authors propose a randomized planning algorithm in the spirit of RRTs that explicitly reasons about disturbances and uncertainty. Specifications of input to output stability with respect to disturbances provide a parameterization of ``tubes'' (analogous to our ``funnels'') that can be composed together to generate motion plans that are collision-free. The factors that distinguish the approach we present in this paper from the one proposed in \cite{Ny12} are our focus on the \emph{real-time} aspect of the problem and use of sums-of-squares programming as a way of computing reachable sets. In \cite{Ny12}, the focus is on generating safe motion plans when the obstacle positions are known \emph{a priori}.  Further, we provide a general technique for computing and explicitly minimizing the size of tubes.

Another approach that is closely related to ours is Model Predictive Control with Tubes  \cite{Mayne05}. The idea is to solve the optimal control problem online with guaranteed ``tubes" that the trajectories stay in. A closely related idea is that of ``flow-tubes'', which have been used for high-level planning for autonomous systems \cite{Li08}. However, these methods are generally limited to linear systems.

\subsection{Lyapunov Theory and SOS programming}
\label{sec:lyapunov}

A critical component of the work presented here is the computation of ``funnels'' for nonlinear systems via Lyapunov functions. The metaphor for thinking about Lyapunov functions as defining funnels was introduced to the robotics community in \cite{Burridge99}, where funnels were \emph{sequentially composed} in order to produce dynamic behaviors in a robot. However, computational tools for automatically searching for Lyapunov functions were lacking until very recently. In recent years, sums-of-squares (SOS) programming has emerged as a way of checking the Lyapunov function conditions associated with each funnel \cite{Parrilo00}. The technique relies on the ability to check nonnegativity of multivariate polynomials by expressing them as a sum of squares of polynomials. This can be written as a semidefinite optimization program and is amenable to efficient computational algorithms such as interior point methods \cite{Parrilo00}. Assuming polynomial dynamics, one can check that a polynomial Lyapunov candidate, $V(x)$, satisfies $V(x) > 0$ and $\dot{V}(x) < 0$ in some region $B_r$. Importantly, the same idea can be used for computing funnels along time-indexed trajectories of a system \cite{Tedrake10} \cite{Tobenkin10b}. In this paper, we will use a similar approach to synthesize feedback controllers that explicitly seek to minimize the effect of disturbances on the system by minimizing the size of the funnel computed along a trajectory. Thus, we are guaranteed that if the system starts off in the set of given initial conditions, it will remain in the computed ``funnel" even if the model of the dynamics is uncertain and the system is subjected to bounded disturbances.

The ability to compute funnels using SOS programming was leveraged by the LQR-Trees algorithm \cite{Tedrake10} for \emph{feedback motion planning} for nonlinear systems. The algorithm works by creating a tree of locally stabilizing controllers which can take any initial condition in some bounded region in state space to the desired goal. However, LQR-Trees lack the ability to handle scenarios in which the task and environment are unknown till runtime: the offline precomputation of the tree does not take into account potential runtime constraints like obstacles, and an online implementation of the algorithm is computationally infeasible.

The SOS programming approach has also been used to guarantee obstacle avoidance conditions for nonlinear systems by computing \emph{barrier certificates} \cite{Prajna06} \cite{Barry12}. Barrier functions are similar to Lyapunov functions in spirit, but are used to guarantee that trajectories starting in some given set of initial conditions will never enter an ``unsafe" region containing obstacles. This approach, however, is limited to settings where the locations and geometry of obstacles are known beforehand since the barrier certificate one computes depends on this data and computing barrier certificates in real-time using SOS programming is not computationally feasible at the present time.


%% file: background.tex
\section{Background}
\label{sec:background}

In this section we provide a brief background on the key computational tools that will be employed throughout this paper. 

\subsection{Semidefinite Programming (SDP)}
\label{sec:background_sdp}

Semidefinite programs (SDPs) form an important class of convex optimization problems. They are optimization problems over the space of symmetric positive semidefinite (psd) matrices. Recall that a $n \times n$ symmetric matrix $Q$ is positive semidefinite if $x^TQx \geq 0, \ \forall x \in \R^n$. Denoting the set of $n \times n$ symmetric matrices as $\mathbf{S}^n$, a SDP in standard form is written as:

\begin{flalign} 
	 \label{eq:sdp}
			& & \min_{X \in \mathbf{S}^n} \hspace*{0.5cm} & \langle C, X \rangle && \\
			& & \text{s.t.} \hspace*{0.5cm} & \langle A_i, X \rangle = b_i \quad \forall i \in \{1,\dots,m\}, && \nonumber \\
			& & & X \succeq 0, && \nonumber 
\end{flalign}
where $C, A_i \in \mathbf{S}^n$ and $\langle X, Y \rangle := \textrm{Tr}(X^T Y) = \sum_{i,j} X_{ij} Y_{ij}$. In other words, a SDP involves minimizing a cost function that is linear in the elements of the decision matrix $X$ subject to linear and positive semidefiniteness constraints on $X$. 

Semidefinite programming includes Linear Programming (LP), Quadratic Programming (QP) and Second-Order Cone Programming (SOCP) as special cases. As in these other cases, SDPs are amenable to efficient numerical solution via interior point methods. The interested reader may wish to consult \cite{Vandenberghe96} and ~\cite[Section 2]{Blekherman13} for a more thorough introduction to SDPs.

\subsection{Sums-of-Squares (SOS) Programming}
\label{sec:background_sos}

An important application of SDPs is to check nonnegativity of polynomials. The decision problem associated with checking polynomial nonnegativity is NP--hard in general \cite{Parrilo00}. However, the problem of determining whether a polynomial is a sum-of-squares (SOS), which is a sufficient condition for nonnegativity, is amenable to efficient computation. A polynomial $p$ in indeterminates\footnote{Throughout this paper, the variables $x$ that a polynomial $p(x)$ depend on will be referred to as ``indeterminates". This is to distinguish these variables from \emph{decision variables} in our optimization problems, which will typically be the coefficients of the polynomial.}  $x_1,x_2,\dots,x_n$ is SOS if it can be written as $p(x)=\sum_{i=1}^m q_i^2(x)$ for a set of polynomials $\{q_i\}_{i=1}^m$. This condition is \emph{equivalent} to the existence of a positive semidefinite (psd) matrix $Q$ that satisfies:
\begin{equation}
\label{eq:sos_sdp}
	p(x) = v(x)^T Q v(x), \ \forall x \in \R^n,
\end{equation}
where $v(x)$ is the vector of all monomials with degree less than or equal to half the degree of $p$ \cite{Parrilo00}. Note that the equality constraint \eqref{eq:sos_sdp} imposes linear constraints on the elements of the matrix $Q$ that come from matching coefficients of the polynomials on the left and right hand sides.  Thus, semidefinite programming can be used to certify that a polynomial is a sum of squares. Indeed, by allowing the coefficients of the polynomial $p$ to be decision variables, we can solve optimization problems over the space of SOS polynomials of some fixed degree.   Such optimization problems are referred to as sums-of-squares (SOS) programs. The interested reader is referred to \cite{Parrilo00} and ~\cite[Sections 3,4]{Blekherman13} for a more thorough introduction to SOS programming.

In addition to being able to prove global nonnegativity of polynomials, the SOS programming approach can also be used to demonstrate nonnegativity of polynomials on basic semialgebraic sets (i.e., sets described by a finite number of polynomial inequalities and equalities). Suppose we are given a set $\mathcal{B}$:
\begin{equation}
\mathcal{B} = \{x \in \R^n \ | \ g_{eq,i}(x) = 0, \ g_{ineq,j}(x) \geq 0 \},  
\end{equation}
where $g_{eq,i}$ and $g_{ineq,j}$ are polynomials for $i \in \{1,...,N_{eq}\}, j \in \{1,...,N_{ineq}\}$. We are interested in showing that a polynomial $p$ is nonnegative on the set $\mathcal{B}$:
\begin{equation}
\label{eq:nonnegative_on_set}
x \in \mathcal{B} \implies p(x) \geq 0.
\end{equation} 
We can write the following SOS constraints in order to impose \eqref{eq:nonnegative_on_set}:
\begin{flalign}
q(x) := p(x)\overbrace{- \sum_{i=1}^{N_{eq}} L_{eq,i}(x) g_{eq,i}(x) - \sum_{j=1}^{N_{ineq}} L_{ineq,j}(x) g_{ineq,j}(x)}^{r(x)} \ \textrm{is SOS}&, \label{eq:s_proc_1} \\
L_{ineq,j}(x) \ \textrm{is SOS}&, \forall j \in \{0,...,N_{ineq}\}. \label{eq:s_proc_2}
\end{flalign}
Here, the polynomials $L_{eq,i}$ and $L_{ineq,j}$ are ``multiplier" polynomials analogous to Lagrange multipliers in constrained optimization. In order to see that \eqref{eq:s_proc_1} and \eqref{eq:s_proc_2} are sufficient conditions for \eqref{eq:nonnegative_on_set}, note that when a point $x$ satisfies $g_{eq,i}(x) = 0$ and $g_{ineq,j}(x) \geq 0$ for  $i \in \{1,...,N_{eq}\}, j \in \{1,...,N_{ineq}\}$ (i.e., when $x \in \mathcal{B}$) then the term $r(x)$ is non-positive. Hence, for $q(x)$ to be nonnegative (which must be the case since $q$ is SOS), $p(x)$ must be nonnegative. Thus, we have the desired implication in \eqref{eq:nonnegative_on_set}. This process for using multipliers to impose nonnegativity constraints on sets is known as the generalized S-procedure \cite{Parrilo00} and will be used extensively in Section \ref{sec:funnels} for computing funnels.

%% file: funnels.tex
\section{Computing Funnels}
\label{sec:funnels}

In this section we describe how the tools from Section \ref{sec:background} can be used to compute outer approximations of reachable sets (``funnels") around trajectories of a nonlinear system. The approach in Section \ref{sec:funnels_sos} builds on the work presented in \cite{Tobenkin10b, Tedrake10} while Sections \ref{sec:funnels_uncertainty} and \ref{sec:funnels_controller} are based on \cite{Majumdar12a} and \cite{Majumdar13} respectively. In contrast to this prior work however, we consider the problem of computing outer approximations of forwards reachable sets as opposed to inner approximations of backwards reachable sets. This leads to a few subtle differences in the cost functions of our optimization problems. 

 Consider the following dynamical system:
\begin{equation}
\label{eq:nonlinear_dynamics}
\dot{x} = f(x(t),u(t)),
\end{equation}
where $x(t) \in \R^n$ is the state of the system at time $t$ and $u(t) \in \R^m$ is the control input. Let  $x_0: [0,T] \rightarrow \R^n$ be the nominal trajectory that we would like the system to follow and $u_0:[0,T] \rightarrow \R^m$ be the nominal open-loop control input. Defining new coordinates $\bar{x} = x - x_0(t)$ and $\bar{u} = u - u_0(t)$, we can rewrite the dynamics \eqref{eq:nonlinear_dynamics} in these variables as:
\begin{equation}
\dot{\bar{x}} = \dot{x} - \dot{x}_0 = f(x_0(t)+\bar{x}(t),u_0(t)+\bar{u}(t)) - \dot{x}_0.
\end{equation}
We will first consider the problem of computing funnels for a closed-loop system subject to no uncertainty. To this end, we assume that we are given a feedback controller $\bar{u}_f(t,\bar{x})$ that corrects for deviations around the nominal trajectory (we will consider the problem of designing feedback controllers later in this section). We can then write the closed-loop dynamics of the system as:
\begin{equation}
\label{eq:dynamics_cl}
\dot{\bar{x}} = f_{cl}(t,\bar{x}(t)).
\end{equation}
Given a set of initial conditions $\mathcal{X}_0 \subset \R^n$ with $x_0(0) \in \mathcal{X}_0$, our goal is to find a tight outer approximation of the set of states the system may evolve to at time $t \in [0,T]$. 
In particular, we are concerned with finding sets $F(t) \subset \R^n$ such that:
\begin{equation}
\label{eq:invariance}
\bar{x}(0) \in \mathcal{X}_0 \implies  \bar{x}(t) \in F(t), \  \forall t \in [0,T].
\end{equation}


\begin{definition}
A \emph{funnel} associated with a closed-loop dynamical system $\dot{\bar{x}} = f_{cl}(t,\bar{x}(t))$  is a map $F: [0,T] \rightarrow \mathcal{P}(\R^n)$ from the time-interval $[0,T]$ to the power set (i.e., the set of subsets)  of $\R^n$ such that the sets $F(t)$ satisfy the condition \eqref{eq:invariance} above.
\end{definition}

\quad

Thus, with each time $t \in [0,T]$, the funnel associates a set $F(t) \subset \R^n$. We will parameterize the sets $F(t)$ as sub-level sets of nonnegative time-varying functions $V:[0,T] \times \R^n \rightarrow \R^+$:
\begin{equation}
F(t) = \{\bar{x}(t) \in \R^n | V(t,\bar{x}(t)) \leq \rho(t) \}.
\end{equation}
Letting $\mathcal{X}_0 \subset F(0,\bar{x})$, the following constraint is a sufficient condition for \eqref{eq:invariance}:
\begin{equation}
\label{eq:Vdot condition}
V(t,\bar{x}) = \rho(t) \implies \dot{V}(t,\bar{x}) < \dot{\rho}(t), \ \forall t \in [0,T].
\end{equation}
Here, $\dot{V}$ is computed as:
\begin{equation}
\dot{V}(t,\bar{x}) = \frac{\partial V(t,\bar{x})}{\partial \bar{x}}f_{cl}(t,\bar{x}) + \frac{\partial V(t,\bar{x})}{\partial t}.
\end{equation}
Intuitively, the constraint \eqref{eq:Vdot condition} says that on the boundary of the funnel (i.e., when $V(t,\bar{x}) = \rho(t)$), the function $V$ grows slower than $\rho$. Hence, states on the boundary of the funnel remain within the funnel. This intuition is formalized in \cite{Tedrake10, Tobenkin10b}. 

While \emph{any} function that satisfies \eqref{eq:Vdot condition} provides us with a valid funnel, we are interested in finding \emph{tight} outer approximations of the reachable set. A natural cost function for encouraging tightness is the volume of the sets $F(t)$. Combining this cost function with our constraints, we obtain the following optimization problem:

\begin{flalign} 
\label{eq:funnel_no_uncertainty}
			& & \underset{V,\rho}{\text{inf}}  \hspace*{1cm} & \int_0^T vol(F(t)) \  \textrm{dt} && \\
			& & \text{s.t.} \hspace*{1cm} & V(t,\bar{x}) = \rho(t) \implies \dot{V}(t,\bar{x}) < \dot{\rho}(t), \ \forall t \in [0,T], \nonumber \\
			& & & \mathcal{X}_0 \subset F(0,\bar{x}). \nonumber
\end{flalign}

\revision{Note that our formulation for computing funnels differs in an important respect from a number of prior approaches on feedback motion planning and sequential composition using funnels \cite{Tedrake10, Burridge99,Conner11}. In particular, the formulation presented here considers a given set of initial conditions and attempts to bound the effects of uncertainty going forwards in time. This leads to the volume of funnels being \emph{minimized} rather than maximized. This is in contrast to \cite{Tedrake10, Burridge99, Conner11}, which consider a fixed goal set and seek to maximize the set of initial conditions that are stabilized to this set (which leads to a formulation where the volume of funnels is maximized). The minimization strategy here is better suited to the use of funnels for real-time planning in previously unseen environments since one is interested in bounding the effect of uncertainties on the system in order to prevent collisions with obstacles.}

\subsection{Numerical implementation using SOS programming}
\label{sec:funnels_sos}

Since the optimization problem \eqref{eq:funnel_no_uncertainty} involves searching over spaces of functions, it is infinite dimensional and hence not directly amenable to numerical computation. However, we can use the SOS programming approach described in Section \ref{sec:background} to obtain finite-dimensional optimization problems in the form of semidefinite programs (SDPs). We first concentrate on implementing the constraints in \eqref{eq:funnel_no_uncertainty}. We will assume that the initial condition set $\mathcal{X}_0$ is a semi-algebraic set (i.e., described in terms of polynomial inequalities):
\begin{equation}
\mathcal{X}_0 = \{\bar{x} \in \R^n \ | \ g_{0,i}(\bar{x}) \geq 0, \ \forall i = 1,\dots,N_0\}.
\end{equation}
Then the constraints in \eqref{eq:funnel_no_uncertainty} can be written as:
\begin{flalign}
V(t,\bar{x}) = \rho(t) \implies \dot{\rho}(t) - \dot{V}(t,\bar{x}) &> 0 \label{eq:funnel_poly_1} \\
g_{0,i}(\bar{x}) \geq 0 \ \forall i \in \{1,\dots,N_0\} \implies \rho(0) - V(0,\bar{x}) &\geq 0. \label{eq:funnel_poly_2}
\end{flalign}
If we restrict ourselves to polynomial dynamics and polynomial functions $V$ and $\rho$, these constraints are precisely in the form of \eqref{eq:nonnegative_on_set} in Section \ref{sec:background_sos}. We can thus apply the procedure described in Section \ref{sec:background_sos} and arrive at the following sufficient conditions for \eqref{eq:funnel_poly_1} and \eqref{eq:funnel_poly_2}:
\begin{flalign}
\dot{\rho}(t) - \dot{V}(t,\bar{x}) - L(t,\bar{x}) [V(t,\bar{x}) - \rho(t)] - L_t(t,\bar{x})[t(T-t)] \ \textrm{is SOS}&, \label{eq:funnel_sos_1} \\
\rho(0) - V(0,\bar{x}) - \sum_i^{N_0} L_{0,i}(\bar{x}) g_{0,i}(\bar{x}) \ \textrm{is SOS}&,\label{eq:funnel_sos_2}  \\
L_t(t,\bar{x}), L_{0,i}(\bar{x}) \ \textrm{are SOS}&, \forall i \in \{1,\dots,N_0\}. \label{eq:funnel_sos_3} 
\end{flalign}
As in Section \ref{sec:background_sos}, the polynomials $L, L_t$ and $L_{0,i}$ are ``multiplier" polynomials whose coefficients are decision variables. 

Next, we focus on approximating the cost function in \eqref{eq:funnel_no_uncertainty} using semidefinite programming. This can be achieved by sampling in time and replacing the integral with the finite sum $\sum_{k = 1}^N vol(F(t_k))$. In the special case where the function $V$ is quadratic in $\bar{x}$:
\begin{equation}
V(t_k,\bar{x}) = \bar{x}^T S_k \bar{x}, \ S_k \succeq 0,
\end{equation}
the set $F(t_k)$ is an ellipsoid and we can use semidefinite programming (SDP) to directly minimize the volume by maximizing the determinant of $S_k$ (recall that the volume of the ellipsoid $F(t_k)$ is a monotonically decreasing function of the determinant of $S_k$). Note that while the problem of maximizing the determinant of a psd matrix is not directly a problem of the form \eqref{eq:sdp}, it can be transformed into such a form ~\cite[Section3]{Ben-Tal01}. Further note that the fact that our cost function can be handled directly in the SDP framework is in distinction to the approaches for computing inner approximations of backwards reachable sets \cite{Tedrake10} \cite{Tobenkin10b} \cite{Majumdar13}. This is because the determinant of a psd matrix is a \emph{concave} function and hence \emph{minimizing} the determinant is not a convex problem. Hence, in the previous work, the authors used heuristics for maximizing volume. 

In the more general case, we can minimize an upper bound on the cost function \\ $\sum_{k = 1}^N vol(F(t_k))$ by introducing ellipsoids $\mathcal{E}(t_k)$:
\begin{equation}
\mathcal{E}(t_k) = \{\bar{x} \in \R^n | \bar{x}^T S_k \bar{x} \leq 1, \ S_k \succeq 0\}
\end{equation}
such that $F(t_k) \subset \mathcal{E}(t_k)$ and minimizing $\sum_{k=1}^N vol(\mathcal{E}(t_k))$. The containment constraint can be equivalently expressed as the constraint:
\begin{equation}
V(t_k,\bar{x}) \leq \rho(t_k) \implies \bar{x}^T S_k \bar{x} \leq 1,
\end{equation}
and can thus be imposed using SOS constraints:
\begin{flalign}
1 - \bar{x}^T S_k \bar{x} - L_{\mathcal{E},k}(\bar{x}) [\rho(t_k) - V(t_k,\bar{x})] \ &\textrm{is SOS}, \\
L_{\mathcal{E},k}(\bar{x}) \ &\textrm{is SOS}.
\end{flalign}
Combining our cost function with the constraints \eqref{eq:funnel_sos_1} - \eqref{eq:funnel_sos_3}, we obtain the following optimization problem:
\begin{flalign}
\label{eq:funnel_no_uncertainty_sos}
			& & \underset{V,\rho,L,L_t,L_{0,i},S_k,L_{\mathcal{E},k}}{\text{inf}}  \hspace*{1cm} & \sum_{k=1}^N vol(\mathcal{E}(t_k)) = \sum_{k=1}^N vol(\{\bar{x} | \bar{x}^T S_k \bar{x} \leq 1\})  \\
			& & \text{s.t.} \hspace*{1cm} & \dot{\rho}(t) - \dot{V}(t,\bar{x}) - L(t,\bar{x}) [V(t,\bar{x}) - \rho(t)] - L_t(t,\bar{x})[t(T-t)] \ \textrm{is SOS},  \label{eq:funnel_no_uncertainty_sos_1}  \\
			& & & \rho(0) - V(0,\bar{x}) - \sum_i^{N_0} L_{0,i}(\bar{x}) g_{0,i}(\bar{x}) \ \textrm{is SOS}, \nonumber \\
			& & & 1 - \bar{x}^T S_k \bar{x} - L_{\mathcal{E},k}(\bar{x}) [\rho(t_k) - V(t_k,\bar{x})]  \ \textrm{is SOS}, &  \hspace*{-5.25cm} \forall k \in \{1,\dots,N\}, \nonumber \\
		 	& & & S_k \succeq 0, & \hspace*{-5.25cm} \forall k \in \{1,\dots,N\}, \nonumber \\
			& & & L_t(t,\bar{x}), L_{0,i}(\bar{x}), L_{\mathcal{E},k}(\bar{x}) \ \textrm{are SOS}, & \hspace*{-7.25cm} \forall i \in \{1,\dots,N_0\}, \ \forall k \in \{1,\dots,N\}. \nonumber
\end{flalign}
While this optimization problem is finite dimensional, it is non-convex in general since the first constraints are \emph{bilinear} in the decision variables (e.g., the coefficients of the polynomials $L$ and $V$ are multiplied together in the first constraint). To apply SOS programming, we require the constraints to be linear in the coefficients of the polynomials we are optimizing. However, note that when $V$ and $\rho$ are fixed, the constraints are linear in the other decision variables. Similarly, when the multipliers $L$ and $L_{\mathcal{E},k}$ are fixed, the constraints are linear in the remaining decision variables. Thus, we can efficiently perform this optimization by alternating between the two sets of decision variables $(L, L_t, L_{0,i},S_k,L_{\mathcal{E},k})$ and $(V,\rho,L_t,L_{0,i},S_k)$. In each step of the alternation, we can optimize our cost function $\sum_{k=1}^N vol(\mathcal{E}(t_k))$. These alternations are summarized in Algorithm \ref{a:funnel_computation}. \revision{Note that Algorithm \ref{a:funnel_computation} requires an initialization for $V$ and $\rho$. We will discuss how to obtain these in Section \ref{sec:funnel_implementation}.}

\begin{algorithm}[h!]
  \caption{Funnel Computation}
  \label{a:funnel_computation}
  \begin{algorithmic}[1]
    \STATE Initialize $V$  and $\rho$.
    \STATE $cost_{prev} = \infty$;
    \STATE converged $=$ false;
    \WHILE{$\neg$converged}
    	\STATE $\bf{STEP}$ $\bf{1}:$ Minimize $\sum_{k=1}^N vol(\mathcal{E}(t_k))$ by searching for multiplier polynomials $(L, L_t, L_{0,i},L_{\mathcal{E},k})$ and $S_k$ while fixing $V$ and $\rho$.
	\STATE $\bf{STEP}$ $\bf{2}:$ Minimize $\sum_{k=1}^N vol(\mathcal{E}(t_k))$ by searching for $(V,\rho,L_t,L_{0,i},S_k)$ while fixing $L$ and $L_{\mathcal{E},k}$. 
	\STATE $cost = \sum_{k=1}^N vol(\mathcal{E}(t_k))$;
	\IF{$\frac{cost_{prev} - cost}{cost_{prev}} <  \epsilon$}
		\STATE converged = true;
	\ENDIF
	\STATE $cost_{prev} = cost$;
    \ENDWHILE    
  \end{algorithmic}
\end{algorithm}

\begin{remark}
\label{rem:convergence}
It is easy to see that Algorithm \ref{a:funnel_computation} converges (though not necessarily to an optimal solution). Each iteration of the alternations is guaranteed to achieve a cost function that is at least as good as the previous iteration (since the solution from the previous iteration is a valid one). Hence, the sequence of optimal values in each iteration form a monotonically non-increasing sequence. Combined with the fact that the cost function is bounded below by 0, we conclude that this sequence converges and hence that Algorithm \ref{a:funnel_computation} terminates.
\end{remark}

\vspace{10pt}
\subsection{Approximation via time-sampling}
\label{sec:funnels_time_sampling}

As observed in \cite{Tobenkin10b} in practice it is often the case that the nominal trajectory $x_0:[0,T] \rightarrow \R^n$ is difficult to approximate with a low degree polynomial in time. This can lead to the constraint \eqref{eq:funnel_no_uncertainty_sos_1} in the problem \eqref{eq:funnel_no_uncertainty_sos} having a high degree polynomial dependence on $t$. Thus it is often useful to implement an approximation of the optimization problem \eqref{eq:funnel_no_uncertainty_sos} where the condition \eqref{eq:funnel_poly_1} is checked only at a finite number of sample points $t_k \in [0,T], \ k \in \{1,\dots,N\}$. We can use a piecewise linear parameterization of $\rho$ and can thus compute:
\begin{equation}
\dot{\rho}(t_k) = \frac{\rho(t_{k+1}) - \rho(t_k)}{t_{k+1} - t_k}.
\end{equation}
Similarly we can parameterize the function $V$ by polynomials $V_k(\bar{x})$ at each time sample and compute:
\begin{equation}
\frac{\partial{V(t,\bar{x})}}{\partial{t}} \approx \frac{V_{k+1}(\bar{x}) - V_k(\bar{x})}{t_{k+1} - t_k}.
\end{equation}
We can then write the following modified version of the problem \eqref{eq:funnel_no_uncertainty_sos}:
\begin{flalign}
\label{eq:funnel_no_uncertainty_sampling}
			& & \underset{V_k,\rho,L_k,L_{0,i},S_k,L_{\mathcal{E},k}}{\text{inf}}  \hspace*{1cm} & \sum_{k=1}^N vol(\mathcal{E}(t_k)) = \sum_{k=1}^N vol(\{\bar{x} | \bar{x}^T S_k \bar{x} \leq 1\})  \\
			& & \text{s.t.} \hspace*{1cm} & \dot{\rho}(t_k) - \dot{V}_k(\bar{x}) - L_k(\bar{x}) [V_k(\bar{x}) - \rho(t_k)],  & &&\forall k \in \{1,\dots,N\}, \nonumber  \\
			& & & \rho(t_1) - V_1(\bar{x}) - \sum_i^{N_0} L_{0,i}(\bar{x}) g_{0,i}(\bar{x}) \ \textrm{is SOS}, \nonumber \\
			& & & 1 - \bar{x}^T S_k \bar{x} - L_{\mathcal{E},k}(\bar{x}) [\rho(t_k) - V_k(\bar{x})]  \ \textrm{is SOS}, & &&\forall k \in \{1,\dots,N\}, \nonumber \\
			& & & S_k \succeq 0, & &&\forall k \in \{1,\dots,N\}, \nonumber \\
			& & & L_{0,i}(\bar{x}), L_{\mathcal{E},k}(\bar{x}) \ \textrm{are SOS}, & && \hspace*{-5.25cm} \forall i \in \{1,\dots,N_0\}, \ \forall k \in \{1,\dots,N\}. \nonumber 
\end{flalign}
This program does not have any algebraic dependence on the variable $t$ and can thus provide significant computational gains over \eqref{eq:funnel_no_uncertainty_sos}. However, it does not provide an exact funnel certificate. One would hope that with a sufficiently fine sampling in time, one would recover exactness. Partial results in this direction are provided in \cite{Tobenkin10b} along with numerical examples showing that the loss of accuracy from the sampling approximation can be quite small in practice. 

The problem \eqref{eq:funnel_no_uncertainty_sampling} is again bilinear in the decision variables. It is linear in the two sets of decision variables $(L_k, L_{0,i},S_k,L_{\mathcal{E},k})$ and $(V_k,\rho,L_{0,i},S_k)$. Thus, Algorithm \ref{a:funnel_computation} can be applied directly to \eqref{eq:funnel_no_uncertainty_sampling} with the minor modification that $V$ and $\rho$ are replaced by their time-sampled counterparts and the multipliers $(L,L_t)$ are replaced by the multipliers $L_k$. 

\subsection{Extensions to the basic algorithm}
\label{sec:funnels_extensions}

Next we describe several extensions to the basic framework for computing funnels described in Section \ref{sec:funnels_sos}. Section \ref{sec:funnels_uncertainty} discusses the scenario in which the dynamics of the system are subject to bounded disturbances/uncertainty, Section \ref{sec:funnels_controller} considers the problem of synthesizing feedback controllers that explicitly attempt to minimize the size of the funnel, Section \ref{sec:funnels_saturations} demonstrates how to handle input saturations, and Section \ref{sec:funnels_costs} considers a generalization of the cost function. 

\subsubsection{Uncertainty in the dynamics}
\label{sec:funnels_uncertainty}

Suppose that the dynamics of the system are subject to an uncertainty term $w(t) \in \R^d$ that models external disturbances or parametric model uncertainties. The closed-loop dynamics \eqref{eq:dynamics_cl} can then be modified to capture this uncertainty:
\begin{equation}
\dot{\bar{x}} = f_{cl}(t,\bar{x}(t),w(t)).
\end{equation}
We will assume that the dynamics $f_{cl}$ depend polynomially on $w$.
Given an initial condition set $\mathcal{X}_0 \subset \R^n$ as before, our goal is to find sets $F(t)$ such that $x(t)$ is guaranteed to be in $F(t)$ for any valid disturbance profile:
\begin{equation}
\label{eq:invariance_disturbance}
\bar{x}(0) \in \mathcal{X}_0 \implies \bar{x}(t) \in F(t), \forall t \in [0,T] ,\forall w:[0,T] \rightarrow \mathcal{W}.
\end{equation}
Parameterizing the sets $F(t)$ as sub-level sets of nonnegative time-varying functions $V: [0,T] \times \R^n \rightarrow \R^+$ as before, the following condition is sufficient to ensure \eqref{eq:invariance_disturbance}:
\begin{equation}
V(t,\bar{x}) = \rho(t) \implies \dot{V}(t,\bar{x},w) < \dot{\rho}(t), \forall t \in [0,T], \forall w(t) \in \mathcal{W},
\end{equation}
where $\dot{V}$ is computed as:
\begin{equation}
\dot{V}(t,\bar{x},w) = \frac{\partial V(t,\bar{x})}{\partial \bar{x}}f_{cl}(t,\bar{x},w) + \frac{\partial V(t,\bar{x})}{\partial t}.
\end{equation}
This is almost identical to the condition \eqref{eq:Vdot condition}, with the exception that the function $V$ is required to decrease on the boundary of the funnel for every choice of disturbance. Assuming that the set $\mathcal{W}$ is a semi-algebraic set $\mathcal{W} = \{w \in \R^d \ | \ g_{w,j}(w) \geq 0, \forall j = 1,\dots,N_w\}$, the optimization problem \eqref{eq:funnel_no_uncertainty_sos} is then easily modified by replacing condition \eqref{eq:funnel_no_uncertainty_sos_1} with the following constraints:
\begin{flalign}
& \label{eq:sos_constraints_uncertainty} \\
\dot{\rho}(t) - \dot{V}(t,\bar{x},w) - L(t,\bar{x},w) [V(t,\bar{x}) - \rho(t)] - L_t(t,\bar{x},w)[t(T-t)] \dots \nonumber \\
 \dots  \ - \sum_{j=1}^{N_w} L_{w,j}(t,\bar{x},w)g_{w,j}(w) \  \textrm{is SOS},& \nonumber \\
L_{w,j}(t,\bar{x},w) \ \textrm{is SOS}, \forall j = \{1,\dots,N_w\}.& \nonumber
\end{flalign}
These SOS constraints now involve polynomials in the indeterminates $t,\bar{x}$ and $w$. Since these constraints are linear in the coefficients of the newly introduced multipliers $L_{w,j}$, Algorithm \ref{a:funnel_computation} can be directly applied to the modified optimization problem by adding $L_{w,j}$ to the list of polynomials to be searched for in both Step 1 and Step 2 of the iterations. Similarly, the time-sampled approximation described in Section \ref{sec:funnels_time_sampling} can also be applied to \eqref{eq:sos_constraints_uncertainty}.

\subsubsection{Feedback control synthesis}
\label{sec:funnels_controller}

So far we have assumed that we have been provided with a feedback controller that corrects for deviations around the nominal trajectory. We now consider the problem of \emph{optimizing} the feedback controller in order to minimize the size of the funnel. We will assume that the system is control affine:
\begin{equation}
\dot{x} = f(x(t)) + g(x(t))u(t),
\end{equation}
and parameterize the control policy as a polynomial $\bar{u}_f(t,\bar{x})$. We can thus write the dynamics in the $\bar{x}$ coordinates as:
\begin{equation}
\dot{\bar{x}} = f(x_0(t)+\bar{x}(t)) + g(x(t))[u_0(t) + \bar{u}_f(t,\bar{x})] - \dot{x}_0.
\end{equation}
The feedback controller can then be optimized by adding the coefficients of the polynomial $\bar{u}_f(t,\bar{x})$ to the set of decision variables in the optimization problem \eqref{eq:funnel_no_uncertainty_sos} while keeping all the constraints unchanged. Note that $\bar{u}_f$ appears in the constraints only through $\dot{V}$, which is now bilinear in the coefficients of $V$ and $\bar{u}_f$ since:
\begin{equation}
\dot{V}(t,\bar{x}) = \frac{\partial V(t,\bar{x})}{\partial \bar{x}}\dot{\bar{x}} + \frac{\partial V(t,\bar{x})}{\partial t}.
\end{equation}
With the (coefficients of) the feedback controller $\bar{u}_f$ as part of the optimization problem, note that the constraints of the problem \eqref{eq:funnel_no_uncertainty_sos} are now bilinear in the two sets of decision variables $(L, L_t, L_{0,i},S_k,L_{\mathcal{E},k}, \bar{u}_f)$ and $(V,\rho,L_t,L_{0,i},S_k)$. Thus, in principle we could use a bilinear alternation scheme similar to the one in Algorithm \ref{a:funnel_computation} and alternatively optimize these two sets of decision variables. However, in this case we would not be searching for a controller that explicitly seeks to minimize the size of the funnel (since the controller would not be searched for at the same time as $V$ or $\rho$, which define the funnel). To get around this issue, we add another step in each iteration where we optimize our cost function $\sum_{k=1}^N vol(\mathcal{E}(t_k))$ by searching for $(\bar{u}_f,\rho, L_t,L_{0,i},S_k)$  while keeping $(V,L,L_{\mathcal{E},k})$ fixed. This allows us to search for $\bar{u}_f$ and $\rho$ at the same time, which can significantly improve the quality of the controllers and funnels we obtain. These steps are summarized in Algorithm \ref{a:control_design}. By a reasoning identical to the one in Remark \ref{rem:convergence} it is easy to see that the sequence of optimal values produced by Algorithm \ref{a:control_design} converges.

\begin{algorithm}[h!]
  \caption{Feedback Control Synthesis}
  \label{a:control_design}
  \begin{algorithmic}[1]
    \STATE Initialize $V$  and $\rho$.
    \STATE $cost_{prev} = \infty$;
    \STATE converged $=$ false;
    \WHILE{$\neg$converged}
    	\STATE $\bf{STEP}$ $\bf{1}:$ Minimize $\sum_{k=1}^N vol(\mathcal{E}(t_k))$ by searching for controller $\bar{u}_f$ and $(L, L_t, L_{0,i},L_{\mathcal{E},k}, S_k)$ while fixing $V$ and $\rho$.
	\STATE $\bf{STEP}$ $\bf{2}:$ Minimize $\sum_{k=1}^N vol(\mathcal{E}(t_k))$ by searching for controller $\bar{u}_f$ and $(\rho, L_t,L_{0,i},S_k)$ while fixing $(V,L,L_{\mathcal{E},k})$.
	\STATE $\bf{STEP}$ $\bf{3}:$ Minimize $\sum_{k=1}^N vol(\mathcal{E}(t_k))$ by searching for $(V,\rho,L_t,L_{0,i},S_k)$ while fixing $(L,L_{\mathcal{E},k},\bar{u}_f)$. 
	\STATE $cost = \sum_{k=1}^N vol(\mathcal{E}(t_k))$;
	\IF{$\frac{cost_{prev} - cost}{cost_{prev}} <  \epsilon$}
		\STATE converged = true;
	\ENDIF
	\STATE $cost_{prev} = cost$;
    \ENDWHILE    
  \end{algorithmic}
\end{algorithm}

We note that the approach for taking into account uncertainty described in Section \ref{sec:funnels_uncertainty} can easily be incorporated into Algorithm \ref{a:control_design}. Similarly, by parameterizing the controller $\bar{u}_f$ as polynomials $u_{f,k}(\bar{x})$ at the times $t_k$, we can also apply the time-sampled approximation described in Section \ref{sec:funnels_time_sampling}.

\subsubsection{Actuator saturations}
\label{sec:funnels_saturations}

{\bf{A. Approach 1}} 

\quad

Our approach also allows us to incorporate actuator limits into the verification procedure. Although we examine the single-input case in this section, this framework is easily extended to handle multiple inputs. Let the control input $u(t)$ at time $t$ be mapped through the following control saturation function:
$$s(u(t)) = \begin{cases} u_{max} & \mbox{if $u(t) \geq u_{max}$}  \\ 
u_{min} & \mbox{if $u(t) \leq u_{min}$} \\  
u(t) & \mbox{o.w.} 
\end{cases}$$
Then, in a manner similar to \cite{Tedrake10}, a piecewise analysis of $\dot{V}(t,\bar{x})$ can be used to check the Lyapunov conditions are satisfied even when the control input saturates. Defining:
\begin{flalign} 
  \dot{V}_{min}(t,\bar{x}) := & \frac{\partial{V(t,\bar{x})}}{\partial{\bar{x}}} \Big{(}f(x_0+\bar{x}) + g(x_0+\bar{x})u_{min}\Big{)} + \frac{\partial{V(t,\bar{x})}}{\partial{t}}, \\
  \dot{V}_{max}(t,\bar{x}) := & \frac{\partial{V(t,\bar{x})}}{\partial{\bar{x}}} \Big{(}f(x_0+\bar{x}) + g(x_0+\bar{x})u_{max}\Big{)} + \frac{\partial{V(t,\bar{x})}}{\partial{t}},
\end{flalign}
we must check the following conditions:
\begin{flalign} \label{eq:tvSOS}
  & u_0(t)+\bar{u}_f(t,\bar{x}) \leq u_{min}, \ V(t,\bar{x}) = \rho(t) \implies \dot{V}_{min}(t,\bar{x}) < \dot{\rho}(t), \\
  & u_0(t)+\bar{u}_f(t,\bar{x}) \geq u_{max}, \ V(t,\bar{x}) = \rho(t) \implies \dot{V}_{min}(t,\bar{x}) < \dot{\rho}(t), \\
  & u_{min} \leq u_0(t)+\bar{u}_f(t,\bar{x}) \leq u_{max}, \ V(t,\bar{x}) = \rho(t) \implies \dot{V}(t,\bar{x}) < \dot{\rho}(t),
\end{flalign}
where $u_0$ is the open-loop control input and $\bar{u}_f$ is the feedback controller as before.
These conditions can be enforced by adding additional multipliers to the optimization program \eqref{eq:funnel_no_uncertainty_sos} or its time-sampled counterpart \eqref{eq:funnel_no_uncertainty_sampling}.

\quad

\noindent{\bf{B. Approach 2}} 

\quad

Although one can handle multiple inputs via the above method, the number of SOS conditions grows exponentially with the number of inputs ($3^m$ conditions for $\dot{V}$ are needed in general to handle all possible combinations of input saturations). Thus, for systems with a large number of inputs, an alternative formulation was proposed in \cite{Majumdar13} that avoids this exponential growth in the size of the SOS program at the cost of adding conservativeness to the size of the funnel. Given limits on the control vector $u \in \RR^m$ of the form $u_{min} < u < u_{max}$, we can ask to satisfy:
\begin{equation}
\bar{x} \in F(t) \implies u_{min} < u_0(t) + \bar{u}_f(t,\bar{x}) < u_{max}, \ \forall t \in [0,T].
\end{equation}
This constraint implies that the applied control input remains within the specified bounds inside the verified funnel (a conservative condition). The number of extra constraints grows linearly with the number of inputs (since we have one new condition for every input), thus leading to smaller optimization problems.

\subsubsection{A more general cost function}
\label{sec:funnels_costs}

We end our discussion of extensions to the basic algorithm for computing funnels presented in Section \ref{sec:funnels_sos} by considering a generalization of the cost function (volume of the funnel) we have used so far. In particular, it is sometimes useful to minimize the volume of the funnel \emph{projected} onto a subspace of the state space. Suppose this projection map is given by $\pi: \R^n \rightarrow \R^{n_p}$ with a corresponding $n_p \times n$ projection matrix $P$.    For an ellipsoid $\mathcal{E} = \{\bar{x} \in \R^n \ | \ \bar{x}^T S_k \bar{x} \leq 1\}$, the projected set $\pi(\mathcal{E})$ is also an ellipsoid $\mathcal{E}_p = \{\bar{x} \in \R^{n_p} \ | \ \bar{x}^T S_k^{(p)} \bar{x} \leq 1\}$ with:
\begin{equation}
\label{eq:ellipsoid projection}
S_k^{(p)} = (PS_k^{-1}P^T)^{-1}.
\end{equation}
Recall that the ability to minimize the volume of the ellipsoid $\mathcal{E}$ using SDP relied on being able to maximize the determinant of $S_k$. In order to minimize the volume of $\mathcal{E}_p$, we would have to maximize det$(S_k^{(p)})$, which is a complicated (i.e. nonlinear) function of $S_k$.  Hence, in each iteration of Algorithm \ref{a:funnel_computation} we linearize the function det$(S_k^{(p)})$ with respect to $S_k$ at the solution of $S_k$ from the previous iteration and maximize this linearization instead. The linearization of det($S_k^{(p)}$) with respect to $S_k$ at a nominal value $S_{k,0}$ can be explicitly computed as:
\begin{equation}
\textrm{Tr}\Big{(}P^T(PS_{k,0}^{-1}P^T)^{-1}PS_{k,0}^{-1}S_kS_{k,0}^{-1}\Big{)},
\end{equation}
where Tr refers to the trace of the matrix.

\subsection{Implementation details}
\label{sec:funnel_implementation}

We end this section on computing funnels by discussing a few important implementation details.

\subsubsection{Trajectory generation}
\label{sec:traj opt}

An important step that is necessary for the success of our approach to computing funnels is the generation of a dynamically feasible open-loop control input $u_0: [0,T] \mapsto \R^m$ and corresponding nominal trajectory $x_0: [0,T] \mapsto \R^n$. A method that has been shown to work well in practice and scale to high dimensions is the direct collocation trajectory optimization method \cite{Betts01}. While this is the approach we use for the results in Section \ref{sec:examples}, other methods like the Rapidly Exploring Randomized Tree (RRT) or its asymptotically optimal version, RRT$^{\star}$ can be used too \cite{Kuffner00,Karaman11}.


\subsubsection{Initializing $V$ and $\rho$}
\label{sec:initialization}

Algorithms \ref{a:funnel_computation} and \ref{a:control_design} require an initial guess for the functions $V$ and $\rho$. In \cite{Tedrake10}, the authors use the Lyapunov function candidate associated with a time-varying LQR controller. The control law is obtained by
solving a Riccati differential equation:
\begin{equation}-\dot S(t) = Q - S(t)B(t)R^{-1}B^TS(t) + S(t)A(t) + A(t)^T S(t)\end{equation}
with final value conditions $S(t) = S_f$. Here $A(t)$ and $B(t)$ describe the time-varying linearization of the dynamics about the nominal trajectory $x_0$. The matrices $Q$ and $R$ are positive-definite cost-matrices. The function:
\begin{equation}V_{guess}(t,\bar{x}) = (x-x_0(t))^T S(t) (x - x_0(t)) = \bar{x}^TS(t)\bar{x}\end{equation}
is our initial Lyapunov candidate. Setting $\rho$ to a quickly increasing function such as an exponential is typically sufficient to obtain a feasible initialization. 
\revision{However, note that since Algorithms  \ref{a:funnel_computation} and \ref{a:control_design} are not guaranteed to converge to globally optimal solutions, their initialization can impact the quality of the resulting solutions. Thus, standard strategies such as multiple (possibly random) initializations can help address issues related to local solutions. Another strategy that can be used when computing libraries of funnels (Section \ref{sec:funnel libraries}) is to initialize a funnel computation with the solution computed from a neighboring maneuver in the library.}

%% file: funnel_libraries.tex
\section{Funnel Libraries}
\label{sec:funnel libraries}

\subsection{Sequential composition}
\label{sec:sequential composition}

One can think of funnels computed using the machinery described in Section \ref{sec:funnels} as \emph{robust} motion primitives (the robustness is to initial conditions and uncertainty in the dynamics).
While we could define a \emph{funnel library} simply as a collection $\F$ of funnels and associated feedback controllers, it will be fruitful to associate some additional structure with $\F$. In particular, it is useful to know how funnels can be \emph{sequenced} together to form composite robust motion plans. In order to consider this more formally, we will first introduce the notion of \emph{sequential composition} of funnels \cite{Burridge99}. \revision{We note that our definition of sequential composition below differs slightly from the one proposed in \cite{Burridge99}, which considers infinite-horizon control policies in contrast to the finite-horizon funnels we consider here. In this regard, our notion of a funnel (and corresponding notion of sequential composition) is akin to that of \emph{conditional invariance} considered in \cite{Kantor03, Conner11}.}

\quad

\begin{definition}
\label{def:seq_comp}
An ordered pair $(F_1,F_2)$ of funnels $F_1:[0,T_1] \rightarrow \mathcal{P}(\R^n)$ and $F_2:[0,T_2] \rightarrow \mathcal{P}(\R^n)$ is \emph{sequentially composable} if $F_1(T_1) \subset F_2(0)$.
\end{definition}
\quad

\begin{figure*}
  \begin{center}
    \includegraphics[trim = 0mm 0mm 0mm 0mm, clip, width=0.7\textwidth]{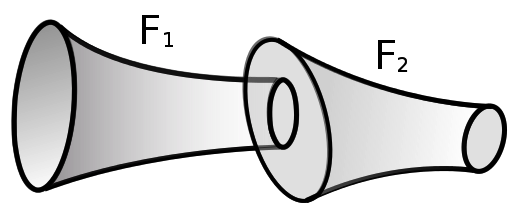}
  \end{center}
  \caption[Sequential composition of funnels]{The ordered pair of funnels $(F_1,F_2)$ is sequentially composable. The outlet of $F_1$ is contained within the inlet of $F_2$, i.e., $F_1(T_1) \subset F_2(0)$.
  \label{fig:funnel_seq_comp}}
\end{figure*}

\noindent In other words, two funnels are sequentially composable if the ``outlet" of one is contained within the ``inlet" of the other. A pictorial depiction of this is provided in Figure \ref{fig:funnel_seq_comp}. We note that the sequential composition of two such funnels is itself a funnel. 

\subsection{Exploiting invariances in the dynamics}
\label{sec:sqmi}

For our purposes here, it is useful to introduce a slightly generalized notion of sequential composability that will allow us to exploit \emph{invariances} (continuous symmetries) in the dynamics. In particular, the dynamics of large classes of mechanical systems such as mobile robots are often invariant under certain transformations of the state space. For Lagrangian systems, the notion of ``cyclic coordinates" captures such invariances. A cyclic coordinate is a (generalized) coordinate of the system that the Lagrangian does \emph{not} depend on. We can then write the dynamics of the system $\dot{x} = f(x(t),u(t))$ as a function of a state vector $x = [x_c, x_{nc}]$ which is partitioned into cyclic coordinates $x_c$ and non-cyclic coordinates $x_{nc}$ in such a way that the dynamics only depend on the non-cyclic coordinates:
\begin{equation}
\dot{x} = f(x_{nc}(t),u(t)).
\end{equation}
For example, the dynamics of a quadrotor or fixed-wing airplane (expressed in an appropriate coordinate system) do not depend on the $x-y-z$ position of the system or the yaw angle.

Invariance of the dynamics of the system also implies that if a curve $t \mapsto (x(t),u(t))$ is a valid solution to the dynamics $\dot{x} = f(x(t),u(t))$, then so is the transformed solution:
\begin{equation}
t \mapsto (\Psi_{c} (x(t)),u(t)),
\end{equation}
where $\Psi_{c}$ \revision{is a mapping from the state space to itself that shifts (i.e., translates) the state vector along the cyclic coordinates (and does not transform the non-cyclic coordinates).} This allows us to make the following important observation. 

\quad

\begin{remark}
\label{rem:shifting funnels}
Suppose we are given a system whose dynamics are invariant to \revision{shifts} $\Psi_c$ along cyclic coordinates $x_c$. Let $F:[0,T] \rightarrow \mathcal{P}(\R^n)$ given by $t \mapsto F(t)$ be a funnel associated with this system. Then the transformed funnel given by $t \mapsto \Psi_c(F(t))$ is also a valid funnel. Hence, one can in fact think of invariances in the dynamics giving rise to an \emph{infinite} family of funnels parameterized by shifts $\Psi_c(F)$ of a funnel $F$ along cyclic coordinates of the system.
\end{remark}
 
\quad

\noindent Note that here we have implicitly assumed that the feedback controller:
\begin{equation}
u_f(t,x) = u_0(t) + \bar{u}_f(t,\bar{x}) = u_0(t) + \bar{u}_f(t,x-x_0(t))
\end{equation}
associated with the funnel has also been transformed to:
\begin{equation}
u_0(t) + \bar{u}_f(t,x-\Psi_c(x_0(t))).
\end{equation}
In other words, we have shifted the reference trajectory we are tracking by $\Psi_c$. Henceforth, when we refer to transformations of funnels along cyclic coordinates we will implicitly assume that the feedback controller has also been appropriately modified in this manner.

These observations allow us to define a generalized notion of sequential composition that exploits invariances in the dynamics. We will refer to this notion as sequential composition \emph{modulo invariances (MI)}.

\quad

\begin{definition}
\label{def:seq_comp_inv}
An ordered pair $(F_1,F_2)$ of funnels $F_1:[0,T_1] \rightarrow \mathcal{P}(\R^n)$ and $F_2:[0,T_2] \rightarrow \mathcal{P}(\R^n)$ is \emph{sequentially composable modulo invariances (MI)} if there exists a \revision{shift} $\Psi_c$ of the state along cyclic coordinates such that $F_1(T_1) \subset \Psi_c\big{(}F_2(0)\big{)}$.
\end{definition}

\quad

\begin{figure*}
  \begin{center}
    \includegraphics[trim = 0mm 0mm 0mm 0mm, clip, width=0.7\textwidth]{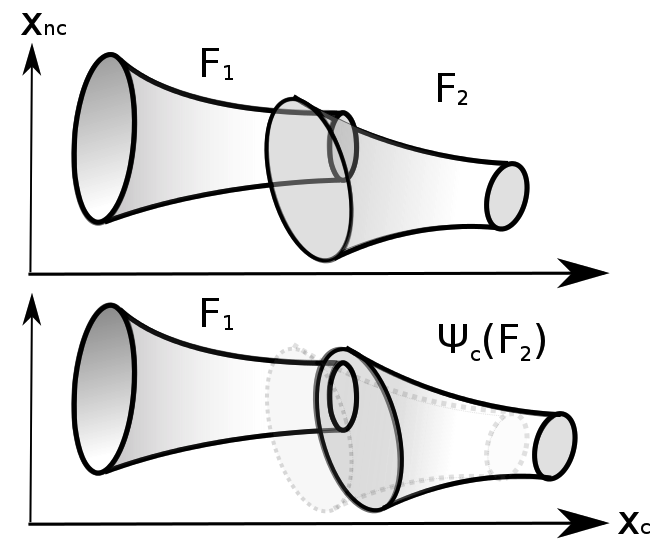}
  \end{center}
  \caption[Sequential composition modulo invariances]{Sequential composition modulo invariances. The top row of the figure shows two funnels that are \emph{not} sequentially composable in the sense of Definition \ref{def:seq_comp}. However, as shown in the bottom row of the figure, they \emph{are} sequentially composable in the more general sense of Definition \ref{def:seq_comp_inv}. By shifting the funnel $F_2$ (whose outline is plotted using dotted lines for reference) in the cyclic coordinate ($x_c$), we can ensure that the outlet of $F_1$ lies in the inlet of this shifted funnel $\Psi_c(F_2)$. 
  \label{fig:funnel_seq_comp_invariance}}
\end{figure*}

\noindent Informally, two funnels $F_1$ and $F_2$ are sequentially composable in this generalized sense if one can shift $F_2$ along the cyclic coordinates of the system and ensure that its inlet contains the outlet of $F_1$ \revision{(note that the definition requires the entire inlet of the shifted funnel to contain the outlet of $F_1$, i.e., not just containment in the cyclic coordinates).} Figure \ref{fig:funnel_seq_comp_invariance} provides a pictorial depiction of this.  

One may think of sequential composability MI of funnels as being analogous to the compatibility condition required for sequencing trajectories in the library of a Maneuver Automaton \cite{Frazzoli05}. Let $\pi_{nc}$ denote the projection operator that maps a state $x = [x_c,x_{nc}]$ to the non-cyclic coordinates $x_{nc}$. In order to be able to sequence together two trajectories $x_{1}:[0,T_1] \rightarrow \R^n$ and $x_2:[0,T_2] \rightarrow \R^n$, one requires:
\begin{equation}
\pi_{nc}(x_1(T_1)) = \pi_{nc}(x_2(0)).
\end{equation}
Note, however that imposing this compatibility condition on the nominal trajectories associated with two funnels is neither necessary nor sufficient for the funnels being sequentially composable MI. Sequentially composability MI is concerned with the compatibility between funnels themselves and not the underlying nominal trajectories, which distinguishes our notion of compatibility between maneuvers from that of \cite{Frazzoli05}.

\subsection{Runtime composability}

The two notions of sequential composability we have considered so far allow us to produce new funnels from a given set of funnels by stitching them together appropriately in an offline preprocessing stage. We now introduce another notion of composability that is particularly important for reasoning about how funnels can be executed sequentially at runtime. We will refer to this notion as \emph{runtime composability}.

\quad

\begin{definition}
\label{def:runtime composability}
An ordered pair $(F_1,F_2)$ of funnels $F_1:[0,T_1] \rightarrow \mathcal{P}(\R^n)$ and $F_2:[0,T_2] \rightarrow \mathcal{P}(\R^n)$ is \emph{runtime composable} if for all $x_{out} \in F_1(T_1)$, there exists a \revision{shift} $\Psi_c$ of the state along cyclic coordinates such that $x_{out} \in \Psi_c\big{(}F_2(0)\big{)}$.
\end{definition}

\quad

\noindent In other words, for any state $x_{out}$ in the outlet of $F_1$, one can shift $F_2$ along cyclic coordinates and ensure that its inlet contains $x_{out}$. Hence, we can guarantee that it will be possible to execute the funnel $F_2$ (after appropriate shifting in the cyclic coordinates) once the funnel $F_1$ has been executed (though the \emph{particular} shift $\Psi_c$ required depends on $x_{out}$ and thus will not be known until runtime). Hence, runtime composability of funnels allows us to exploit invariances in the dynamics of the system at runtime and effectively reuse our robust motion plans in different scenarios. As a simple example, a UAV flying through a cluttered environment can reuse a funnel computed for a certain starting position by shifting the funnel so its inlet contains the UAV's current state.

\quad

\begin{remark}
\label{rem:runtime_comp_condition}
It is easy to see from Definition \ref{def:runtime composability} that two funnels $F_1$ and $F_2$ are runtime composable if and only if:
\begin{align}
\label{eq:runtime_comp_condition_1}
& \forall x_{out} = [x_c,x_{nc}] \in F_1(T_1), \ \exists x_{0,c} \ s.t. \  [x_{0,c}, x_{nc}] \in F_2(0). 
\end{align}
This condition is simply stating that we can shift the point $x_{out}$ along cyclic coordinates in such a way that it is contained in the inlet of $F_2$. This condition in turn is equivalent to:
\begin{equation}
\pi_{nc}(F_1(T_1)) \subset \pi_{nc}(F_2(0)),
\end{equation}
where as before $\pi_{nc}$ denotes the projection onto the non-cyclic coordinates of the state space.
\end{remark}

\quad

\begin{figure*}
  \begin{center}
    \includegraphics[trim = 0mm 0mm 0mm 0mm, clip, width=0.9\textwidth]{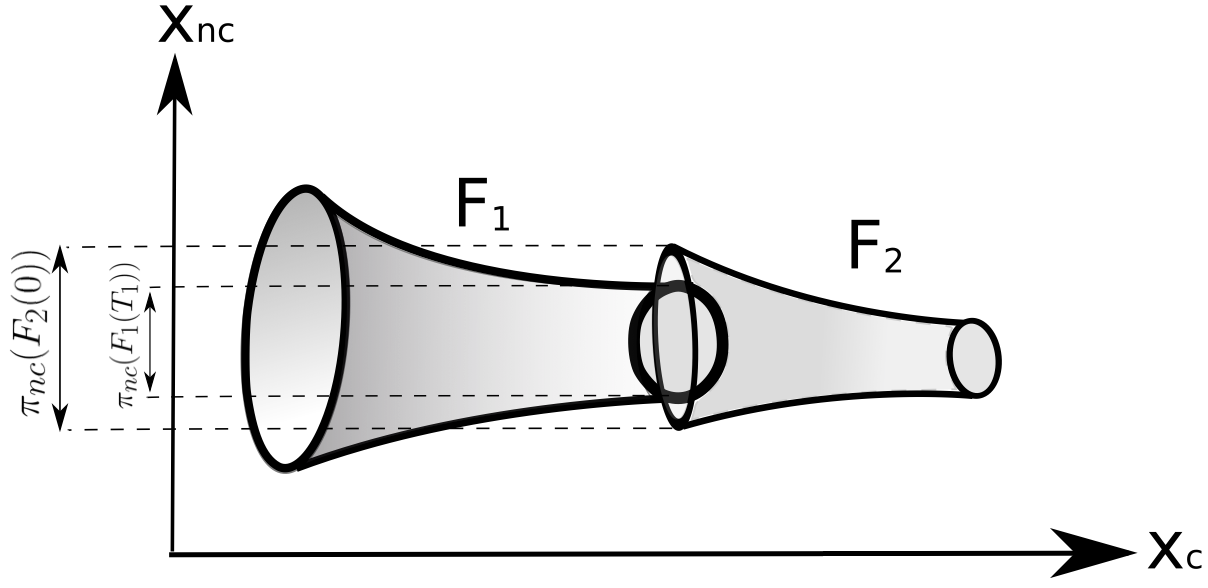}
  \end{center}
  \caption[Runtime composability of funnels]{The figure shows two funnels that are \emph{not} sequentially composable MI (since the inlet of funnel $F_2$ isn't large enough in the $x_c$ dimension). However, since $\pi_{nc}(F_1(T_1)) \subset \pi_{nc}(F_2(0))$,  they \emph{are} runtime composable.
  \label{fig:funnel_runtime_comp}}
\end{figure*}

\noindent Figure \ref{fig:funnel_runtime_comp} shows two funnels that are \emph{not} sequentially composable MI. However, since $\pi_{nc}(F_1(T_1)) \subset \pi_{nc}(F_2(0))$,  they \emph{are} runtime composable.

We end our discussion on composability of funnels by noting the following relationship between the three notions of sequential composability we have discussed.

\quad

\begin{remark}
The three notions of sequential composability we have discussed are related as follows:
\begin{equation}
\textrm{Sequential composability} \implies \textrm{Sequential composability MI} \implies \textrm{Runtime composability}.\nonumber
\end{equation}
\end{remark}

\quad

\noindent The first implication is immediate from Definitions \ref{def:seq_comp} and \ref{def:seq_comp_inv}. The second implication follows from the following reasoning:
\begin{flalign}
& F_1(T_1) \subset \Psi_c\big{(}F_2(0)\big{)} \ &\textrm{(Sequential composability MI, ref. Definition \ref{def:seq_comp_inv})} \nonumber \\
& \implies \pi_{nc} (F_1(T_1)) \subset \pi_{nc} \big{(}\Psi_c(F_2(0))\big{)} = \pi_{nc} (F_2(0)) \ \hspace{-2cm} &\textrm{(Runtime composability, ref. Remark \ref{rem:runtime_comp_condition})}. \nonumber
\end{flalign}

\subsection{Checking composability}

Given two funnels $F_1:[0,T_1] \rightarrow \mathcal{P}(\R^n)$ and $F_2:[0,T_2] \rightarrow \mathcal{P}(\R^n)$ defined as $F_1(t) = \{\bar{x} \in \R^n \ | \  V_1(t,\bar{x}) \leq \rho_1(t) \}$ and $F_2(t) = \{\bar{x} \in \R^n \ | \  V_2(t,\bar{x}) \leq \rho_2(t) \}$ for polynomials $V_1$ and $V_2$, this section describes how we can check sequential composability, sequential composability MI and runtime composability in an offline preprocessing stage. 

\subsubsection{Sequential composability}

Sequential composability of $(F_1,F_2)$ is equivalent to the following condition:
\begin{equation}
V_1(T_1,\bar{x}) \leq \rho_1(T_1) \implies V_2(0,\bar{x}) \leq \rho_2(0).
\end{equation}
We can thus apply the generalized S-procedure (described in Section \ref{sec:background_sos}) and check sequential composability using the following simple SOS program:
\begin{flalign} 
\label{eq:seq_comp_sos}
			& & \text{Find}  \hspace*{1cm} & L(\bar{x}) & \\
			& & \text{s.t.} \hspace*{1cm} & \rho_2(0) - V_2(0,\bar{x}) - L(\bar{x}) (\rho_1(T_1) - V_1(T_1,\bar{x})) \ \textrm{is SOS}, & \nonumber \\
			& & & L(\bar{x}) \ \textrm{is SOS}. & \nonumber
\end{flalign}

\subsubsection{Sequential composition MI}

In order to check sequential composability MI, we need to search for a shift $\Psi_c$ along cyclic coordinates of the state such that  $F_1(T_1) \subset \Psi_c(F_2(0))$. For the important special case in which the sets $F_1(T_1)$ and $F_2(0)$ are ellipsoids (corresponding to $V_1(T_1,\bar{x})$ and $V_2(0,\bar{x})$ being quadratic\footnote{Note that the restriction to quadratic $V_1(T_1,\bar{x})$ and $V_2(0,\bar{x})$ is a relatively mild one. We are not imposing any conditions on the degree of $V_1$ and $V_2$ at times other than the endpoints.} in $\bar{x}$), we can cast this search as a semidefinite program (SDP). Suppose the sets $F_1(T_1)$ and $F_2(0)$ are given by:
\begin{flalign}
F_1(T_1) &= \{x \in \R^n \ | \ (x - x_1(T_1))^T S_1 (x-x_1(T_1)) \leq 1\} \\
F_2(0)     &= \{x \in \R^n \ | \ (x-x_2(0))^T S_2 (x-x_2(0)) \leq 1\},
\end{flalign}
where $x_1:[0,T_1] \rightarrow \R^n$ and $x_2:[0,T_2] \rightarrow \R^n$ are the nominal trajectories around which the funnels were computed and $S_1$ and $S_2$ are positive definite matrices. We would like to search for a shift $\Delta_c \in \R^n$ (where the components of $\Delta_c$ corresponding to the non-cyclic coordinates are set to zero) such that the ellipsoid:
\begin{equation}
\Psi_c(F_2(0)) = \{x \in \R^n \ | \ [x-(x_2(0)+\Delta_c)]^T S_2 [x-(x_2(0)+\Delta_c)] \leq 1\}
\end{equation}
is contained within the ellipsoid $F_1(T_1)$. The first step in obtaining the desired SDP is to note that the set $\Psi_c(F_2(0))$ can be represented equivalently as the image of the unit ball under an affine map:
\begin{equation}
\Psi_c(F_2(0)) = \{Bu + x_2(0)+\Delta_c \ | \ \|u\|_2 \leq 1 \},
\end{equation} 
where $B = \textrm{chol}(S_2))^{-1}$. Here, $\textrm{chol}(S_2)$ is the Cholesky factorization of $S_2$ (guaranteed to exist and be invertible since $S_2$ is positive definite). Such a representation is a standard trick in semidefinite programming (see for example ~\cite[Section 8.4.2]{Boyd04a}). 

Introducing the notation  $b := -S_1x_1(T_1)$ and $c := x_1(T_1)^T S_1 x_1(T_1)-1$,
 the condition that the ellipsoid $F_1(T_1)$ is a subset of  the ellipsoid $\Psi_c(F_2(0))$ is then \emph{equivalent} to being able to find a scalar $\lambda > 0$ such that the following matrix semidefiniteness constraint holds (~\cite[Section 8.4.2]{Boyd04a}):
\begin{equation}
\label{eq:seq_comp_inv_sdp}
\left[ \begin{array}{ccc} 
-\lambda-c+b^TS_1^{-1}b & 0_{1\times n} & (x_2(0)+\Delta_c+S_1^{-1}b)^T \\
0_{n \times 1} &  \lambda I_{n \times n} & B \\
x_2(0)+\Delta_c+S_1^{-1}b & B & S_1^{-1} 
\end{array} \right] \succeq 0.
\end{equation}
Here, the matrices $I$ and $0$ represent the identity matrix and the all-zeros matrix respectively.
Since this semidefiniteness condition is linear in $\lambda$ and $\Delta_c$, the problem of searching for these decision variables subject to $\lambda > 0$ and \eqref{eq:seq_comp_inv_sdp} is a SDP. This SDP will be feasible if and only if $F_1$ and $F_2$ are sequentially composable MI.

The problem of checking sequential composability MI in the more general non-ellipsoidal case is not directly amenable to such a SDP based formulation. However, if we \emph{fix} $\Psi_c$, then the problem of checking sequential composability MI reduces to the problem of checking sequential composability. Thus, we can use the SOS program \eqref{eq:seq_comp_sos} to verify if a \emph{given} $\Psi_c$ yields the desired containment constraint $F_1(T_1) \subset \Psi_c(F_2(0))$. One natural choice is to set $\Psi_c$ such that $\pi_c(\Psi_c(x_2(0))) = \pi_c(x_1(T_1))$, where $\pi_c$ is the projection of the state onto the cyclic coordinates. Intuitively, this corresponds to shifting the funnel $F_2$ so that the start of its nominal trajectory is lined up along cyclic coordinates with the end of the nominal trajectory of $F_1$. If the SOS program is infeasible, a local search around this $\Psi_c$ could yield the desired shift.

\subsubsection{Runtime composability}

In order to check runtime composability of $F_1$ and $F_2$, we need to check the inclusion $\pi_{nc} (F_1(T_1)) \subset \pi_{nc} (F_2(0))$. For the important special case where the sets $F_1(T_1)$ and $F_2(0)$ are ellipsoids, we can compute the projections exactly. This is because the projection of an ellipsoid onto a linear subspace is also an ellipsoid (see equation \eqref{eq:ellipsoid projection} for the exact formula for the projected ellipsoid). Checking if a given ellipsoid contains another is a straightforward application of semidefinite programming ~\cite[Example B.1, Appendix B]{Boyd04a}.

For the more general case, one might hope for a SOS programming based condition for checking  $\pi_{nc} (F_1(T_1)) \subset \pi_{nc} (F_2(0))$. However, the existential quantifier inherent in the projection (see the equivalent condition \eqref{eq:runtime_comp_condition_1}) makes it challenging to formulate such SOS conditions. Nevertheless, there exist general purpose tools such as \emph{quantifier elimination} \cite{Collins98} for checking quantified polynomial formulas such as \eqref{eq:runtime_comp_condition_1}. While the worst-case complexity of doing general purpose quantifier elimination is poor, software packages such as QEPCAD \cite{Brown03} (or dReal \cite{Gao13}, which is based on a different theoretical framework) can often work well in practice for specialized problems. We note however that in the examples considered in Sections \ref{sec:examples} and \ref{sec:hardware} we will only be using funnels with ellipsoidal inlets and outlets and thus will not be concerned with this complexity.

\subsection{Funnel library}
\label{sec:funnel library definition}

A simple but useful generalization of the notions of composability introduced above can be obtained by checking the associated containment conditions at a given time $\tau_1$ rather than at time $T_1$. For example, we will say that the ordered pair of funnels $(F_1,F_2)$ is sequentially composable at time $\tau_1$ if $F_1(\tau_1) \subset F_2(0)$. We will use similar terminology for the other notions of composability. Given a collection $\mathcal{F}$ of funnels associated with a dynamical system, we will associate a directed graph $\GF$ whose vertices correspond to funnels $F \in \mathcal{F}$. Two vertices corresponding to funnels $F_i$ and $F_j$ are connected by a directed edge $(F_i,F_j)$ if and only if the ordered pair $(F_i,F_j)$ is runtime composable at some specified time $\tau_i$. We will sometimes refer to $\tau_i$ as the execution time of funnel $F_i$. 

We now introduce the primary data structure that will be used for robust real-time planning (Section \ref{sec:planning}). 

\quad

\begin{definition}
A \emph{funnel library FL} associated with a given dynamical system is a tuple FL = $(\mathcal{F}, \GF, \mathcal{C}, \{\tau_i\})$, where $\mathcal{F}$ is a set of funnels for the dynamical system, $\GF$ is the directed graph representing which funnels are runtime composable, $\mathcal{C}$ is the set of feedback controllers associated with the funnels in $\mathcal{F}$, and $ \{\tau_i\}$ is the set of execution times.
\end{definition}

\quad

\noindent \revision{We note that while the definition of funnel libraries only makes explicit reference to runtime composability of funnels, we are implicitly exploiting the notion of sequential composability (MI) since each funnel in $\F$ may be constructed from simpler funnels that have been sequentially composed.} It is also important to note that while we do not impose restrictions such as connectedness or strong connectedness on the graph $\GF$, it may be useful to impose such conditions based on the task at hand. For example, for tasks which require continuous operation (such as a UAV navigating indefinitely through a forest or a factory arm continuously placing objects onto a conveyor belt), we should require that $\GF$ be strongly connected (since if $\GF$ is not strongly connected, we are ruling out the possibility of executing certain funnels in the future). On a similar note, properties of the graph such as its \emph{diameter} or \emph{girth} may be related to the efficiency with which a certain task can be accomplished. Identifying which graph theoretic properties should be imposed on $\GF$ for different tasks is an interesting research avenue, but we do not pursue this in the present work.

%% file: planning.tex
\section{Real-time Planning with Funnels}
\label{sec:planning}

Given a funnel library FL = $(\mathcal{F}, \GF, \mathcal{C}, \{\tau_i\})$ computed offline, we can proceed to use it for robust real-time planning in previously unseen environments. The robot's task specification may be in terms of a goal region that must be reached (as in the case of a manipulator arm grasping an object), or in terms of a nominal direction the robot should move in while avoiding obstacles (as in the case of a UAV flying through a forest or a legged robot walking over rough terrain). For the sake of concreteness, we adopt the latter task specification although one can adapt the contents of this section to the former specification. We further assume that the robot is provided with regions in the configuration space that obstacles are guaranteed to lie in and that the robot's sensors only provide this information up to a finite spatial horizon around the robot. Our task is to choose funnels from our library in a way that avoids obstacles while moving forward in the nominal direction.

The key step in our real-time planning approach is the selection of a funnel from the funnel library that doesn't intersect any obstacles in the environment. This selection process is sketched in the ReplanFunnel algorithm (Algorithm \ref{a:replan_funnel}). Given the current state $x$ of the robot and the locations and geometry of obstacles $\OO$ in the environment, the algorithm searches through the funnels in the library that are runtime composable with the previous funnel that was executed. \revision{We assume that for any given funnel, the set of funnels in the library that are runtime composable with it (easily computed given the graph $\GF$) is (totally) ordered in a preference list. This ordering may be prespecified offline or generated online based on the current state of the robot. For example, this ordering can be based on aggressiveness of the maneuvers for a UAV (less aggressive maneuvers are given preference) or likely progress towards the goal.} For each funnel $F_i$, the algorithm tries to find a shift $\Psi_c$ along cyclic coordinates of the system such that the shifted funnel $\Psi_c(F_i)$ satisfies two properties: (i) the current state $x$ is contained in the inlet of the shifted funnel, (ii) the projection of the shifted funnel onto the coordinates of the state space corresponding to the configuration space\footnote{Note that these projections can be computed in the offline computation stage.} doesn't intersect any obstacles in $\OO$. If we are able to find such a shift, we are \emph{guaranteed} that the system will remain collision-free when the funnel is executed despite the uncertainties and disturbances that the system is subjected to.

The simplest way to try to choose $\Psi_c$ is to set it such that the nominal trajectory $x_i$ associated with the funnel lines up with the current state in the cyclic coordinates (i.e., using the notation of Section \ref{sec:funnel libraries}, $\pi_c(\Psi_c(x_i(0))) = \pi_c(x)$). Intuitively, this corresponds to shifting the funnel so that it is executed from the current location of the robot. We can then use standard collision-checking libraries such as the Bullet Collision Detection \& Physics Library \cite{Coumans06} to check if $\Psi_c(F_i)$ (projected onto the configuration space) intersects any obstacles. We will discuss more sophisticated ways of finding $\Psi_c$ in Section \ref{sec:shifting funnels}.

If the search for a collision-free funnel in Algorithm \ref{a:replan_funnel} doesn't succeed, \revision{we assume the existence of a failsafe maneuver ($F_{\textrm{failsafe}}$ in Algorithm \ref{a:replan_funnel}) that can be executed in order to keep the robot safe (we assume that the failsafe is verified to be executable from any state the robot may find itself in when replanning).} For a ground vehicle, this could entail coming to a stop while for a quadrotor or fixed-wing airplane this may involve transitioning to a hover or propellor-hang mode. In certain cases, it is possible to derive geometric conditions on the environment that the robot will operate in (e.g., constraints on obstacle size and gaps between obstacles) that \emph{guarantee} that a collision-free funnel will always be found by Algorithm \ref{a:replan_funnel} if the environment satisfies these conditions. We will see an example of this in Section \ref{sec:quadrotor} for a quadrotor system navigating through a forest of polytopic obstacles.


\begin{algorithm}[t!]
  \caption{ReplanFunnel}
  \label{a:replan_funnel}
  \begin{algorithmic}[1]
   \STATE Inputs: $x$ (current state of system), $\OO$ (reported obstacles in environment), previousFunnel (previous funnel that was executed)
    \FOR{$i=1,\dots,\#(\F)$ such that $($previousFunnel$, F_i) \in \mathcal{G}(\F)$}
    		\STATE  success $\Leftarrow$ Find a shift $\Psi_c$ along cyclic coordinates such that $x$ is contained in the inlet of $\Psi_c(F_i)$ and $\Psi_c(F_i)$ is collision-free w.r.t. $\OO$
  	  \IF{success}
   	 	 \RETURN $\Psi_c(F_i)$
 	   \ENDIF
    \ENDFOR
    \RETURN $F_{\textrm{failsafe}}$
      \end{algorithmic}\vspace{10pt}
\end{algorithm}

Algorithm \ref{a:online_planning} provides a sketch of the real-time planning and control loop, which applies the ReplanFunnel algorithm in a receding-horizon manner. At every control cycle, the robot's sensors provide it with a state estimate and report the locations and geometry of the set of obstacles $\OO$ in the sensor horizon. The algorithm triggers a replanning of funnels if any of the following three criteria are met: (i) if the system has executed the current funnel $F_i$ for the associated execution time $\tau_i$, (ii) if the current state of the system is no longer in the funnel being executed, or (iii) if the current funnel being executed is no longer collision-free. In principle, (ii) should not happen. However, in practice this can happen if the system received a disturbance that was larger than the ones taken into account for the funnel computations. Option (iii) can happen if the robot's sensors report new obstacles that were previously unseen.

\begin{algorithm}[h!]
  \caption{Real-time planning loop}
  \label{a:online_planning}
  \begin{algorithmic}[1]
    \STATE $x \Leftarrow$ Initialize current state of the robot
    \STATE $\OO \Leftarrow$ Initialize obstacles in sensor horizon
    \STATE $F_i \in \F \Leftarrow$ Initialize current planned funnel
    \FOR{$t=0,\dots$}
   	\STATE $x \Leftarrow$ Update current state of robot
    	\STATE $\OO \Leftarrow$ Update obstacles in sensor horizon
	\STATE replan $\Leftarrow$ Check if we have finished executing current funnel $F_i$ for the associated execution time $\tau_i$
	\STATE insideFunnel $\Leftarrow$ Check if current state is still inside the current funnel $F_i$ being executed
	\STATE collisionFree $\Leftarrow$ Check if current funnel $F_i$ is still collision-free with $\OO$
	\IF {replan \OR $\neg$insideFunnel \OR $\neg$collisionFree}
		\STATE $F_i \Leftarrow$ ReplanFunnels($x, \OO$, $F_i$)
	\ELSE
		\STATE apply feedback control input $u$ associated with current funnel $F_i$ 
	\ENDIF	
    \ENDFOR 
  \end{algorithmic}
\end{algorithm}

\vspace{40pt}
\subsection{Shifting funnels at runtime}
\label{sec:shifting funnels}

The main step in Algorithm \ref{a:replan_funnel} is the search for a shift $\Psi_c$ along cyclic coordinates such that the shifted funnel $\Psi_c(F_i)$ will be collision-free with respect to the obstacles in $\OO$ while containing the current state $x$ in its inlet:

\begin{flalign} 
\label{eq:find_shift}
			& & \text{Find}  \hspace*{1cm} & \Psi_c & \\
			& & \text{s.t.} \hspace*{1cm} & x \in \Psi_c(F_i(0)), & \label{eq:containment constraint} \\
			& & & \pi_{conf}(\Psi_c(F_i)) \cap \OO = \O. & \nonumber
\end{flalign}
Here, $\pi_{conf}$ is the projection onto the configuration space variables of the state space. This optimization problem is non-convex in general since the free-space of the environment is non-convex. However, the number of decision variables is very small. In particular, if we parameterize the shift $\Psi_c$ by a vector $\Delta_c \in \R^n$ (where the coordinates of $\Delta_c$ corresponding to the non-cyclic coordinates are set to zero) such that $\Psi_c(x) = x + \Delta_c$, the number of decision variables is equal to the number of cyclic coordinates of the system. 

We can thus apply general-purpose nonlinear optimization tools such as gradient-based methods to solve this problem. In particular, the first constraint is equivalent to checking that the value of the Lyapunov function $V(0,\bar{x}) = V(0,x - (x_i(0) + \Delta_c))$ at time 0 is less than or equal to $\rho(0)$ (recall that the funnel is described as the $\rho(t)$ sub-level set of the function $V(t,\bar{x})$). As in Section \ref{sec:funnels}, $x_i$ here is the nominal trajectory corresponding to the funnel $F_i$. The second constraint can be evaluated using off-the-shelf collision-checking libraries. 

\begin{figure*}[h!]
  \begin{center}
    \includegraphics[trim = 0mm 0mm 0mm 0mm, clip, width=0.45\textwidth]{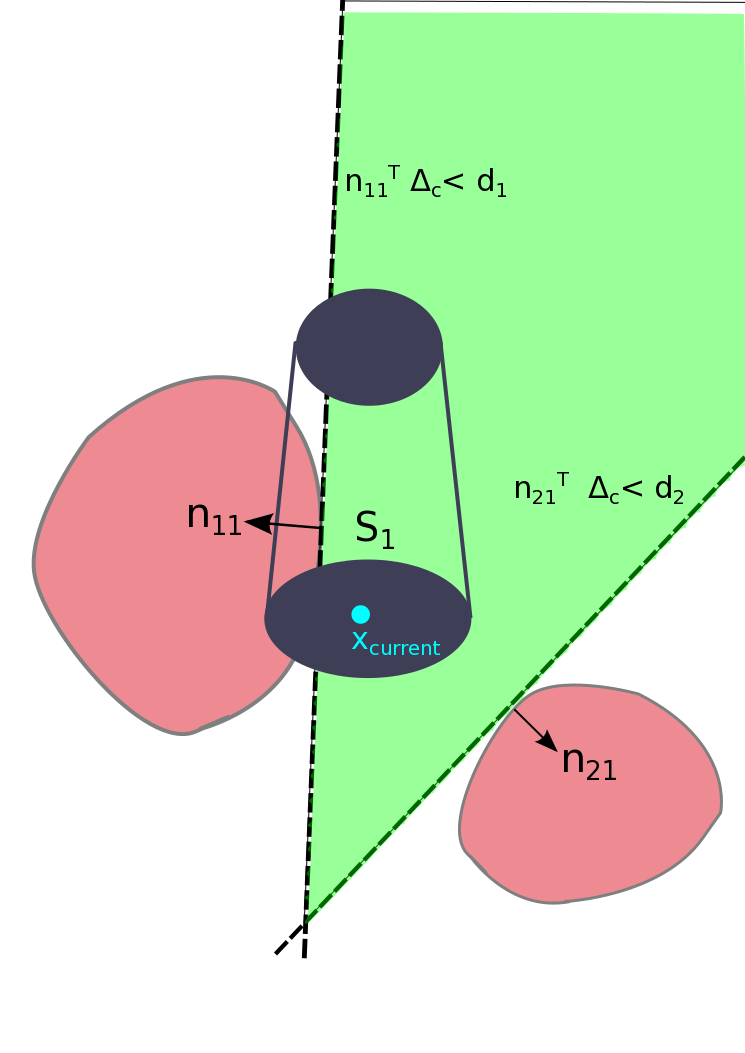}
  \end{center}
  \caption[Shifting funnels out of collision by exploiting invariances in dynamics]{The two red regions are obstacles. \revision{The shaded green region represents the set in which the funnel segment must be entirely contained in order to satisfy the linear constraints given by the collision normals and collision distances.}
  \label{fig:qcqp shifting}}
  \vspace{-5pt}
\end{figure*}

Despite the small number of decision variables, for some applications general-purpose nonlinear optimization may be too slow to run in real-time. For the important special case where the cyclic coordinates form a subset of the configuration space variables (e.g., for a UAV), we propose another approach that allows us to search over a restricted family of shifts $\Psi_c$ but has the advantage of being posed using \emph{convex} quadratic constraints. The first observation is that the containment constraint \eqref{eq:containment constraint} is a convex quadratic constraint for the case where the inlet of the funnel is an ellipsoid (i.e., the function $V(0,\bar{x})$ is a positive definite quadratic). If the inlet is not an ellipsoid, we can find an ellipsoidal inner approximation by solving a simple SOS program offline.

Next, we seek to find a set of convex constraints that will guarantee non-collision of the funnel. We will assume that $\pi_{conf}(F_i)$ is represented as a union of convex segments $\mathcal{S}_k$ and that the obstacles in the environment are also convex (we assume that non-convex obstacles have been decomposed into convex segments). For each funnel segment, we can find collision normals $n_{jk}$ and collision distances $d_{jk}$ to each obstacle $o_j \in \OO$ (these are easily extracted from a collision-checking software). Figure \ref{fig:qcqp shifting} provides an illustration of this for a single convex segment $\mathcal{S}_1$. The two regions colored in red are obstacles. The collision normals are $n_{11}$ and $n_{21}$. By definition, collision normals provide us with constraints such that any shift $\Delta_c$ of the segment $\mathcal{S}_1$ satisfying the linear constraint $n_{jk}^T \Delta_c < d_{jk}$ will be collision-free. The region corresponding to this is shaded in green in the figure.

The constraints described above (containment of the current state in an ellipsoid and the linear constraints given by the collision normals) form a special case of convex Quadratically Constrained Quadratic Programs (QCQPs), for which there exist very mature software packages. For our examples in Section \ref{sec:examples}, we use the FORCES Pro package \cite{Domahidi14}. The package generates solver code tailored to the specific optimization problem at hand and is faster in our experience than using general-purpose convex QCQP solvers.

\subsection{Global Planning}

Algorithm \ref{a:online_planning} employed a receding horizon strategy for real-time planning. For tasks such as robot navigation through previously unmapped environments, such a replanning-based strategy is unavoidable since the robot's sensors will only report obstacles in the environment in some finite sensor horizon around the robot. Thus, the robot will need to replan as it makes progress through the environment and more obstacles are reported by its sensors. However, in certain scenarios where the robot has access to a larger portion of the environment, a planning strategy that is more \emph{global} in nature may be appropriate. In general, the funnel primitives provide a discrete action space which can be searched by any heuristic planner - the primary considerations here are the additional constraint of containment of the current state in the inlet and the moderately more significant cost of collision checking. Here we briefly mention a few ways in which global planning may be achieved.

\quad

\noindent {\bf{Grid-based planners:}} One can build on existing trajectory library-based planning approaches to quickly find collision-free sequences of funnels that minimize a certain performance criterion (e.g., distance traversed by the nominal trajectories in the sequence). For example, \cite{Stolle06} uses a variant of the grid-based planning algorithm $A^\star$ to search through possible sequences of trajectories in a library. The additional cost of extending such an approach for planning with funnels is the cost that comes from pruning sequences of funnels that are in collision with obstacles in the environment.

\quad

\noindent {\bf{Maneuver automata:}} The Maneuver Automaton \cite{Frazzoli05} provides an alternative trajectory library-based approach that exploits continuous symmetries (such as shift invariance for ground vehicles or UAVs) to achieve efficient planning. The approach relies on having a number of ``trim trajectories" and maneuvers that transition between them. One can then efficiently plan sequences of trims and maneuvers that start and end at prescribed points in space by solving a set of equations that has the same structure as an inverse kinematics problem. In order to extend this approach for real-time planning with funnels, one needs to (potentially approximately) solve the inverse kinematics problem subject to the constraint that the funnel sequence is collision-free. 

\quad

\noindent {\bf{Planning backwards:}} The real-time planning approaches we have discussed so far all plan forward in time. However, in cases where there is a well-defined goal set that one needs to reach, it may be more efficient to plan backwards. This kind of planning is a direct analog of the \emph{preimage backchaining} approach \cite{Lozano-Perez84} in the motion planning literature. One way to perform this backwards search would be to employ a randomized strategy similar to the Rapidly-exploring Randomized Tree (RRT). In particular, we can grow a tree of funnels backwards from the goal set by using the funnels in our funnel library as primitives for the extension operator for the RRT. One can again exploit invariances in the dynamics when performing this extension operation by randomly sampling shifts/rotations of funnels (while still maintaining sequential composability and non-collision of funnels). The termination criterion for this RRT-like algorithm is the containment of the current state of the robot in the tree of funnels. This is very similar to the LQR-Trees approach \cite{Tedrake10}, with the following differences: (i) one would only use a pre-computed library of funnels in the tree rather than computing new funnels (which would be computationally infeasible for real-time implementation), and (ii) the extension operator for the tree \revision{would take into account possible collisions of funnels with obstacles in the environment.}

%% file: examples.tex
\section{Examples}
\label{sec:examples}

In this section we apply the techniques presented in this paper on two simulation examples. We will consider a hardware example in Section \ref{sec:hardware}.  \revision{An example implementation of our approach for a model of the quadrotor system (Section \ref{sec:quadrotor}) is available as part of the Drake toolbox \cite{Drake}.} The computations in this section were performed on a 3.4 GHz desktop computer with 16 GB RAM and 4 cores.


\subsection{Ground Vehicle Model}
\label{sec:ground vehicle}

\begin{figure*}
  \begin{center}
    \includegraphics[trim = 0mm 0mm 0mm 0mm, clip, width=0.4\textwidth]{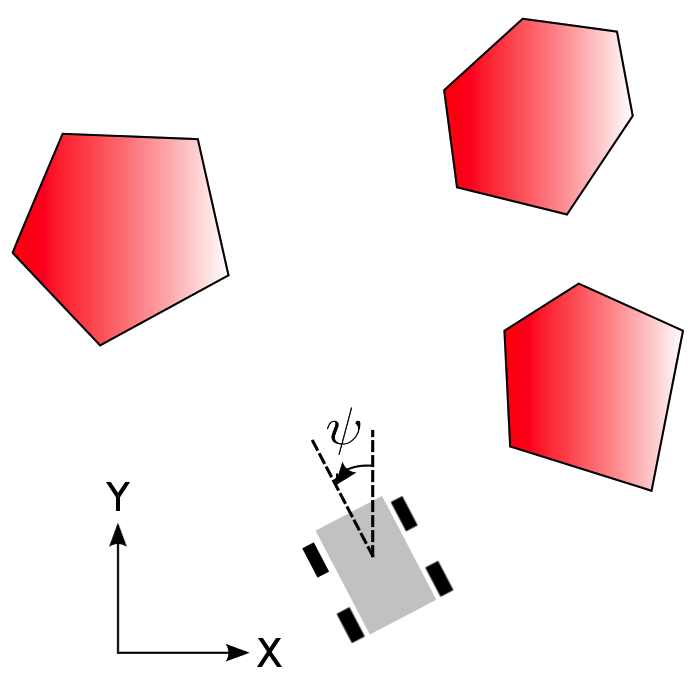}
  \end{center}
  \caption{Illustration of the ground vehicle model.
  \label{fig:dubins_cartoon}}
\end{figure*}

As our first example we consider a ground vehicle model based on the Dubins car \cite{Dubins57} navigating an environment of polytopic obstacles. A pictorial depiction of the model is provided in Figure \ref{fig:dubins_cartoon}. The vehicle is constrained to move at a fixed forward speed and can control the second derivative of its yaw angle $\psi$. We introduce uncertainty into the model by assuming that the speed of the vehicle is only known to be within a bounded range and is potentially time-varying. The full non-linear dynamics of the system are then given by:
\begin{equation}
\bf{x} = \left[ \begin{array}{c}
x \\
y \\
\psi \\
\dot \psi \end{array} \right],
\qquad 
\dot{\bf{x}} = \left[ \begin{array}{c} 
-v(t) \sin \psi \\
v(t) \cos \psi \\
\dot{\psi} \\
u \end{array} \right]
\label{eq:dubins_dynamics}
\end{equation}
with the speed of the plane $v(t) \in [9.0,11.0]$ m/s. The control input is bounded in the range $[-1000,1000]$ rad/s$^2$.

\begin{figure}
\centering

\subfigure[]{\includegraphics[trim = 135mm 20mm 130mm 30mm, clip, width=0.48\columnwidth]{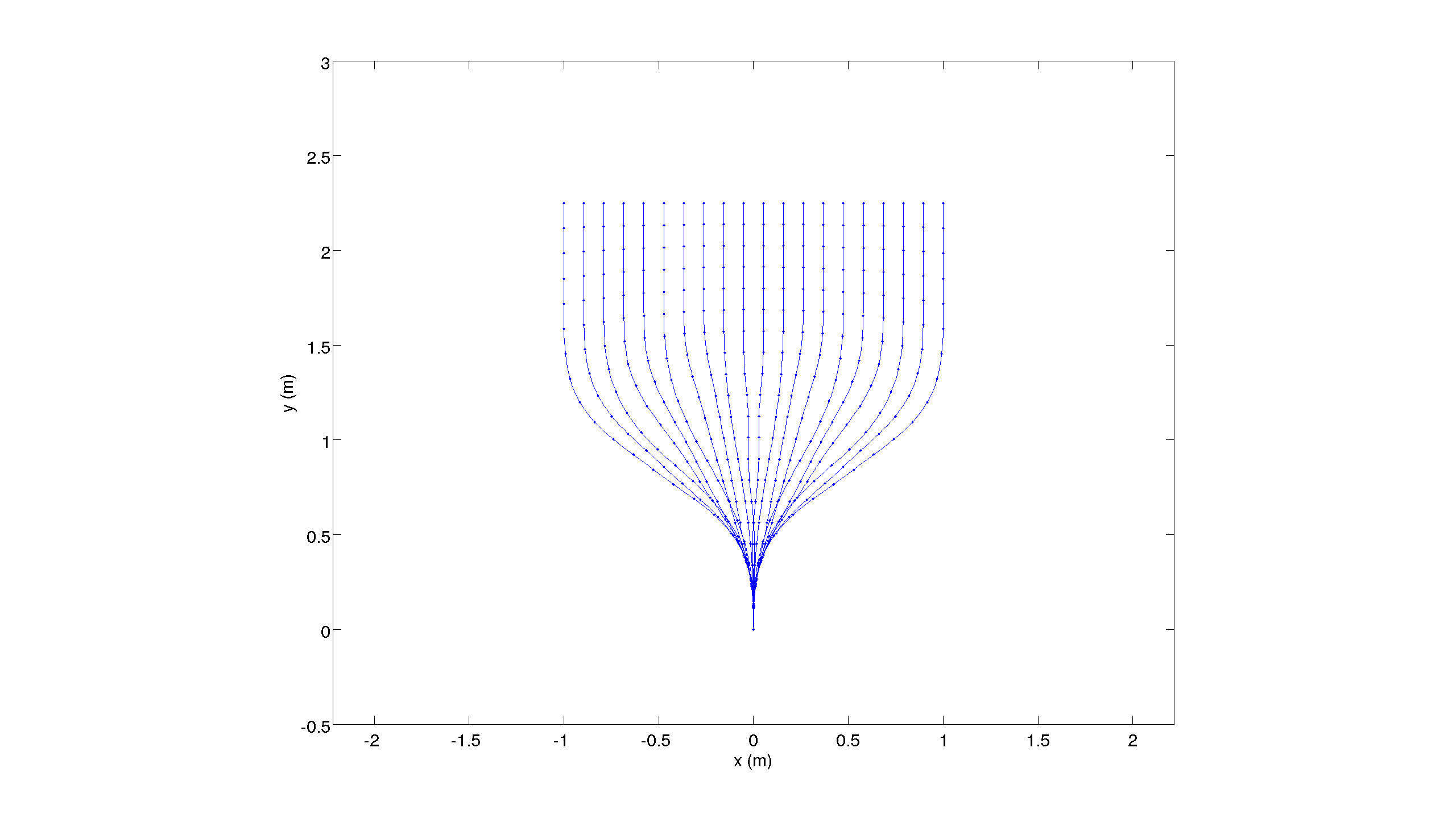} \label{fig:dubins_traj_library}} 
\subfigure[]{\includegraphics[trim = 135mm 20mm 130mm 30mm, clip, width=0.48\columnwidth]{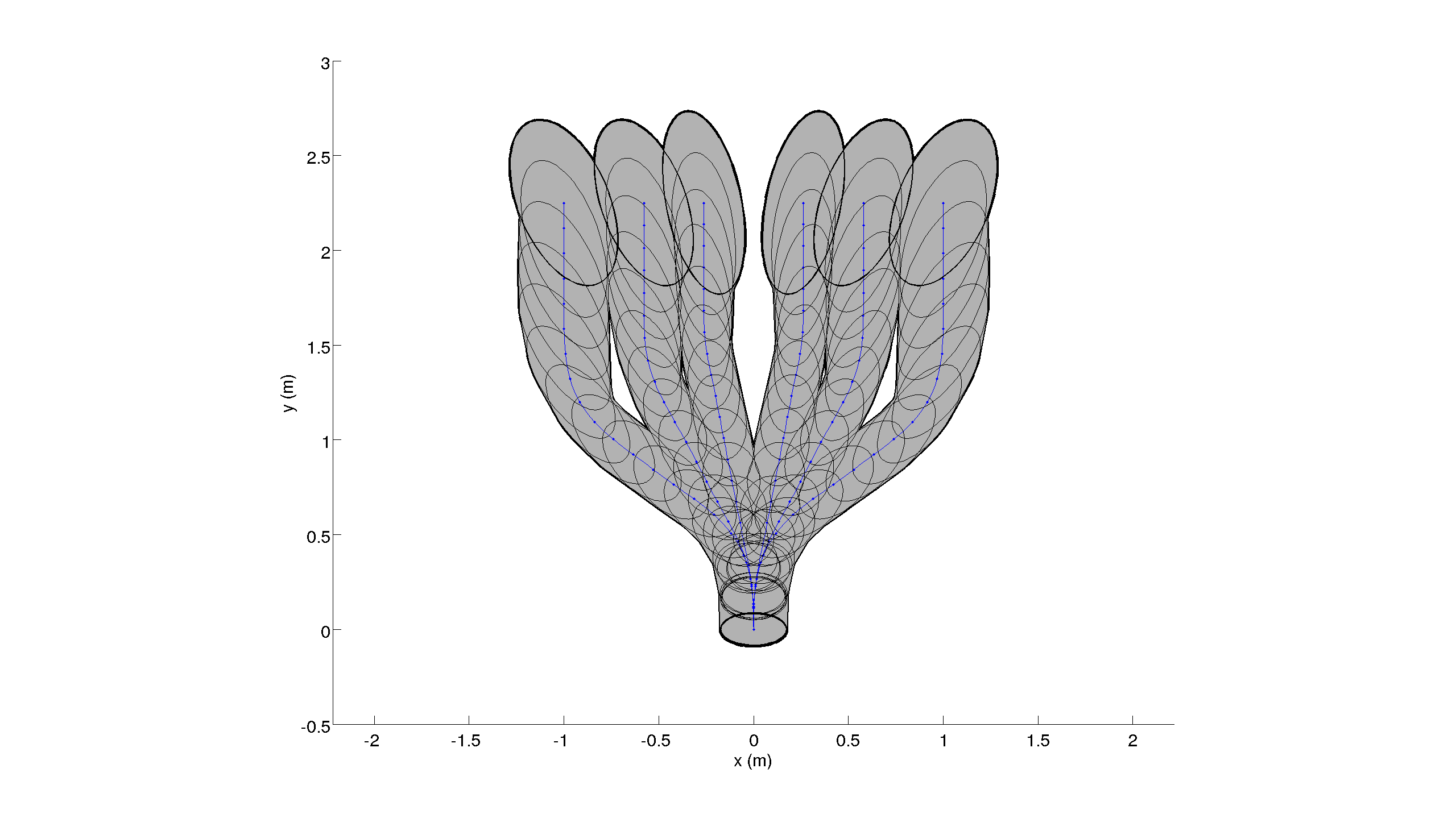} \label{fig:dubins_funnel_library}}    

\caption[Trajectory and funnel libraries for ground vehicle model]{The plot on the left shows the trajectory library for the ground vehicle model. The plot on the right shows a selection of funnels from the funnel library projected onto the $x-y$ plane. 
}
\end{figure}

The trajectory library, $\T$, computed for the ground vehicle consists of 20 trajectories and is shown in Figure \ref{fig:dubins_traj_library}. The trajectories $x_i(t):  [0,T_i] \mapsto \RR^4$ and the corresponding nominal open-loop control inputs were obtained via the direct collocation trajectory optimization method \cite{Betts01} for the vehicle dynamics in  \eqref{eq:dubins_dynamics} with $v(t) = 10$ m/s.  The initial state $x_i(0)$ was constrained to be $[0,0,0,0]$  and the final state $x_i(T_i)$ was varied in the x direction while keeping the y component fixed. We locally minimized a cost of the form:
 $$J = \int^{T_i}_{0}[1 + u_0(t)^TR(t)u_0(t)] \,dt $$
where $R$ is a positive-definite matrix. We constrained the nominal control input to be in the range $[-500,500]$ rad/s$^2$ to ensure that feedback controllers computed around the nominal trajectories do not immediately saturate. 

For each $x_i(t)$ in $\T$ we obtain controllers and funnels using the method described in Section \ref{sec:funnels}. In order to obtain polynomial dynamics, we computed a (time-varying) degree 3 Taylor expansion of the dynamics of the system around the nominal trajectory. We note that with the right change of coordinates, one can express the dynamics of this system directly as a polynomial. In particular, we can introduce new indeterminates $s$ and $c$ for $\sin(\psi)$ and $\cos(\psi)$, and impose the constraint that $s^2 + c^2 = 1$ (this equality constraint is easily imposed in the sums-of-squares programming framework). However, this increases the dimensionality of the state space and in practice we find that the time-varying Taylor approximation accurately captures the nonlinearities of the system. 

The approach from Section \ref{sec:funnels_controller} along with the time-sampled approximation described in Section \ref{sec:funnels_time_sampling} (with 15 time samples) was used to synthesize a (time-varying) linear feedback controller around each trajectory. The methods described in Section \ref{sec:funnels_uncertainty} and \ref{sec:funnels_saturations} were used to take into account the parametric uncertainty and input saturations that the system is subject to. As described in Section \ref{sec:initialization}, we used a time-varying LQR controller to initialize the funnel computations. The computation time for each funnel was approximately 5-10 minutes. A subset of the funnels is shown in Figure \ref{fig:dubins_funnel_library}. Note that the four-dimensional funnels have been projected down to the $x-y$ dimensions for the purpose of visualization; \revision{the true funnels take into account variations in all four state dimensions}. The directed graph $\GF$ that encodes real-time composability between funnels (ref. Section \ref{sec:funnel library definition}) is fully connected.

\begin{figure*}[t!]
  \begin{center}
    \includegraphics[trim = 0mm 0mm 0mm 0mm, clip, width=0.9\textwidth]{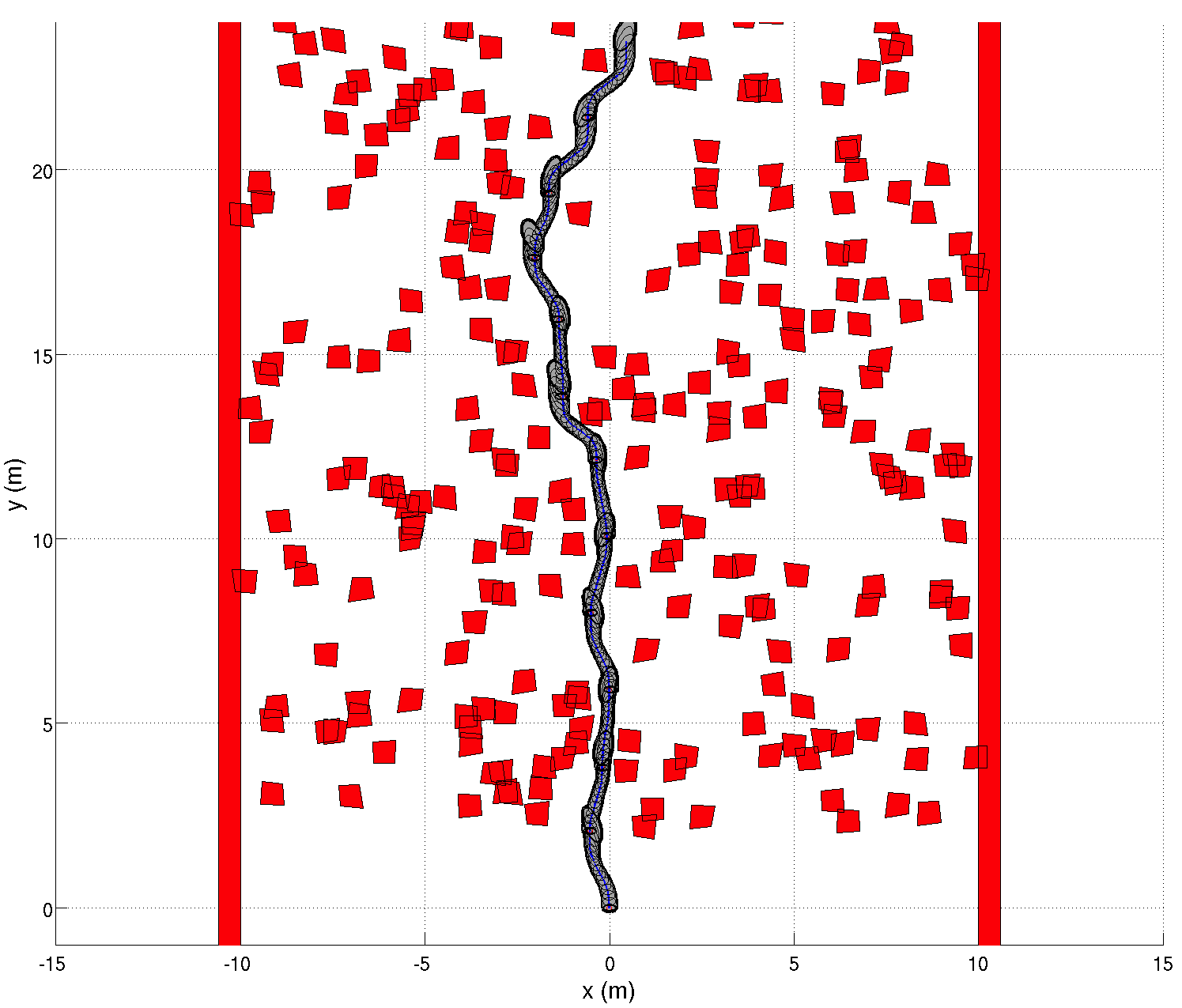}
  \end{center}
  \caption[Ground vehicle model navigating through a cluttered environment using funnel library]{This plot shows the funnels that were executed in order for the ground vehicle model to traverse a randomly generated obstacle environment using the funnel library based real-time planning approach.
  \label{fig:dubins_example_run}}
\end{figure*}

The resulting funnel library was employed by Algorithm \ref{a:online_planning} for planning in real-time through environments with randomly placed obstacles. Figure \ref{fig:dubins_example_run} shows the funnels that were executed in order to traverse a representative environment. The obstacle positions were randomly generated from a spatial Poisson process (with a density/rate parameter of 0.6 obstacles per $m^2$). Two further ``barrier" obstacles were placed on the sides of the environment to prevent the vehicle from leaving the region containing obstacles.
The planner was provided with a sensor horizon of 3m in the $y$ direction (forward) and $\pm 2$m in the $x$ direction (side-to-side) relative to the position of the vehicle. Only obstacles in this sensor window were reported to the planner at every instant in time. The execution times ${\tau_i}$ for each funnel were set such that replanning occurred once $80\%$ of the funnel was executed. 
The parametric uncertainty in the speed of the vehicle was taken into account in our simulation by randomly choosing a speed $v(t) \in \{9.0,11.0\}$ m/s after every execution time period has elapsed.
The real-time planner employs the QCQP-based algorithm from Section \ref{sec:shifting funnels} for shifting funnels in the $x-y$ directions (we did not exploit invariances in the yaw dimension of the state space in order to ensure that the vehicle moves in the forward direction and doesn't veer off to the sides). We use the FORCES Pro solver \cite{Domahidi14} for our QCQP problems. This resulted in our implementation of the real-time planner running at approximately $50$ Hz \revision{in the worst-case (i.e., when all the funnels in the library have to be evaluated for collisions).}

We performed extensive simulation experiments to compare our funnel library based robust planning approach with a more traditional trajectory library based method.
In order to facilitate a meaningful comparison, we used the underlying trajectory library corresponding to our funnel library. The trajectory-based planner employs essentially the same outer-loop as our funnel-based approach (Algorithm \ref{a:online_planning}). The key difference is that the planner chooses which maneuver to execute by evaluating which trajectory has the maximal clearance from the obstacles (as measured by Euclidean distance):
\begin{equation}
\label{eq:clearance}
\underset{i}{\max} \ \underset{t,j}{\min} \ \textrm{dist}(x_i(t), o_j)
\end{equation}
where $x_i(t) \in \T$ is a trajectory in the library and $o_j \in \OO$ is an obstacle in the environment. Once a maneuver is chosen based on this metric, a time-varying LQR feedback controller computed along this trajectory is applied (the same LQR controller that is used to initialize the funnel computations).

\begin{figure*}[t!]
  \begin{center}
    \includegraphics[trim = 0mm 0mm 0mm 0mm, clip, width=0.9\textwidth]{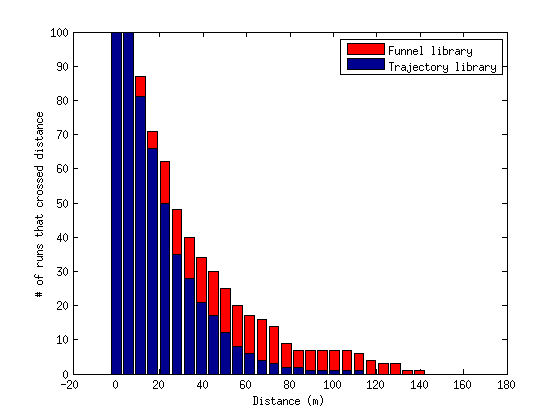}
  \end{center}
  \caption[Comparison of the funnel library and trajectory library based planning approaches for the ground vehicle model]{A bar plot comparing the performance of the funnel library approach with one based on a trajectory library.  \label{fig:bar_plot_dists}}
\end{figure*}

To compare the two planners, we generated 100 obstacle environments randomly as described previously. For each environment, we ran the different planning algorithms until there was a collision of the vehicle with an obstacle. The distance in the $y$ direction at the time of collision was recorded for each run. Figure \ref{fig:bar_plot_dists} compares the performance of the different approaches. For each distance on the x-axis of the plot the height of the bar indicates the number of runs for which the vehicle traveled beyond this distance. As is evident from the plot, our funnel library based approach provides a significant advantage over the trajectory-based method.

\begin{figure*}[h!]
  \begin{center}
    \includegraphics[trim = 220mm 30mm 200mm 0mm, clip, width=0.5\textwidth]{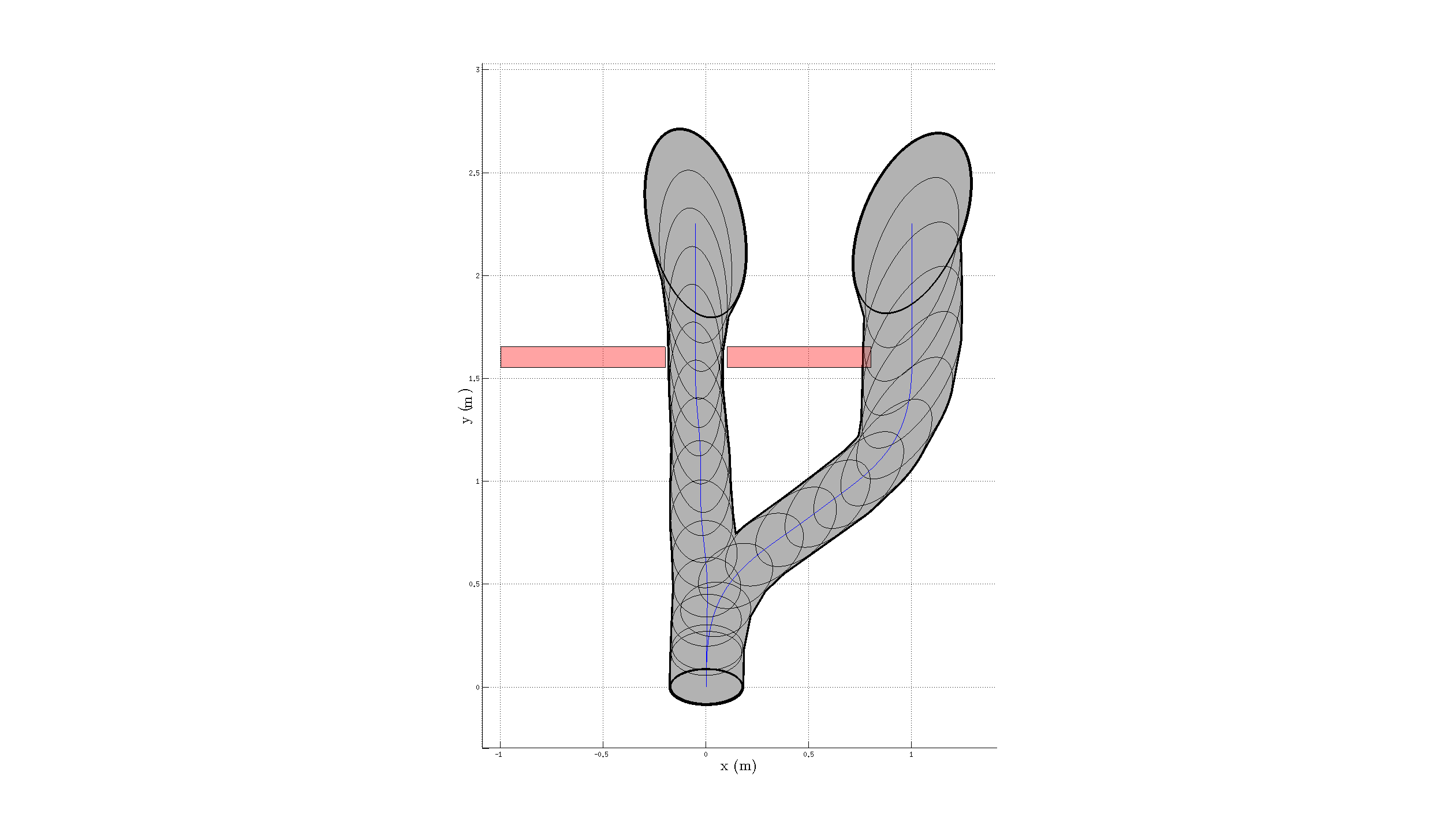}
  \end{center}
  \caption[The utility of making decisions using funnel libraries]{This figure shows the utility of explicitly taking uncertainty into account while planning.
The intuitively more “risky” strategy of flying in between two closely spaced obstacles is guaranteed to be safe, while the path that avoids going in between obstacles is less robust to uncertainty
and could lead to a collision.  \label{fig:dubins_traj_vs_funnel}}
\end{figure*}

This advantage can be partially understood by considering a specific example. In particular, Figure \ref{fig:dubins_traj_vs_funnel} demonstrates the utility of explicitly taking into account uncertainty during the planning process. There are two obstacles in front of the vehicle. The two options available to the plane are to fly straight in between the obstacles or to maneuver aggressively to the right and attempt to go around them. If the motion planner didn't take uncertainty into account and simply chose to maximize the clearance in terms of Euclidean distance from the nominal trajectory to the obstacles (see equation \eqref{eq:clearance}), it would choose the trajectory that goes right around the obstacles. However, taking the funnels into account leads to a different decision: going straight in between the obstacles is guaranteed to be safe even though the distance to the obstacles is smaller. In contrast, the maneuver that avoids going in between obstacles is less robust to uncertainty and could lead to a collision. The utility of safety guarantees in the form of funnels is especially important when the margins for error are small and making the wrong decision can lead to disastrous consequences.

\subsection{Quadrotor Model}
\label{sec:quadrotor}

\begin{figure}
\centering

\includegraphics[trim = 0mm 0mm 0mm 0mm, clip, width=0.7\columnwidth]{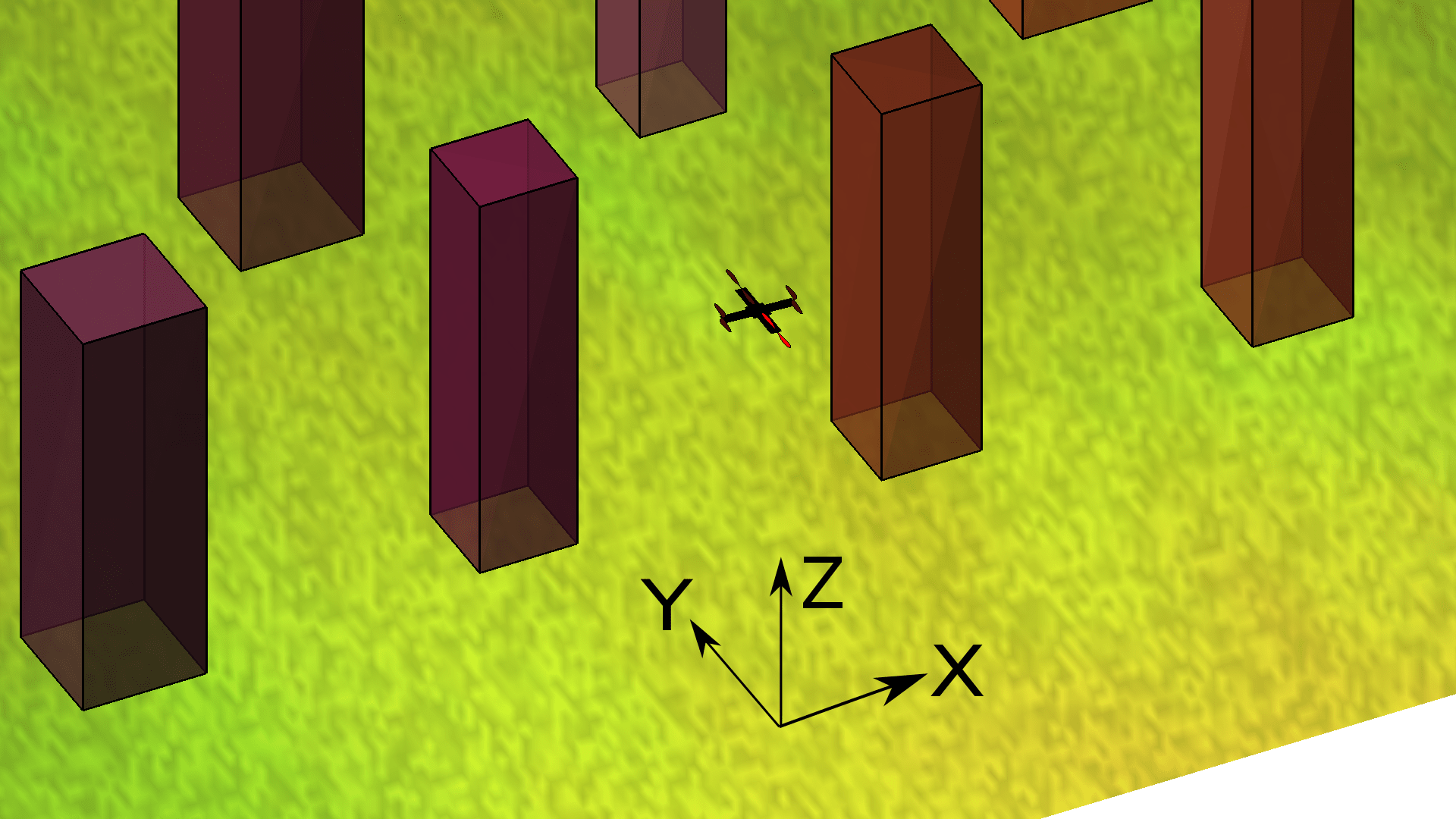}

\caption[Visualization of the quadrotor system]{Visualization of the quadrotor system. \label{fig:quad_visualization}}
\end{figure}

The next example we consider is a model of a quadrotor system navigating through a forest of polygonal obstacles. A visualization of the system is provided in Figure \ref{fig:quad_visualization}. The goal of this example is to demonstrate that we can derive simple geometric conditions on the environment that guarantee collision-free flight. In other words if the environment satisfies these conditions, the Algorithm \ref{a:replan_funnel} presented in Section \ref{sec:planning} will always succeed in finding a collision-free funnel from the library and the quadrotor will fly forever through the environment with no collisions.

The quadrotor model has a 12 dimensional state space consisting of the x-y-z position of the centre of mass, the roll-pitch-yaw of the body, and the time derivatives of these configuration space variables. The dynamics model we use is identical to the one presented in \cite{Mellinger10} with an additional uncertainty term in the form of a ``cross wind". We modeled this with a bounded uncertainty term on the acceleration of the $x$ position: $\ddot{x} = \ddot{x}_{nominal} + \Delta$, with $\Delta \in [-0.1,0.1] \ m/s^2$.

\begin{figure}
\centering

\subfigure[]{\includegraphics[trim = 0mm 0mm 0mm 0mm, clip, width=0.48\columnwidth]{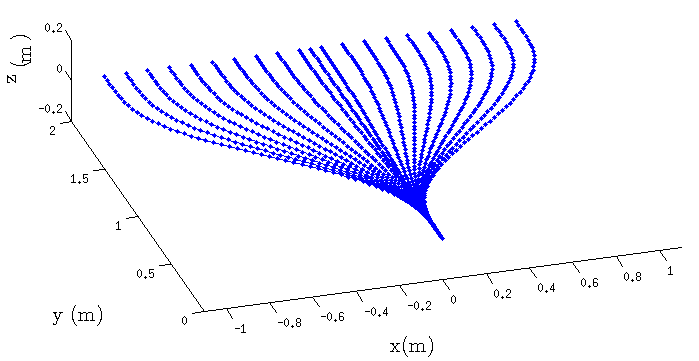} \label{fig:quad_traj_library}} 
\subfigure[]{\includegraphics[trim = 0mm 0mm 0mm 0mm, clip, width=0.48\columnwidth]{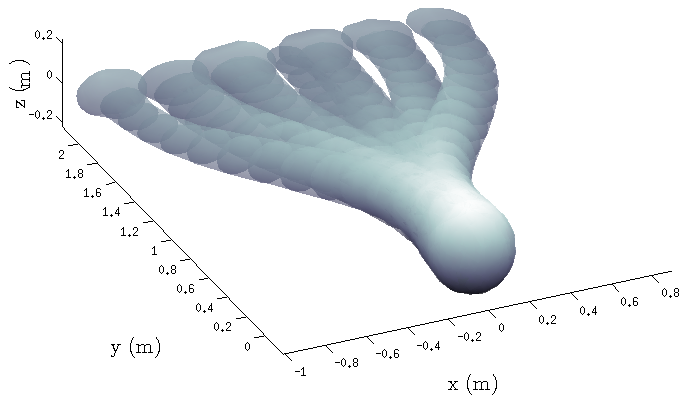} \label{fig:quad_funnel_library}}    

\caption[Trajectory and funnel libraries for quadrotor model]{The plot on the left shows the trajectory library for the quadrotor model. The plot on the right shows a selection of funnels from the funnel library projected onto the $x-y-z$ space.
}
\end{figure}

Figure \ref{fig:quad_traj_library} plots the trajectory library we use. The funnel library consists of 20 maneuvers and was created in a manner similar to the ground vehicle example. In particular, the trajectories $x_i(t):  [0,T_i] \mapsto \RR^{12}$ and the corresponding nominal open-loop control inputs were obtained via the direct collocation trajectory optimization method \cite{Betts01}. The initial state $x_i(0)$ was constrained to have a forwards speed of $2 \ m/s$ (i.e, $x_i(0) = [0,0,0,0,0,0,0,2,0,0,0,0]$). \revision{The final state $x_i(T_i)$ was varied in the x-direction while keeping the y-component fixed to $2$m (all other components of the final state were chosen to be the same as the initial state)}. We locally minimized a cost of the form:
 $$J = \int^{T_i}_{0}[1 + u_0(t)^TR(t)u_0(t)] \,dt $$
where $R$ is a positive-definite matrix. In addition to the 20 maneuvers, the library also consists of a ``trim" trajectory corresponding to the quadrotor flying forward at a constant speed of $2 \ m/s$.

For each $x_i(t)$ in $\T$ we obtain controllers and funnels using the method described in Section \ref{sec:funnels}. We obtained time-varying Taylor expansions of degree 3 computed around the nominal trajectory. The approach from Section \ref{sec:funnels_controller} along with the time-sampled approximation described in Section \ref{sec:funnels_time_sampling} (with 15 time samples) was used to synthesize a (time-varying) linear feedback controller around each trajectory. The methods described in Section \ref{sec:funnels_uncertainty} were used to take into account the parametric uncertainty that the system is subject to. The computation time for each funnel was approximately 20-25 minutes. The directed graph $\GF$ that encodes real-time composability between funnels (ref. Section \ref{sec:funnel library definition}) is fully connected. Moreover, the funnel corresponding to the trim trajectory is sequentially composable modulo invariances (ref. Section \ref{sec:sqmi}) with the other maneuvers in the library, allowing us to apply the trim trajectory before or after any of the maneuvers. A subset of the funnels is shown in Figure \ref{fig:quad_funnel_library}. Note that the twelve-dimensional funnels have been projected down to the $x-y-z$ dimensions for the purpose of visualization.

Given this funnel library $\mathcal{F}$, we will show how one can derive simple geometric conditions on the environment that guarantee that a collision-free funnel will always be found during real-time planning. In order to simplify the analysis, we will assume that the quadrotor is navigating through a 2.5D environment (i.e., the obstacles in the environment are 2D polygons extruded in the z-direction). But, we note that our analysis can be extended to fully three dimensional environments. We will treat the geometry of the quadrotor as a sphere of radius $r_{quad}$. 

\begin{figure*}
  \begin{center}
    \includegraphics[trim = 0mm 0mm 0mm 0mm, clip, width=0.5\textwidth]{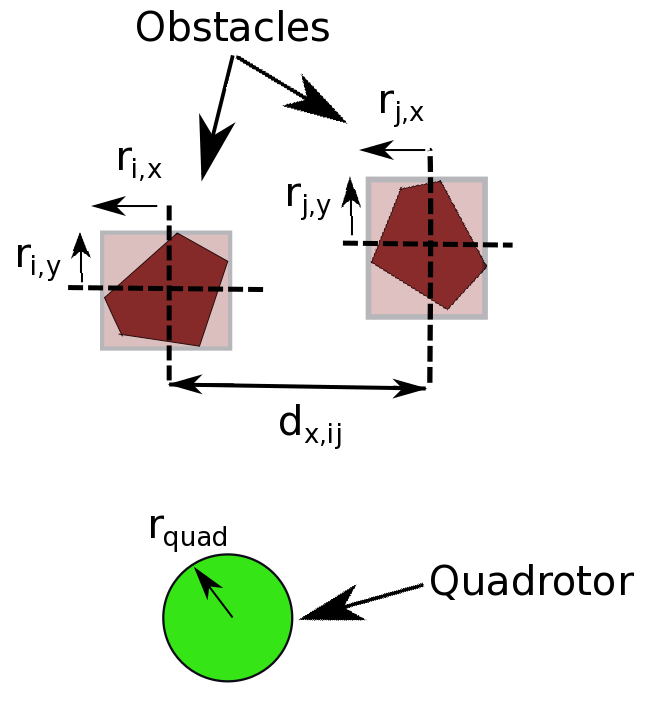}
  \end{center}
  \caption{Obstacle bounding boxes and quadrotor radius.
  \label{fig:quad_analysis_obs}}
\end{figure*}

Since we are only considering 2.5D environments, it is sufficient to project the environment and funnels down to the $x-y$ plane. Consider (axis aligned) bounding boxes for the 2D obstacles (see Figure \ref{fig:quad_analysis_obs}). Denote the center of the bounding box for obstacle $o_i$ as $(o_{i,x,},o_{i,y})$ for $i \in \{1,\dots,N_{obs}\}$ ($N_{obs}$ is the number of obstacles in the robot's sensor horizon). Suppose without loss of generality that the quadrotor's x-y-z position is $(0,0,0)$.  We will first derive conditions on the distances $o_{i,y}$, $d_{x,ij} := |o_{i,x} - o_{j,x}|$, and sizes $r_{x,i}$, $r_{y,i}$ (ref. Figure \ref{fig:quad_analysis_obs}) such that there exists a collision-free funnel in $\F$ \emph{assuming} the quadrotor's x-y-z position is $(0,0,0)$. We will then extend this analysis to derive conditions that guarantee that a collision-free funnel will always be found as the quadrotor applies Algorithm \ref{a:replan_funnel} in a receding horizon manner to fly continuously through the environment.

\begin{figure}
\centering

\subfigure[]{\includegraphics[trim = 0mm 0mm 0mm 0mm, clip, width=0.52\columnwidth]{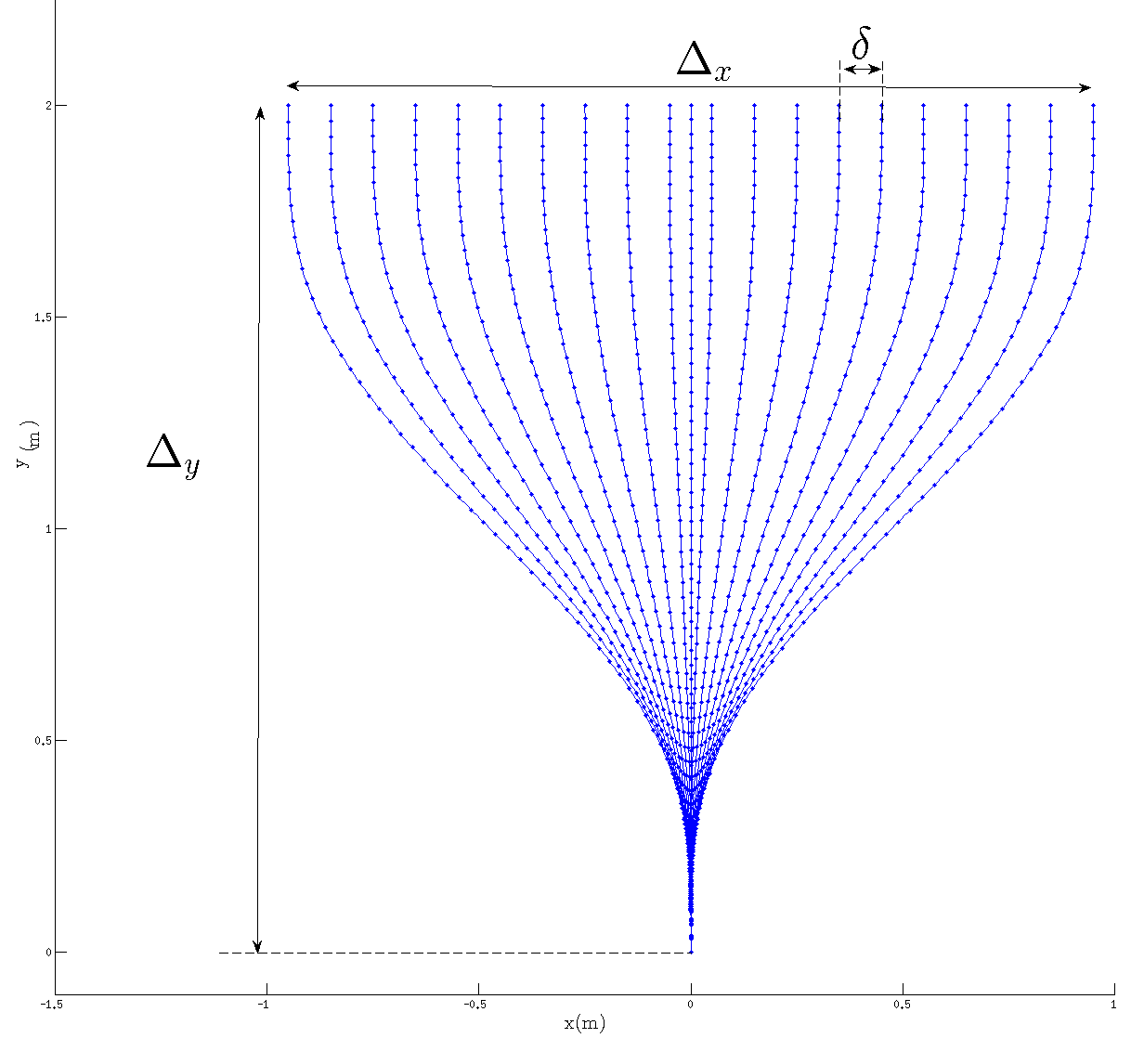} \label{fig:quad_analysis_traj}}    
\subfigure[]{\includegraphics[trim = 0mm 0mm 0mm 0mm, clip, width=0.44\columnwidth]{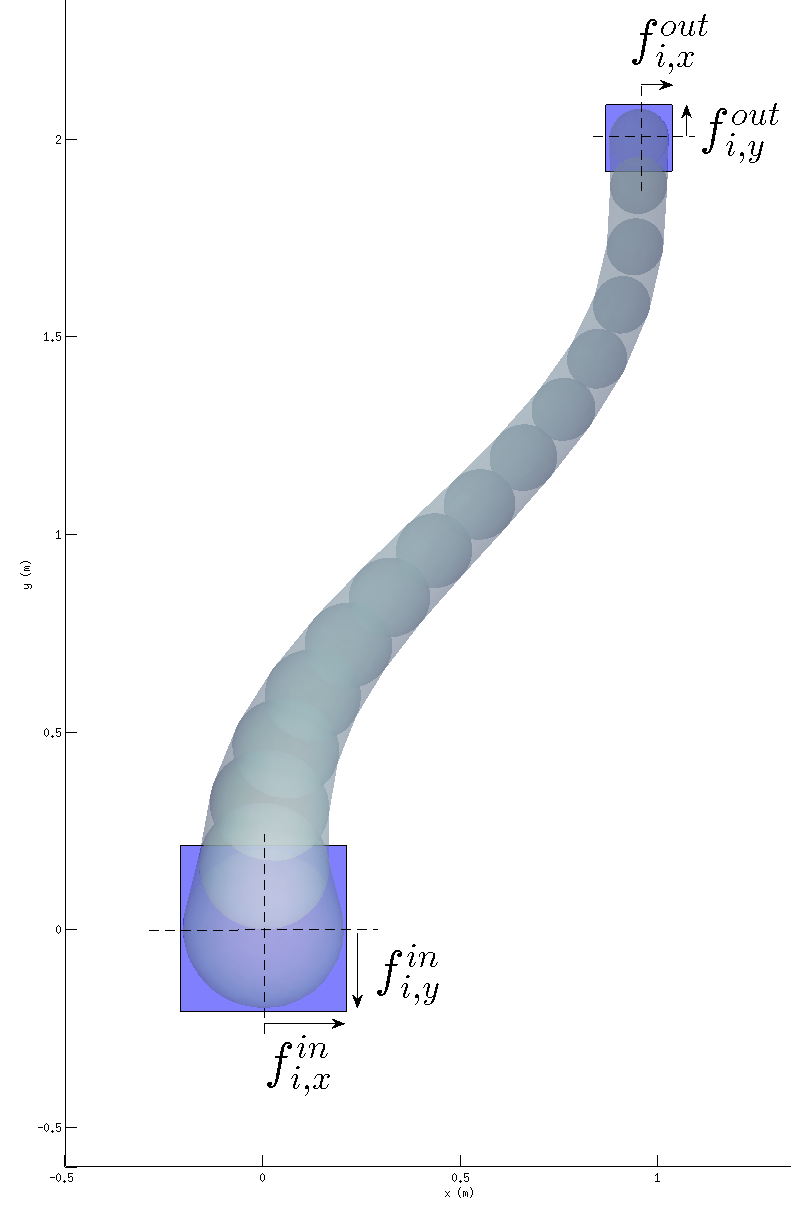} \label{fig:quad_analysis_funnel}} 

\caption[Notation for geometry of trajectories and funnels]{Notation for geometry of trajectories and funnels.}
\end{figure}

For each funnel $F_i \in \F$ compute bounding boxes for the inlet and outlet of the funnel (see Figure \ref{fig:quad_analysis_funnel}). We will refer to the dimensions of these bounding boxes as depicted in the figure as $f_{i,y}^{out}, f_{i,x}^{out}, f_{i,y}^{in}, f_{i,x}^{in}$. Define 
$$f_x := \max (f_{1,x}^{out},...,f_{N,x}^{out},f_{1,x}^{in},...,f_{N,x}^{in}) + r_{quad},$$
where $N$ is the number of funnels in the library (21 in our case). Similarly, define:
$$f_y := \max (f_{1,y}^{out},...,f_{N,y}^{out},f_{1,y}^{in},...,f_{N,y}^{in}) + r_{quad}.$$

Next, define $\delta$ as shown in Figure \ref{fig:quad_analysis_traj} as the distance (in the $x$-dimension) of the endpoints of trajectories in $\mathcal{T}$. Denote by $\Delta_x$ the distance (in the $x$-dimension) between the endpoints of the most aggressive trajectories in $\mathcal{T}$ and let $\Delta_y$ be the distance in the $y$-direction that each trajectory in the library covers (see Figure \ref{fig:quad_analysis_traj}). 
\vspace{40pt}
Define $r_x := \max(r_{1,x},...,r_{N_{obs},x})$,  $r_y := \max(r_{1,y},...,r_{N_{obs},y})$. Then consider the following conditions on the obstacles:

\begin{align}
& o_{i,y} > \Delta_y \label{eq:obs_cond_1}, \ \forall i \in \{1,\dots,N_{obs}\} \\
& r_{y} < f_y \label{eq:obs_cond_2} \\
& r_{x} < \Delta_x - 2f_x \label{eq:obs_cond_3} \\
& d_{x,ij} > 2f_x + \delta, \ \forall i,j \in \{1,\dots,N_{obs}\}, \ i \neq j. \label{eq:obs_cond_4}  
\end{align}

It is straightforward to see that these conditions imply the existence of a funnel in $\F$ that is collision-free when executed from the quadrotor's location at $(0,0,0)$ (i.e., Algorithm \ref{a:replan_funnel} will succeed assuming that the full 12 dimensional state of the quadrotor is contained within the inlet of all the funnels in $\F$). In particular, the condition \eqref{eq:obs_cond_1} ensures that the funnels are not too close to the quadrotor's starting location. Condition \eqref{eq:obs_cond_2} prevents an obstacle from extending so far in the $y$-direction that it collides with the middle segments of a funnel. Condition \eqref{eq:obs_cond_3} ensures that obstacles do not extend so far in the $x$-direction that there is no funnel in the library that goes around the obstacle (e.g., imagine a wall in front of the quadrotor). And finally \eqref{eq:obs_cond_4} ensures that the gap in the $x$-direction between any two obstacles is large enough for \emph{some} funnel in $\mathcal{F}$ to fit through. Then conditions \eqref{eq:obs_cond_3} and \eqref{eq:obs_cond_4} ensure that this gap occurs in a portion of the space that is reachable by one of the funnels in the library (and not in some far off location in the $x$-direction where a funnel in $\F$ does not extend to).

\begin{figure*}
  \begin{center}
    \includegraphics[trim = 0mm 0mm 0mm 0mm, clip, width=0.95\textwidth]{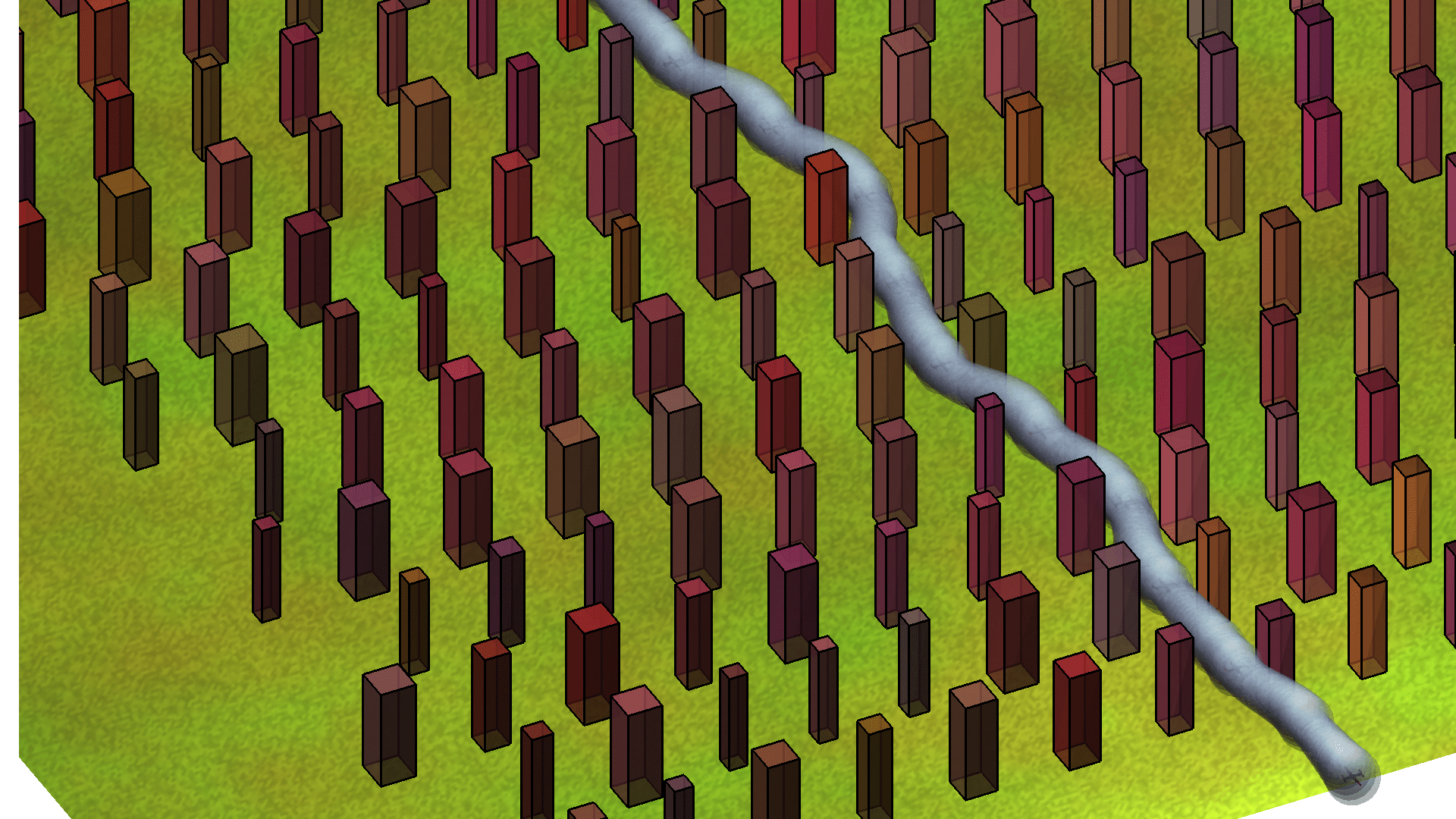}
  \end{center}
  \caption[Quadrotor system navigating through an environment]{The plot shows the quadrotor maneuvering through a forest of polytopic obstacles in a collision-free manner. The environment satisfies simple geometric conditions that allow us to guarantee collision-free flight forever. Note that the visualized funnels have been inflated to take into account the size of the quadrotor. 
  \label{fig:quad_navigation}}
\end{figure*}

We can now extend this analysis to the scenario in which the quadrotor is flying along continuously through the environment by applying Algorithm \ref{a:replan_funnel} in a receding horizon manner. We will assume that the sensor horizon of the quadrotor (i.e., the range in which obstacles are reported) is greater than $\Delta_y$ in the $y$-direction and greater than $\Delta_x/2$ in the $x$-direction. We then need to ensure that when the quadrotor finishes executing a funnel, it does not find itself in a position where there are obstacles that are very close to it. In order to do this, we will first replace the conditions \eqref{eq:obs_cond_1} and \eqref{eq:obs_cond_4} with the following condition on the separation between obstacles:
\begin{align}
& d_{y,ij} := |o_{i,y} - o_{j,y}| > \Delta_y + f_y \\
& \textrm{OR} \nonumber \\
& d_{x,ij} > 2f_x + \delta.                        
\end{align}
This condition ensures that obstacles are separated enough in \emph{either} the $x$ or $y$ directions. However, this is still not enough to prevent cases where the quadrotor executes a funnel and finds itself too close to an obstacle. In particular, suppose that the quadrotor starts off at location $(0,0,0)$ and that there is a single obstacle in the environment located at $(0,1.1\Delta_y+f_y)$ (the size of the obstacle will not matter in this example). When the quadrotor applies Algorithm \ref{a:replan_funnel} at location $(0,0,0)$  to find a funnel, all the funnels in the library will be collision-free. Let us suppose that the algorithm chooses the funnel corresponding to the quadrotor flying straight and ends up in location $(0,\Delta_y)$. At this point, the obstacle is too close to the quadrotor and a collision-free funnel will not be found. To prevent cases such as this, we can use the trim maneuver in $\F$ corresponding to the quadrotor flying straight in order to ``pad" the distance to the obstacles. In particular, we can apply the trim maneuver until the quadrotor is at distance $\Delta_y$ from the obstacle (in the $y$-direction) and \emph{then} use Algorithm \ref{a:replan_funnel} to replan. This will guarantee that Algorithm \ref{a:replan_funnel} will succeed and thus our analysis is complete.

Figure \ref{fig:quad_navigation} shows an example of the quadrotor system navigating through an environment that satisfies the geometric conditions derived above. The figure shows the sequence of funnels executed by the quadrotor to traverse the environment. Note that the funnels depicted in the plot have been inflated by $r_{quad}$ in order to take into account the physical extent of the quadrotor. 

We end this section by noting that our goal here has been to demonstrate the possibility of imposing geometric conditions on the environment that guarantee collision-free flight. \revision{While fairly straightforward, this analysis may provide a useful starting point for more sophisticated analyses.} Going forwards, our analysis may be varied or tightened in many ways.
For example, we can extend it in a straightforward manner to fully 3D environments. Further, we have assumed that the quadrotor is only planning a single funnel at a time (in contrast to sequences of funnels). Planning sequences of funnels may help loosen some of the restrictions on the environment that we have made in our analysis. \revision{Another avenue is to consider cases where the locations of obstacles in the environment are drawn from a (known) stochastic process. We discuss a few possibilities for handling such settings in Section \ref{sec:designing libraries}.}

%% file: hardware.tex
\section{Hardware experiments on a fixed-wing airplane}
\label{sec:hardware}

In this section, we will validate the key components of the approach presented in this paper on a small fixed-wing airplane performing a challenging obstacle avoidance task. The goal of these hardware experiments is to answer the following important practical questions: 

\begin{itemize}

\item Can we obtain models of a real-world challenging nonlinear dynamical system that are accurate enough to compute funnels that are valid in reality? 

\item Can we implement the real-time planning algorithm described in Section \ref{sec:planning} to operate at the required rate given realistic computational constraints?

\item Can we demonstrate our planning algorithm on a realistic and challenging obstacle avoidance task? 

\end{itemize}

\subsection{Hardware platform}

\begin{figure}
 \centering
   \subfigure[The E-flite SBach RC airplane used for hardware experiments. \label{fig:sbach}]{\includegraphics[width=0.7\columnwidth]{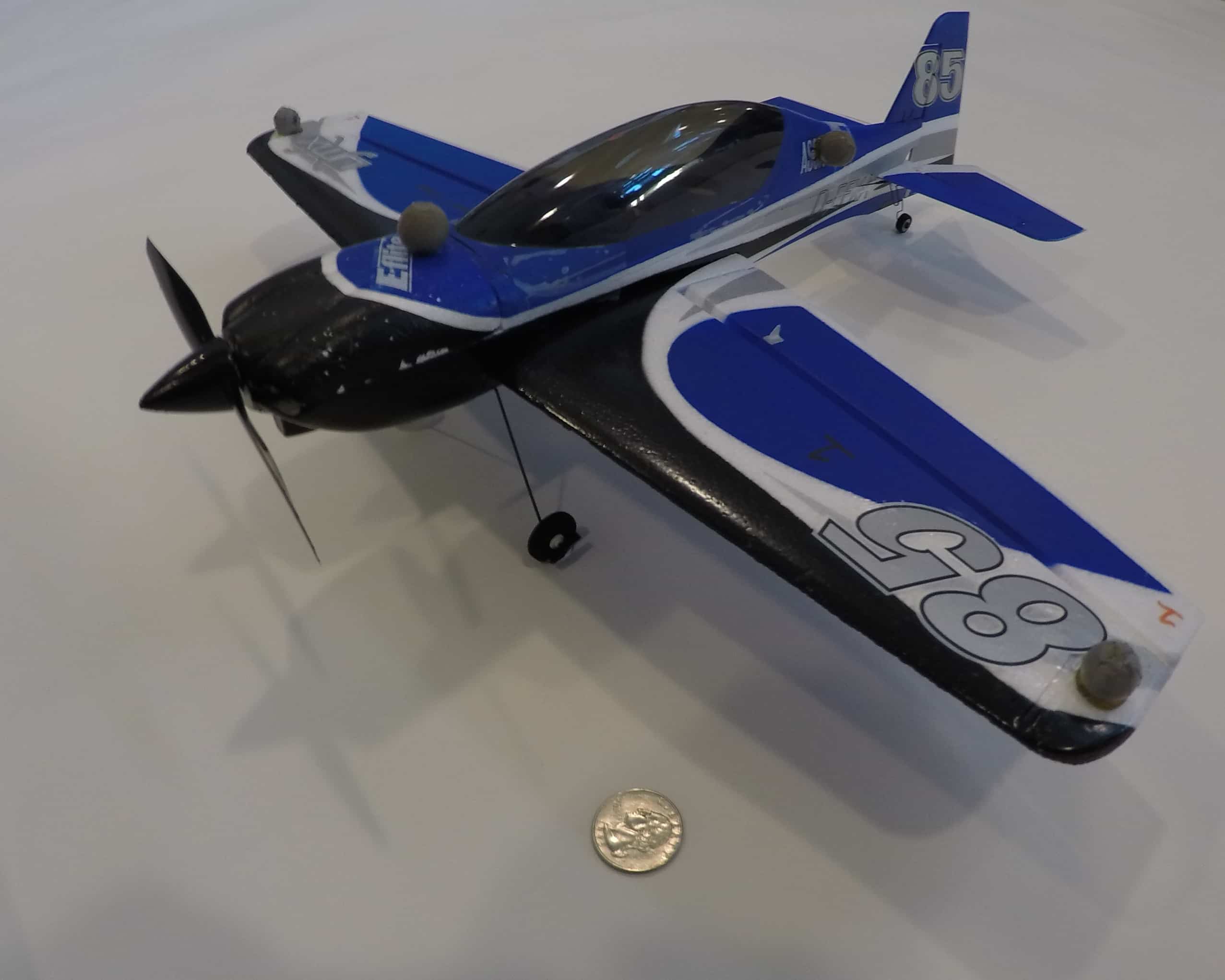}}
   \subfigure[Experimental setup and coordinate system. \label{fig:experimental_setup}]{\includegraphics[trim = 0mm 0mm 0mm 0mm, clip, width=0.99\columnwidth]{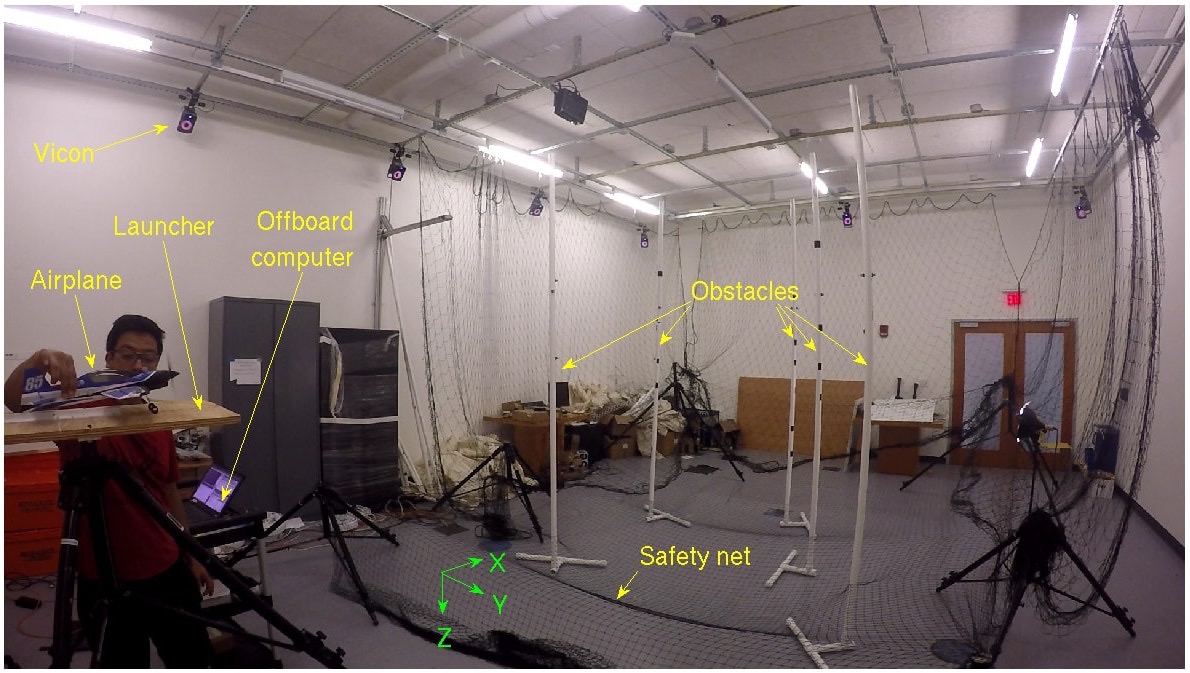}}
    \caption[Fixed-wing airplane and setup used for hardware experiments]{Airplane hardware and experimental setup. \label{fig:sbach_hardware_setup}}
\end{figure}

The hardware platform chosen for these experiments is the SBach RC airplane manufactured by E-flite shown in Figure \ref{fig:sbach}. The airplane is very light ($76.6$ g) and highly maneuverable, thus allowing for dramatic obstacle avoidance maneuvers in a tight space. The control inputs to the SBach are raw servo commands to the control surfaces (ailerons, rudder, elevator) and a raw throttle setting. These commands are sent through a modified 2.4 GHz RC transmitter at an update rate of $50$ Hz.

\subsection{Task and experimental setup}
\label{sec:sbach_task}

The experimental setup is shown in Figure \ref{fig:experimental_setup}. The airplane is launched from a simple rubber-band powered launch mechanism at approximately $4-5$ m/s. The goal is to traverse the length of the room while avoiding the obstacles placed in the experimental arena. The airplane's planner is \emph{not} informed where the obstacles are beforehand; rather, the obstacle positions and geometry are reported to the planner only once the airplane clears the launcher. This simulates the receding-horizon nature of realistic obstacle avoidance tasks where the obstacle positions are not known beforehand and planning decisions must be taken in real-time. The experiments are performed in a Vicon motion capture arena that reports the airplane and obstacle poses at 120 Hz. All the online computation is performed on an off-board computer with four Intel i7 2.9GHz processors and 16 GB RAM.

\subsection{Modeling and system identification}
\label{sec:sbach_model}

Our dynamics model of the airplane is based on the model described in \cite{Sobolic09} (\cite{Stevens92} is also a good reference for modeling fixed-wing airplanes). The model has $12$ states:
$${\bf{x}} = [x,y,z,\phi,\theta,\psi,\dot{x},\dot{y},\dot{z},P,Q,R].$$
Here, $+x$ is in the forward direction, $+y$ is to the right and $+z$ is downwards as depicted in Figure \ref{fig:experimental_setup} (this is the standard North-East-Down coordinate frame used in aeronautics). The states $\phi,\theta,\psi$ are the roll, pitch and yaw angles. The variables $P,Q,R$ are the components of the angular velocity expressed in the body coordinate frame. The control inputs of the model are the angles of the ailerons, rudder and elevator, along with the speed of the propellor. Thus, we have $4$ control inputs since the ailerons are coupled to deflect in opposite directions by the commanded magnitude.

The airplane is treated as a rigid body with aerodynamic and gravitational forces acting on it. Here, we provide a brief description of our model of the aerodynamic forces:

\quad 

\noindent {\bf{Propellor thrust} } The thrust from the propellor is proportional to the square of the propellor speed. The constant of proportionality was obtained by hanging the airplane (with the propellor pointing downwards) from a digital fish scale and measuring the thrust produced for a number of different throttle settings. 

\quad

\noindent {\bf{Lift/drag on aerodynamic surfaces}} The lift and drag forces on the ailerons, rudder, elevator and tail of the airplane were computed using the flat-plate model. The flat-plate model was also used as a baseline for the lift and drag coefficients of the wings, but an angle-of-attack dependent correction term was added. This correction term was fit from experimental data obtained from passive (i.e., unactuated) flights in a manner similar to \cite{Moore14b}. Since lift and drag forces are dependent on the airspeed over the aerodynamic surface, we need to take into account the effect of ``propwash" (i.e. the airflow from the propellor). The relationships between the throttle speed command and the propellor downwash speed over the different control surfaces were measured using a digital anemometer in a manner similar to \cite{Sobolic09}. 

\quad

\noindent {\bf{Body drag}} The drag on the airplane body is approximated as a quadratic drag term whose drag coefficient is fit from data.

\quad

As described above, many of the parameters in the model were obtained directly from physical experiments and measurements. However, some model parameters are more difficult to measure directly. These include the moments of inertia of the airplane and the coefficient of drag associated with the airplane body. The prediction-error minimization method in the Matlab System Identification Toolbox \cite{Ljung07} was used to fit these parameters and to fine-tune the measured parameters. In particular, we collected data from 15-20 flights (each lasting approximately $0.5$--$0.7$ seconds) where the control inputs were excited using sinusoidal signals of varying frequency and amplitude.

\subsection{Funnel validation}
\label{sec:sbach_funnel}

\begin{figure}
 \centering
   \subfigure[A depiction of the funnel that was validated on hardware. The funnel has been projected down to the $x-y-z$ coordinates of the state space and then reprojected onto the camera image. \label{fig:sbach_funnel_hardware}]{\includegraphics[trim = 0mm -50mm 0mm 0mm, clip, width=0.5\columnwidth]{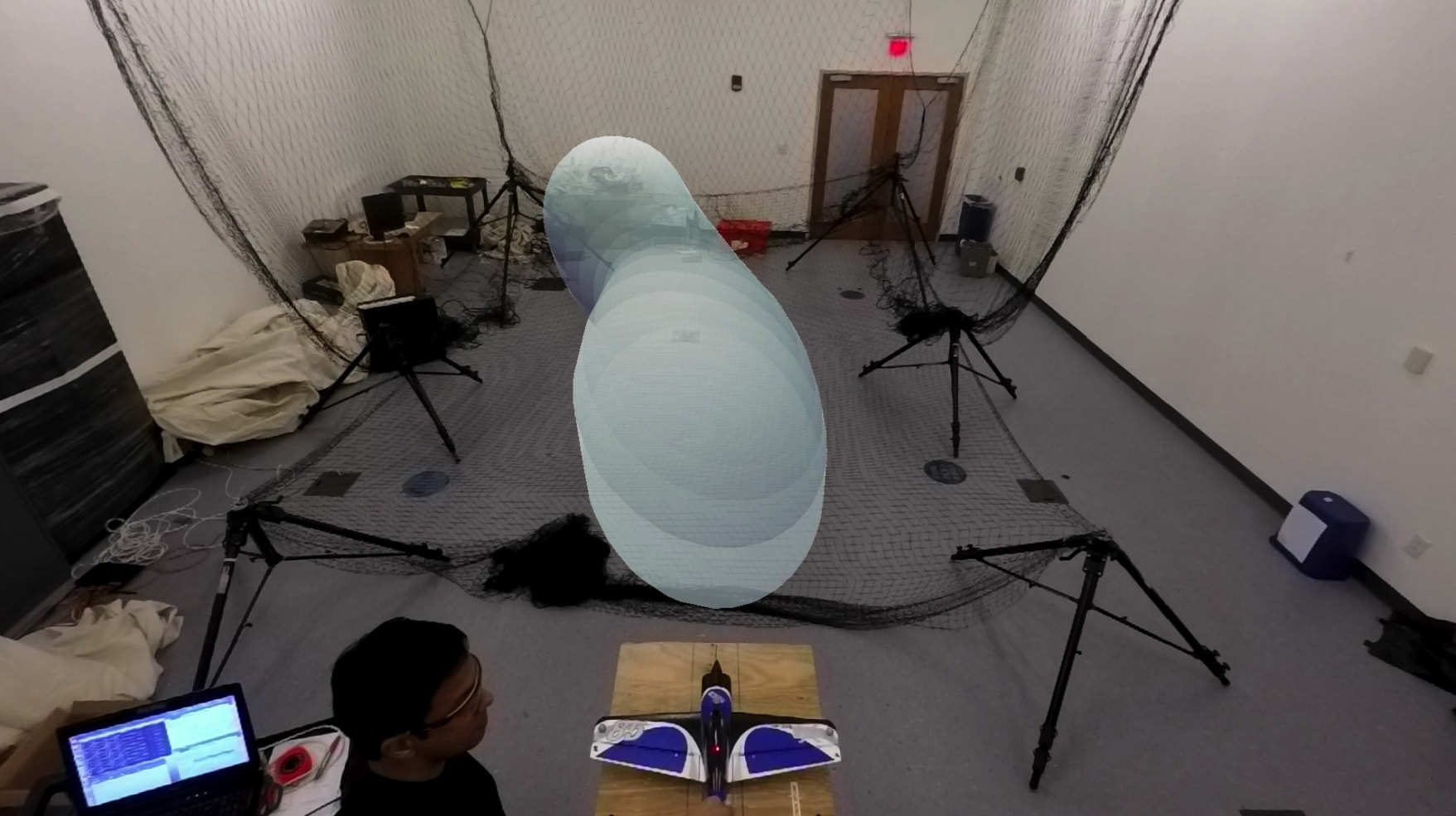}}
   \subfigure[The value of the Lyapunov function ($V$) plotted as a function of time for $30$ different trials of the airplane started from different initial conditions in the inlet of the funnel. The 1-sublevel set of the Lyapunov function corresponds to the funnel (i.e.,  a Lyapunov function of 1 or less corresponds to the airplane being inside the funnel). All 30 of the trajectories remain inside the computed funnel for the entire duration of the maneuver. \label{fig:sbach_V_vs_t}]{\includegraphics[width=0.49\columnwidth]{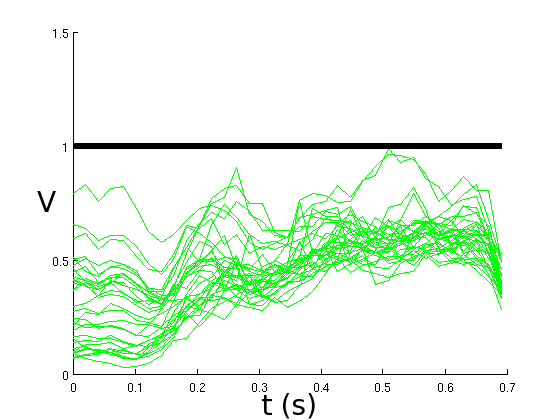}}
    \caption{Validating funnels on the fixed-wing airplane.  \label{fig:sbach_funnel_validation}}
\end{figure}

The first major goal of our hardware experiments was to demonstrate that our model of the SBach airplane is accurate enough to compute funnels that are truly meaningful for the hardware system. To this end, we computed a funnel (shown in Figure \ref{fig:sbach_funnel_hardware}) for the airplane using the approach described in Section \ref{sec:funnels}. We first estimated the set of initial states that the launcher mechanism causes the airplane to start off in (here, by ``initial state" we mean the state of the airplane as soon as it has cleared the launcher mechanism). This was done by fitting an ellipsoid around the initial states observed from approximately 50 experimental trials. We then used direct collocation trajectory optimization \cite{Betts01} to design an open-loop maneuver that makes the airplane bend towards the left. The initial state of the trajectory is constrained to be equal to the mean of the experimentally observed initial states and the control inputs are constrained to satisfy the limits imposed by the hardware. Next we computed a time-varying LQR (TVLQR) controller around this nominal trajectory. This controller was tuned largely in simulation to ensure good tracking of the trajectory from the estimated initial condition set. The resulting closed-loop dynamics were then Taylor expanded around the nominal trajectory to degree 3 in order to obtain polynomial dynamics. Finally, we used SOS programming to compute the funnel depicted in Figure \ref{fig:sbach_funnel_hardware} using the time-sampled approximation described in Section \ref{sec:funnels_time_sampling} (with 10 time samples). The inlet of the funnel was constrained to contain the experimentally estimated set of initial conditions. We observed that the tuned TVLQR controller does not saturate the control inputs for the most part and thus we did not find it necessary to take input saturations into account in our funnel computation. Further, we wanted to assess the validity of our funnel for the nominal dynamics model of the airplane and did not take into account any uncertainty in the model. The funnel computation took approximately 1 hour.

\begin{figure}[t!]
 \centering
   \subfigure[]{\includegraphics[trim = 45mm 15mm 100mm 30mm, clip, width=0.49\columnwidth]{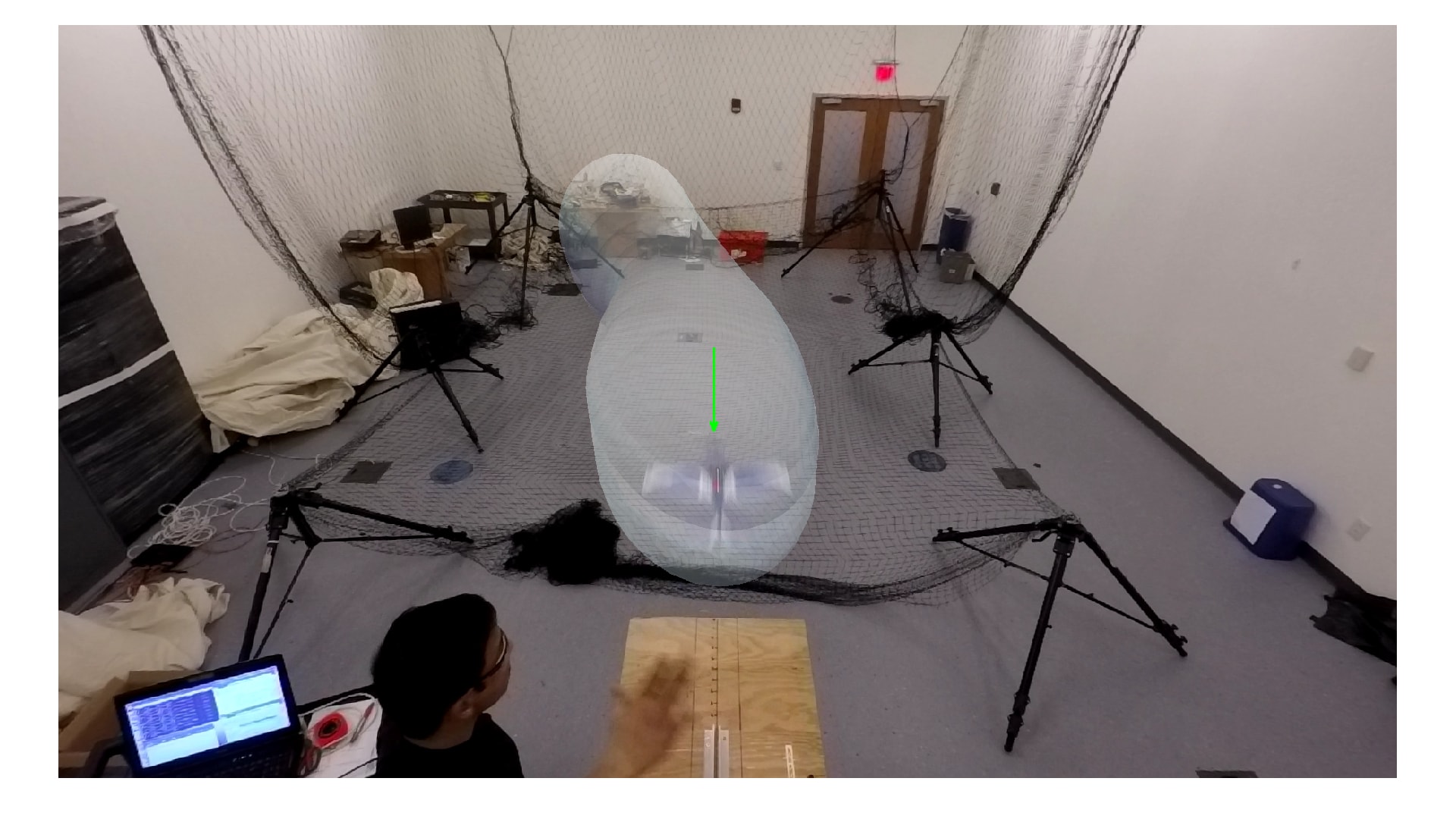}}
   \subfigure[]{\includegraphics[trim = 90mm 39mm 120mm 48mm, clip, width=0.49\columnwidth]{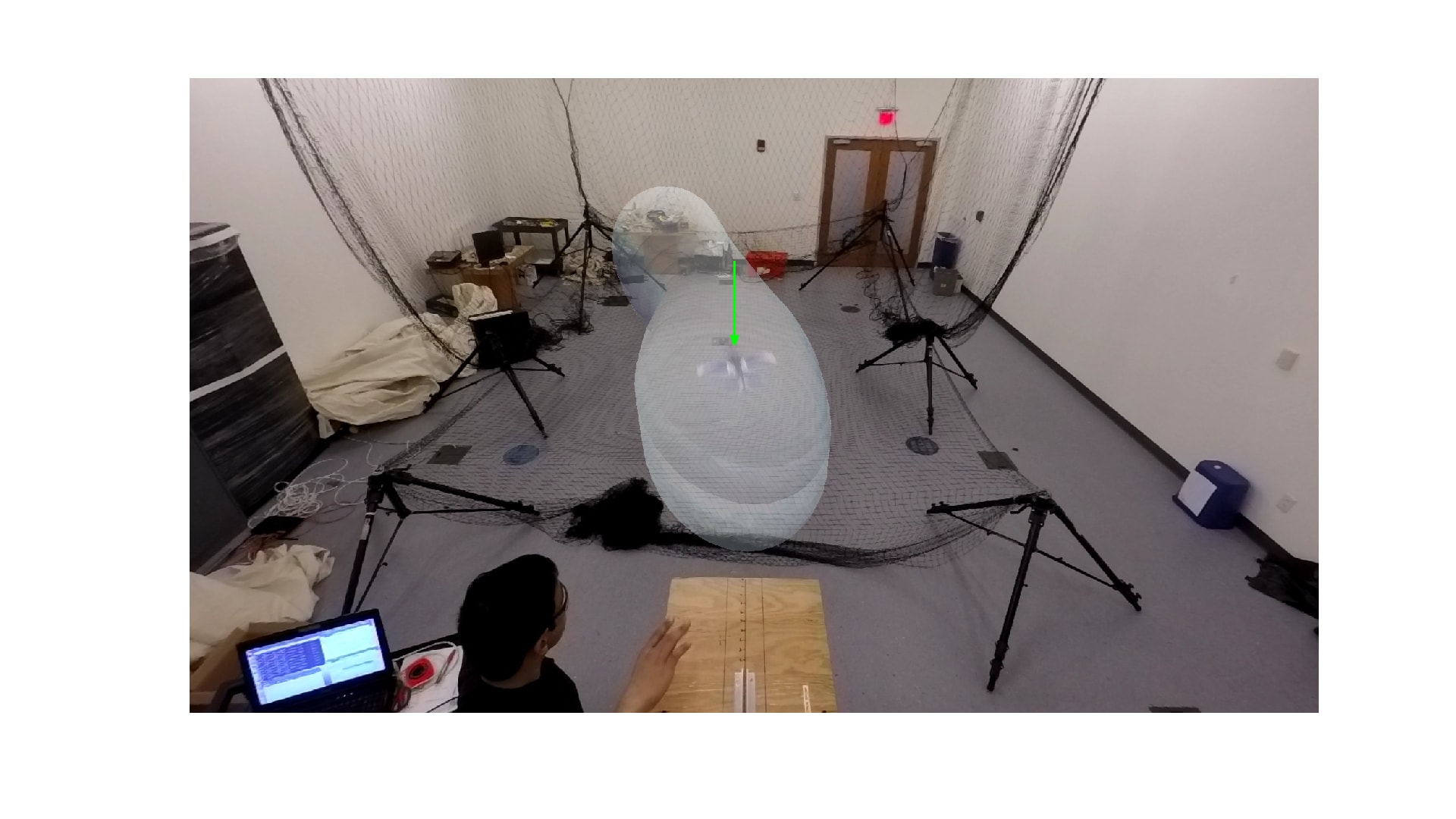}}
\subfigure[]{\includegraphics[trim = 90mm 39mm 120mm 48mm, clip, width=0.49\columnwidth]{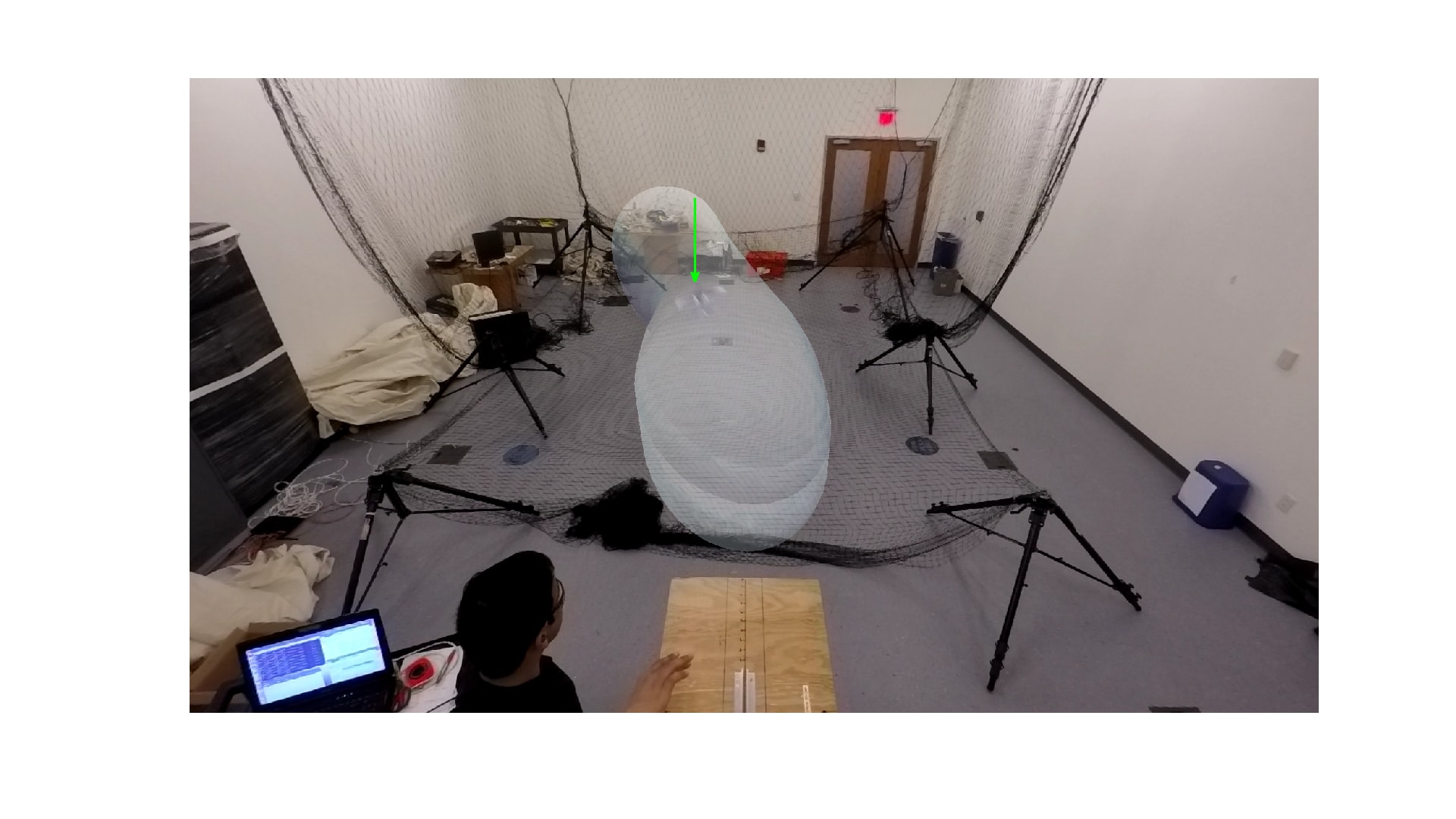}}    
\subfigure[]{\includegraphics[trim = 90mm 39mm 120mm 48mm, clip, width=0.49\columnwidth]{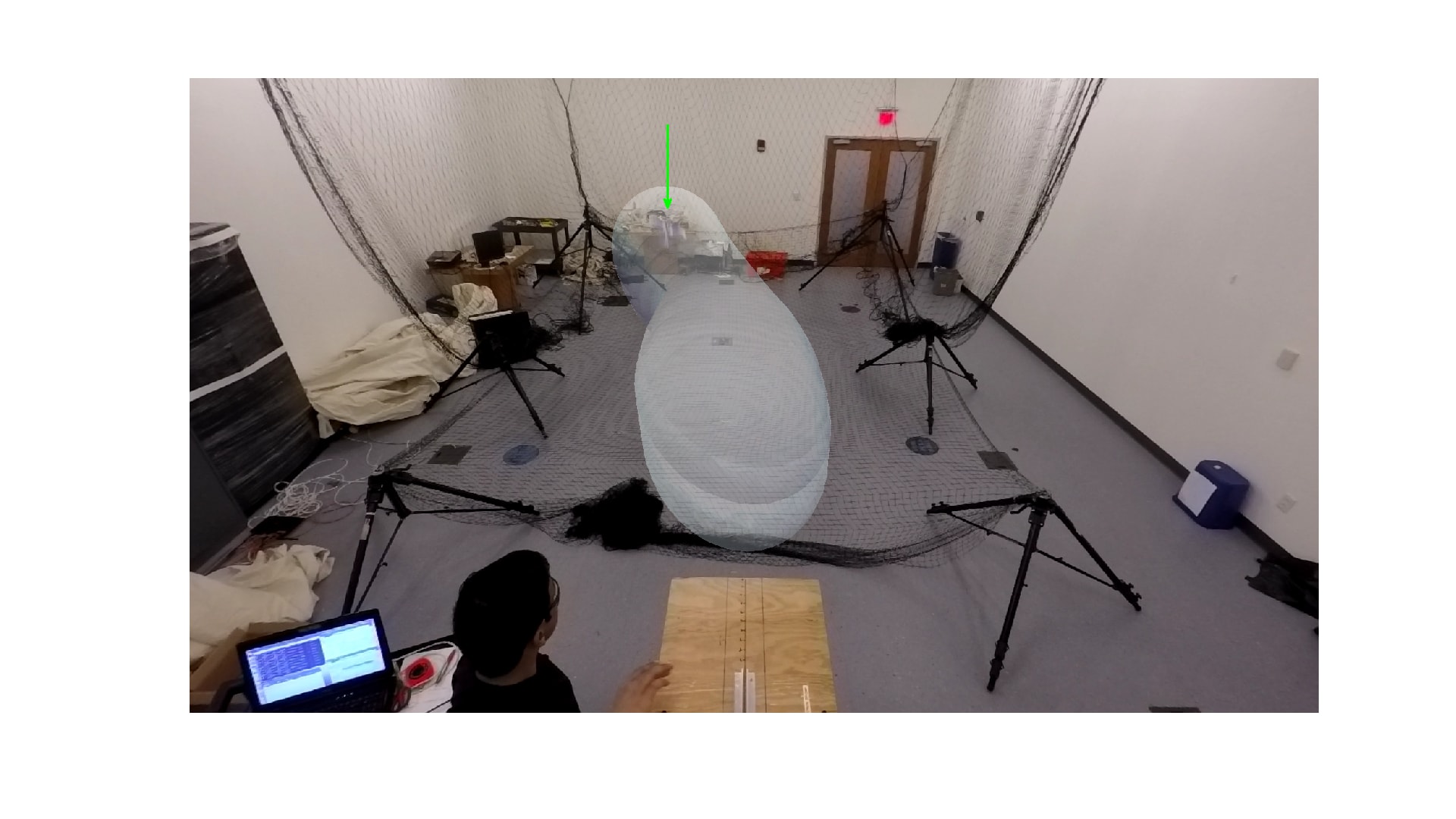}} \caption[Still images of the airplane flying through a funnel]{Still images of the SBach flying through a funnel. The funnel has been projected down to the $x-y-z$ coordinates of the state space and then reprojected onto the camera image. \revision{A video with a visualization of the funnel and flights through it is available online at \href{https://youtu.be/cESFpLgSb50}{https://youtu.be/cESFpLgSb50}.} \label{fig:sbach_funnel_stills}}
\end{figure}

We validate the funnel shown in Figure \ref{fig:sbach_funnel_hardware} with $30$ experimental trials of the airplane executing the maneuver corresponding to the funnel. The airplane is started off in different initial conditions in the inlet of the funnel and the TVLQR controller is applied for the duration of the maneuver. Figure \ref{fig:sbach_funnel_stills} shows still images of a sample flight of the airplane executing the maneuver with the funnel superimposed onto the images. A video with a visualization of the funnel and flights through it is available online at \href{https://youtu.be/cESFpLgSb50}{https://youtu.be/cESFpLgSb50}.

Figure \ref{fig:sbach_V_vs_t} provides a more quantitative perspective on the flights. In particular, the figure shows the value of the Lyapunov function $V(\bar{x},t)$ as a function of time achieved during the $30$ experimental trials. Here, the Lyapunov function has been normalized so that the 1-level set corresponds to the boundary of the funnel. As the plot illustrates, all $30$ trajectories remain inside the computed 12 dimensional funnel for the entire duration of the maneuver. This suggests that our model of the airplane is accurate enough to produce funnels that are indeed valid for the hardware system.

\subsection{Obstacle avoidance experiments}
\label{sec:sbach obstacles}

\begin{figure*}[h!]
  \begin{center}
    \includegraphics[trim = 0mm 0mm 0mm 0mm, clip, width=0.7\textwidth]{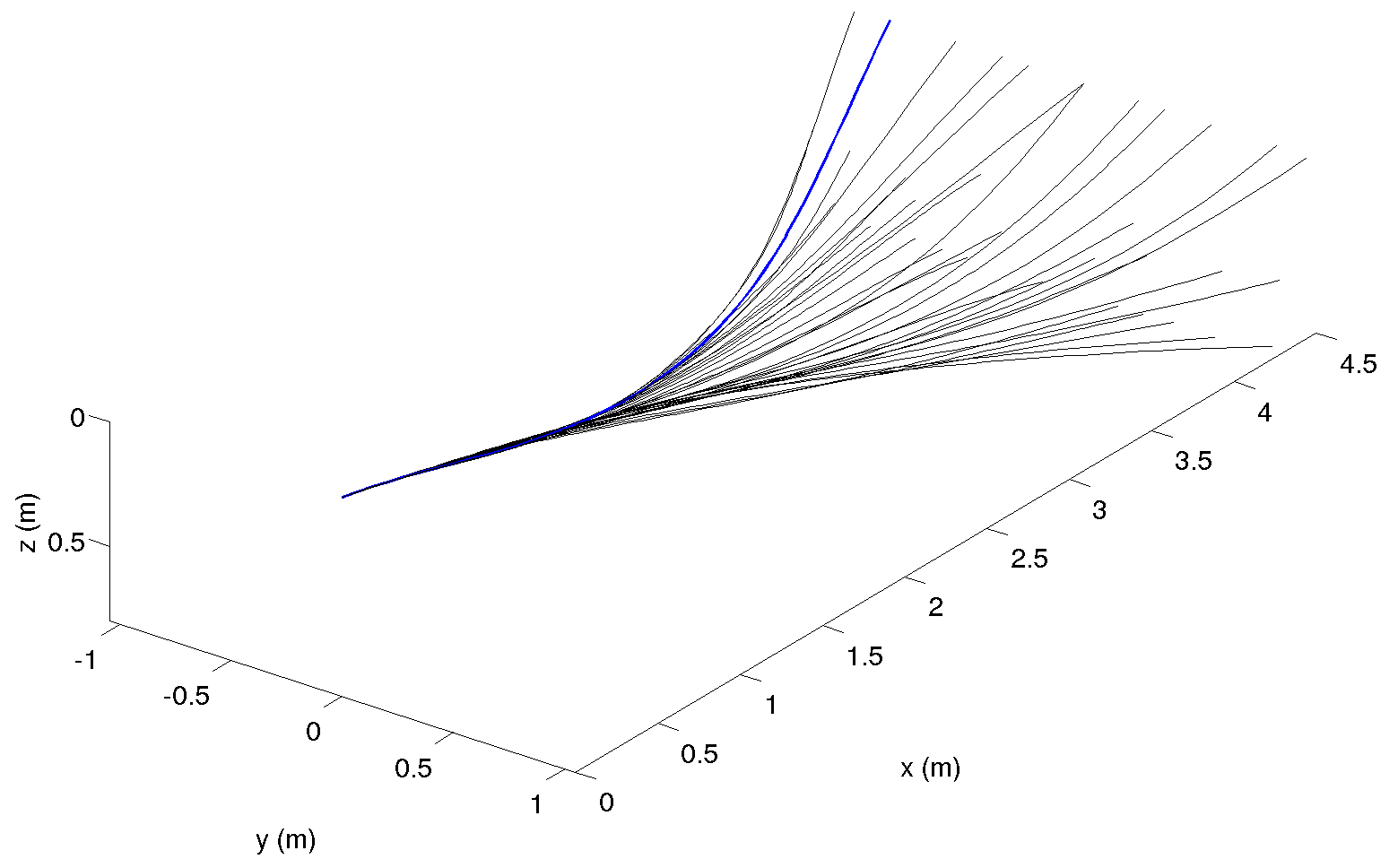}
  \end{center}
  \caption[Trajectory library for the airplane]{Trajectory library for the SBach airplane. The library consists of $40$ different maneuvers. The maneuver from Section \ref{sec:sbach_funnel} is highlighted in blue. \label{fig:sbach_traj_library}}
\end{figure*}

The second major goal of our hardware experiments was to demonstrate the funnel library based real-time planning algorithm proposed in Section \ref{sec:planning} on the obstacle avoidance task described in Section \ref{sec:sbach_task}. Our first step was to design a rich trajectory library consisting of a large number of different maneuvers. We initialized the library with the maneuver from Section \ref{sec:sbach_funnel} and augmented it by computing trajectories with varying final states. In particular, the library consists of $40$ trajectories which were obtained by discretizing the final state in the $y$ and $z$ coordinates and using direct collocation trajectory optimization to compute locally optimal trajectories for the airplane model described in Section \ref{sec:sbach_model}. The $x-y-z$ components of this trajectory library are depicted in Figure \ref{fig:sbach_traj_library}, with the trajectory from Section \ref{sec:sbach_funnel} highlighted in blue.

For each trajectory in our library, we computed a TVLQR controller with the same state/action costs as the controller for the maneuver in Section \ref{sec:sbach_funnel}. As in Section \ref{sec:sbach_funnel}, we used SOS programming to compute funnels for each trajectory in the library. We note that since \revision{our model of the} dynamics of the airplane is symmetric with respect to reflection about the $x$ axis, we were able to halve the amount of computation involved in constructing the trajectory/funnel library by exploiting this symmetry. 

As mentioned in Section \ref{sec:sbach_task}, the positions and geometry of the obstacles are not reported to the planner until the airplane has cleared the launcher. This forces planning decisions to be made in real-time. We use the planning algorithm described in Section \ref{sec:planning} to choose a funnel from our library. The planner employs the QCQP-based algorithm from Section \ref{sec:shifting funnels} for shifting funnels in the $x-y-z$ directions. As in the other examples considered in this paper, we use the FORCES Pro solver \cite{Domahidi14} for our QCQP problems. If no satisfactory funnel is found by the planner, we revert to a ``failsafe" option which involves switching off the propellor and gliding to a halt. \revision{A more sophisticated failsafe would be to transition to a ``propellor-hang" mode (with a corresponding funnel whose inlet contains the set of initial conditions the airplane may be in when it is launched).} However, we found that our failsafe provides adequate protection to the airplane's hardware as it usually glides on to the safety net before colliding with any obstacles. Finally, we note that since the experimental arena is quite limited in space, we do not replan funnels once one has been chosen; the airplane executes the feedback controller corresponding to the chosen funnel for the whole duration of the flight. 

\begin{figure}[h!]
 \centering
 \subfigure{\includegraphics[trim = 0mm 0mm 0mm 0mm, clip, width=0.49\columnwidth]{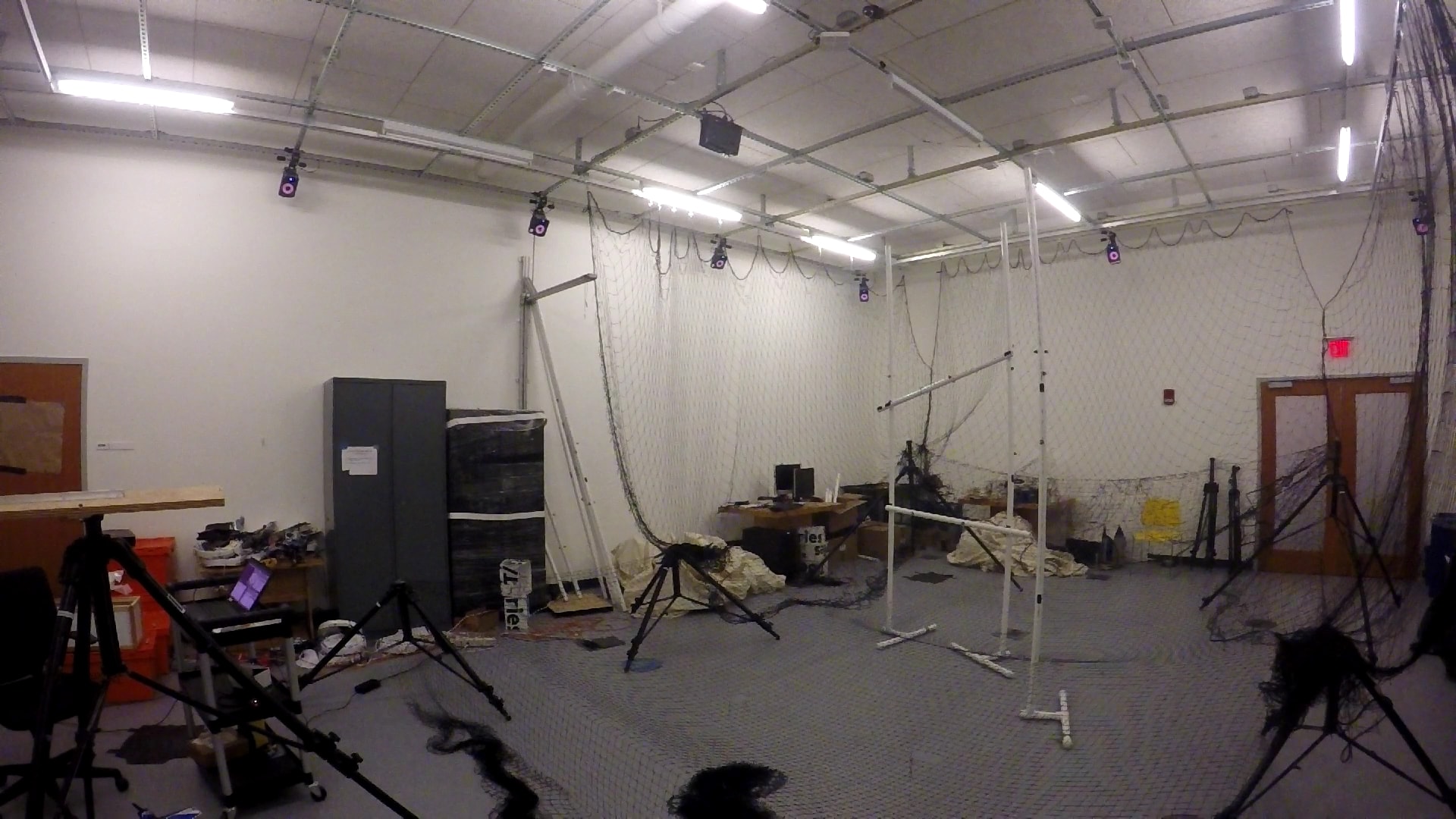}} 
 \subfigure{\includegraphics[trim = 0mm 0mm 0mm 0mm, clip, width=0.49\columnwidth]{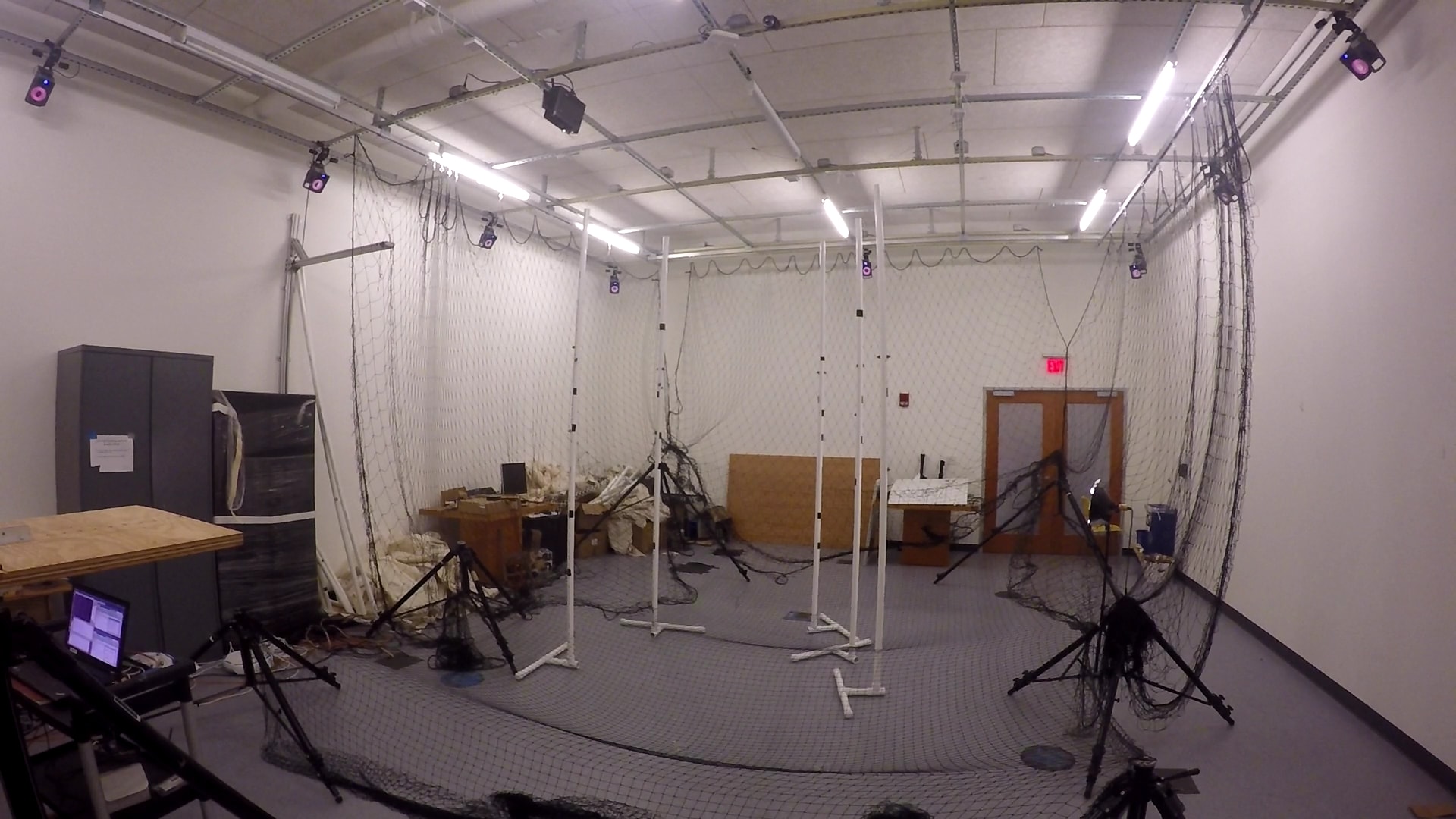}}    
\subfigure{\includegraphics[trim = 0mm 0mm 0mm 0mm, clip, width=0.49\columnwidth]{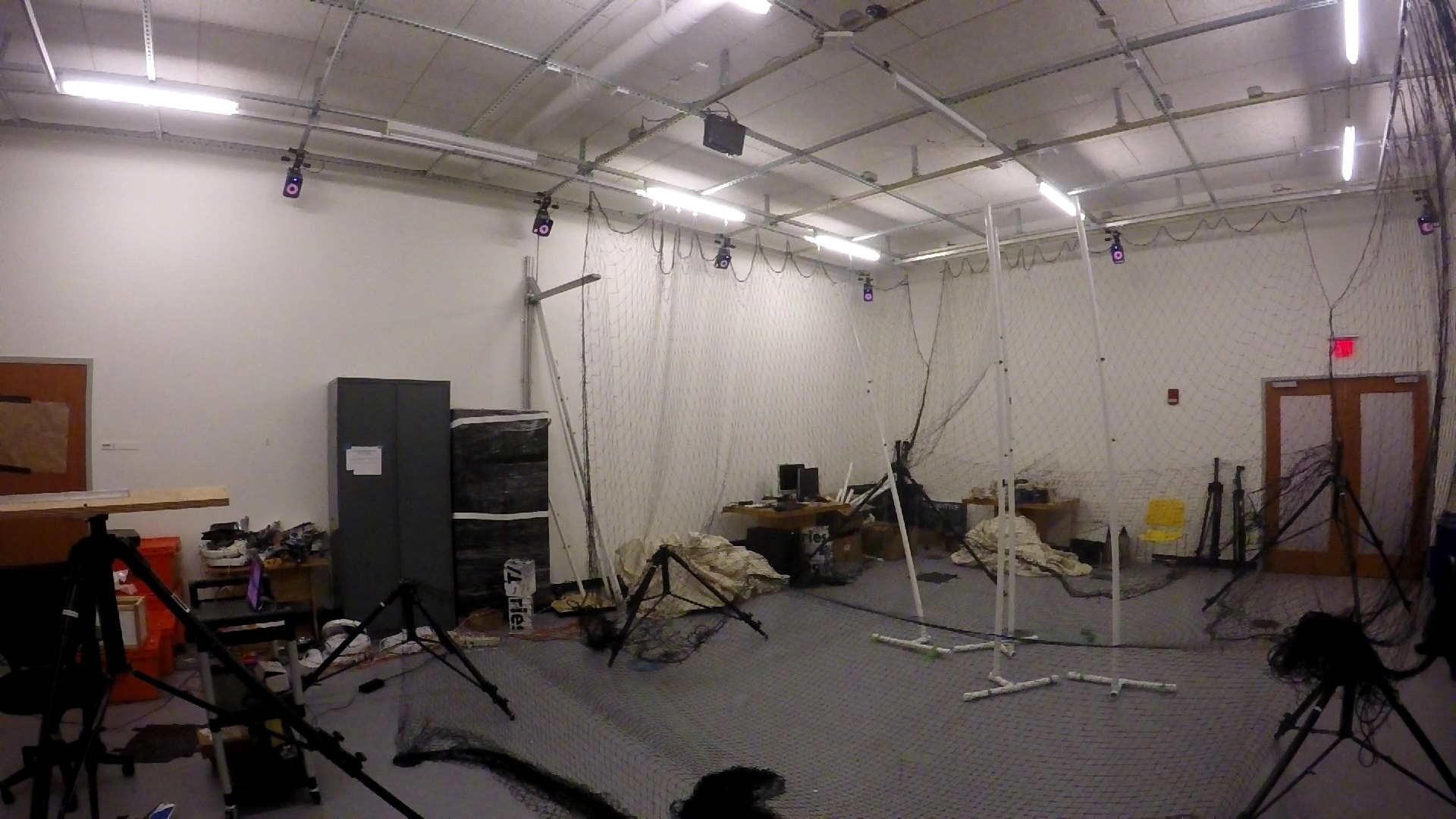}}    
   \subfigure{\includegraphics[trim = 0mm 0mm 0mm 0mm, clip, width=0.49\columnwidth]{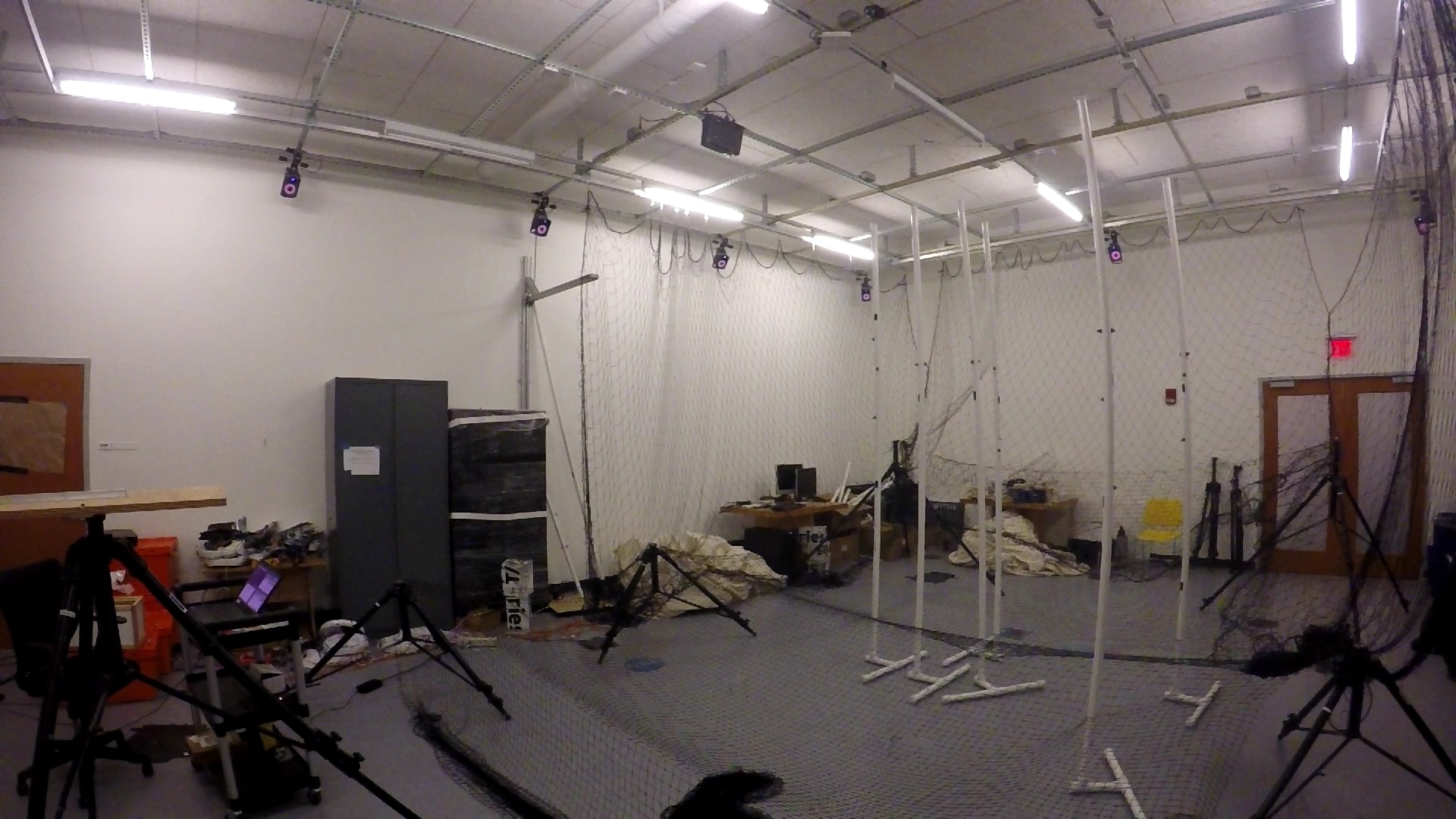}}
\subfigure{\includegraphics[trim = 0mm 0mm 0mm 0mm, clip, width=0.49\columnwidth]{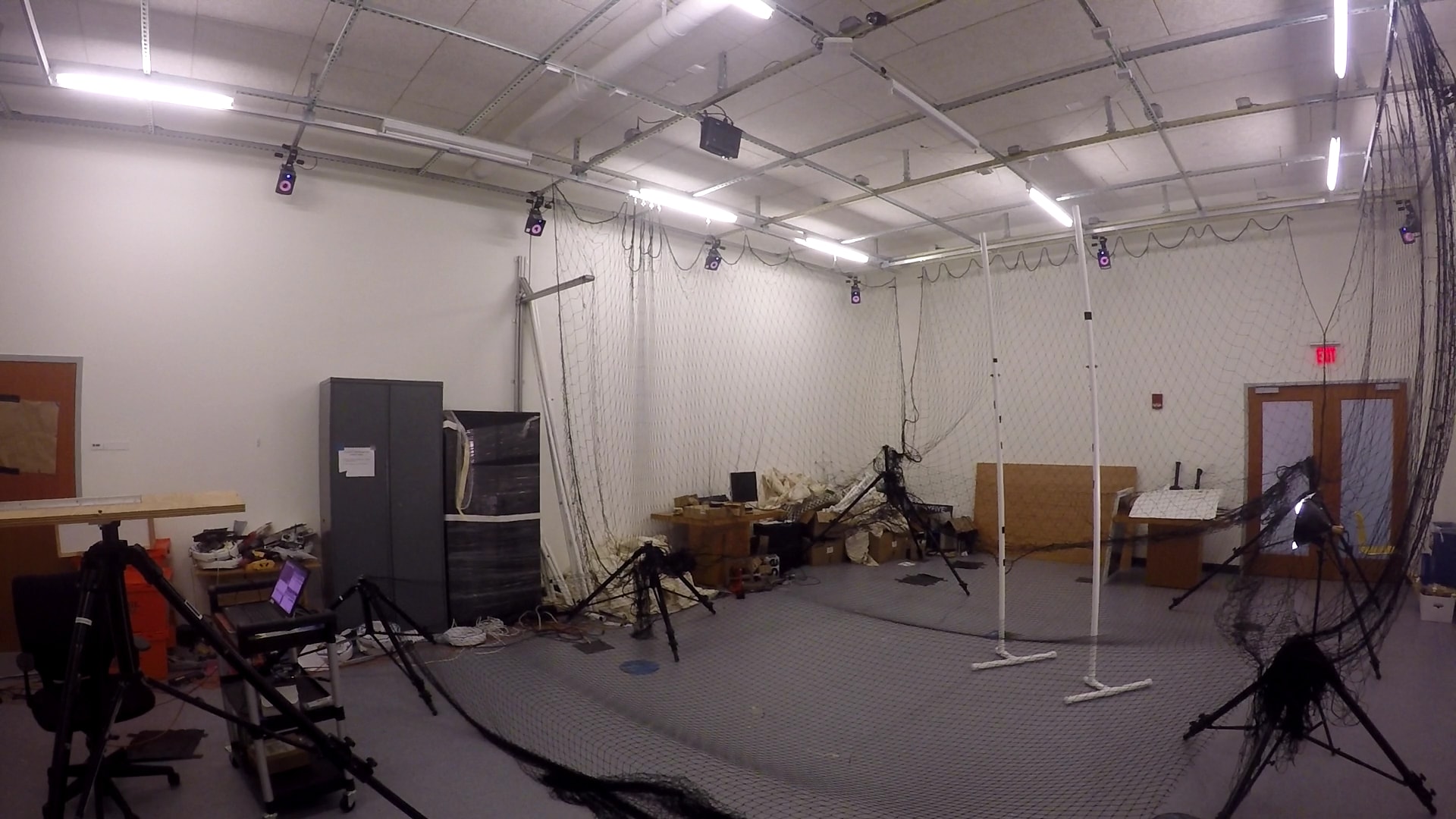}}    
\subfigure{\includegraphics[trim = 0mm 0mm 0mm 0mm, clip, width=0.49\columnwidth]{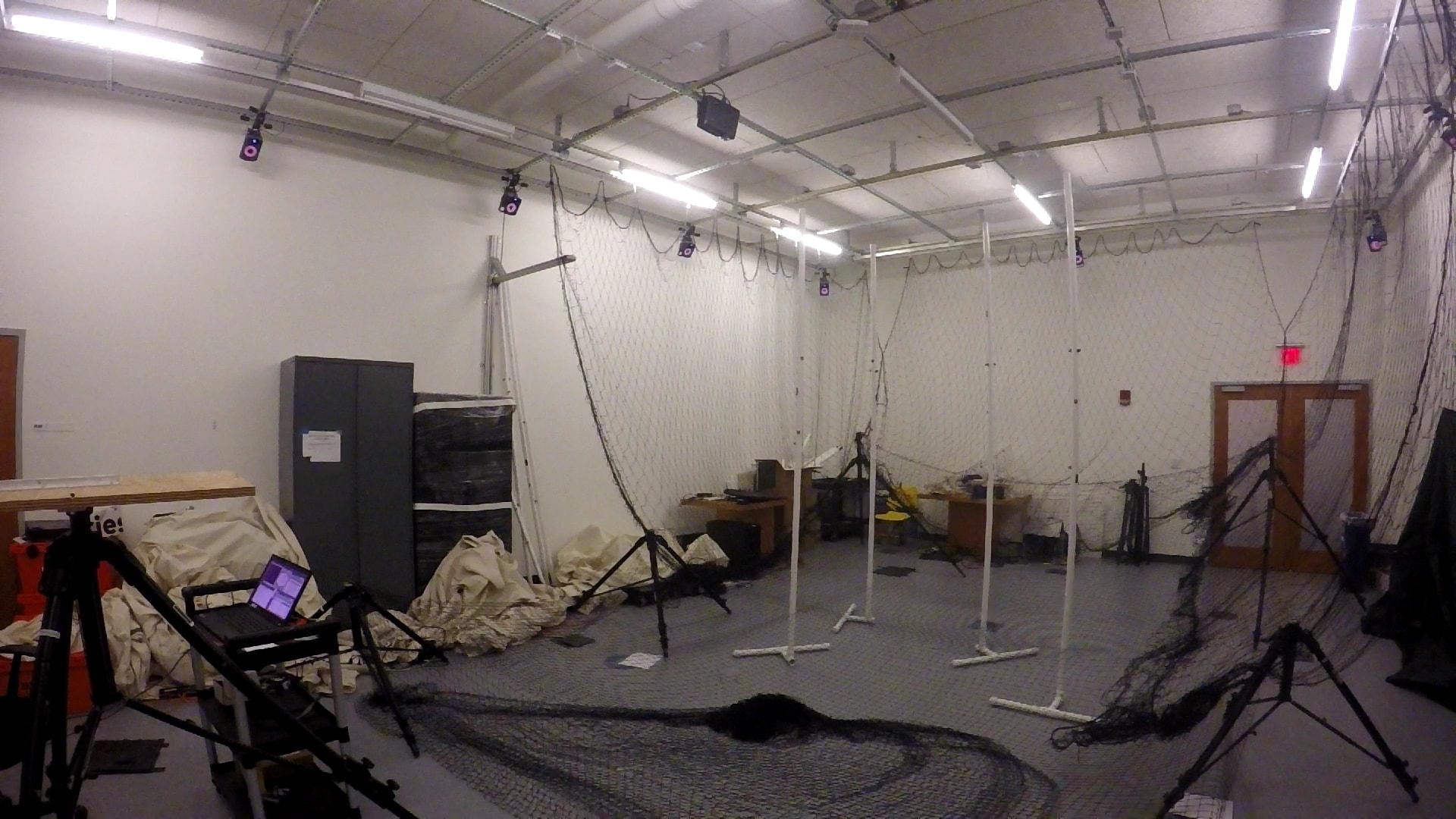}}   
\subfigure{\includegraphics[trim = 0mm 0mm 0mm 0mm, clip, width=0.49\columnwidth]{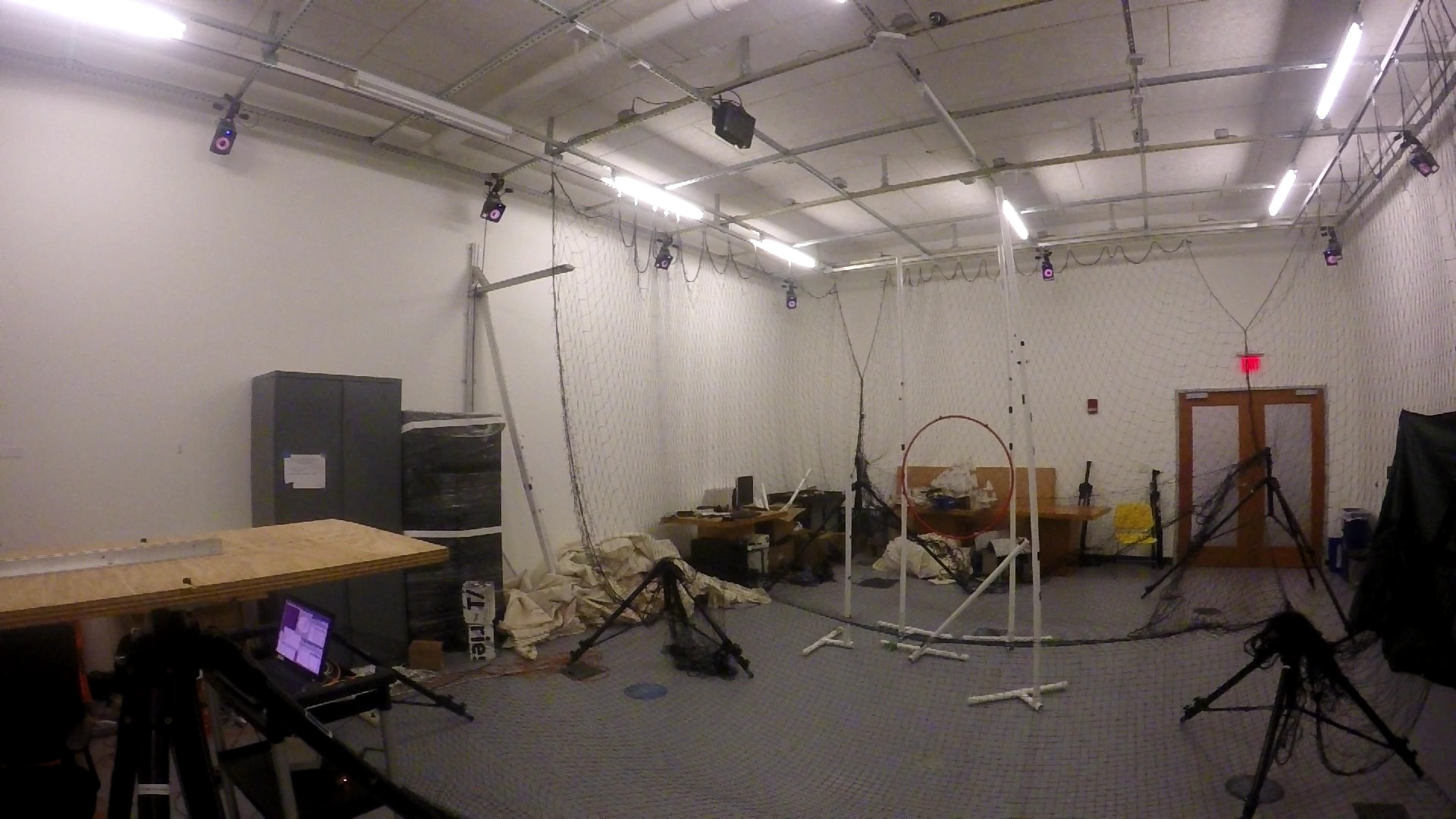}}    
\subfigure{\includegraphics[trim = 0mm 0mm 0mm 0mm, clip, width=0.49\columnwidth]{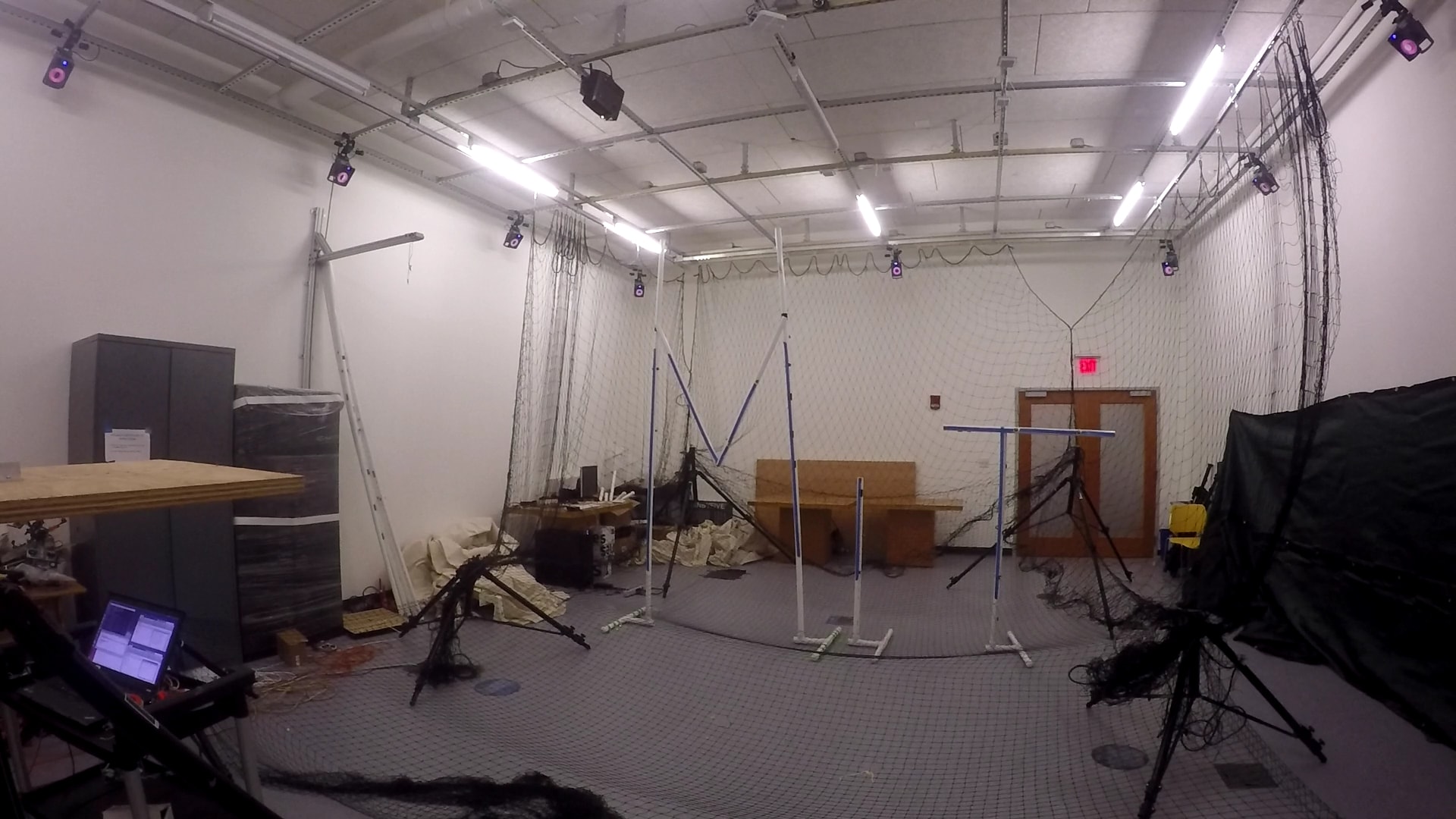}}      
\caption[Environments 1-8 on which the online planning algorithm was tested]{Environments 1-8 on which the online planning algorithm was tested. The obstacles include poles of different lengths in varying orientations and also a hoop of diameter equal to 0.9 m.  \label{fig:env_images_1}}
\end{figure}

\begin{figure}[h!]
\centering
\subfigure{\includegraphics[trim = 0mm 0mm 0mm 0mm, clip, width=0.49\columnwidth]{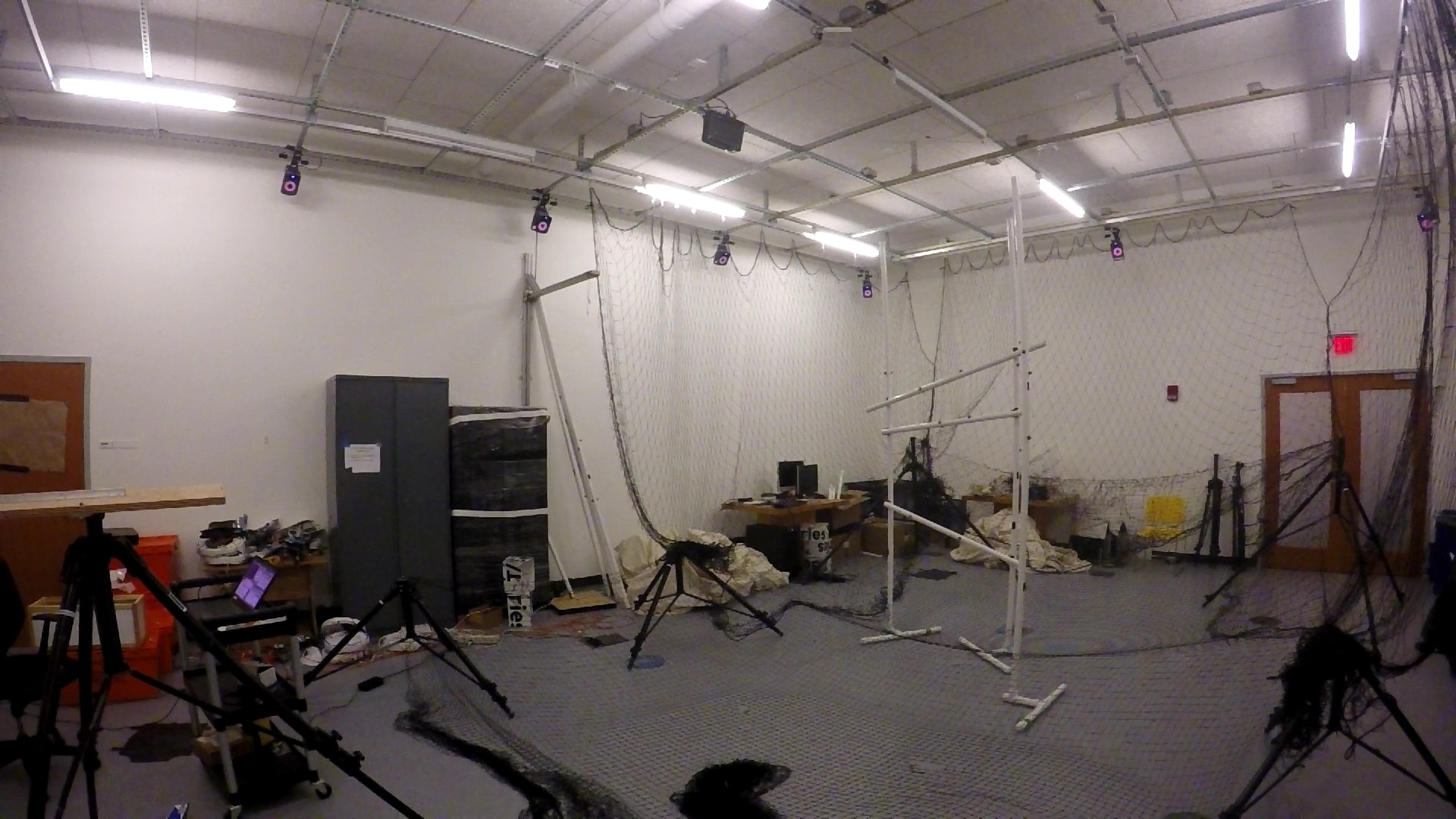}}    
\subfigure{\includegraphics[trim = 0mm 0mm 0mm 0mm, clip, width=0.49\columnwidth]{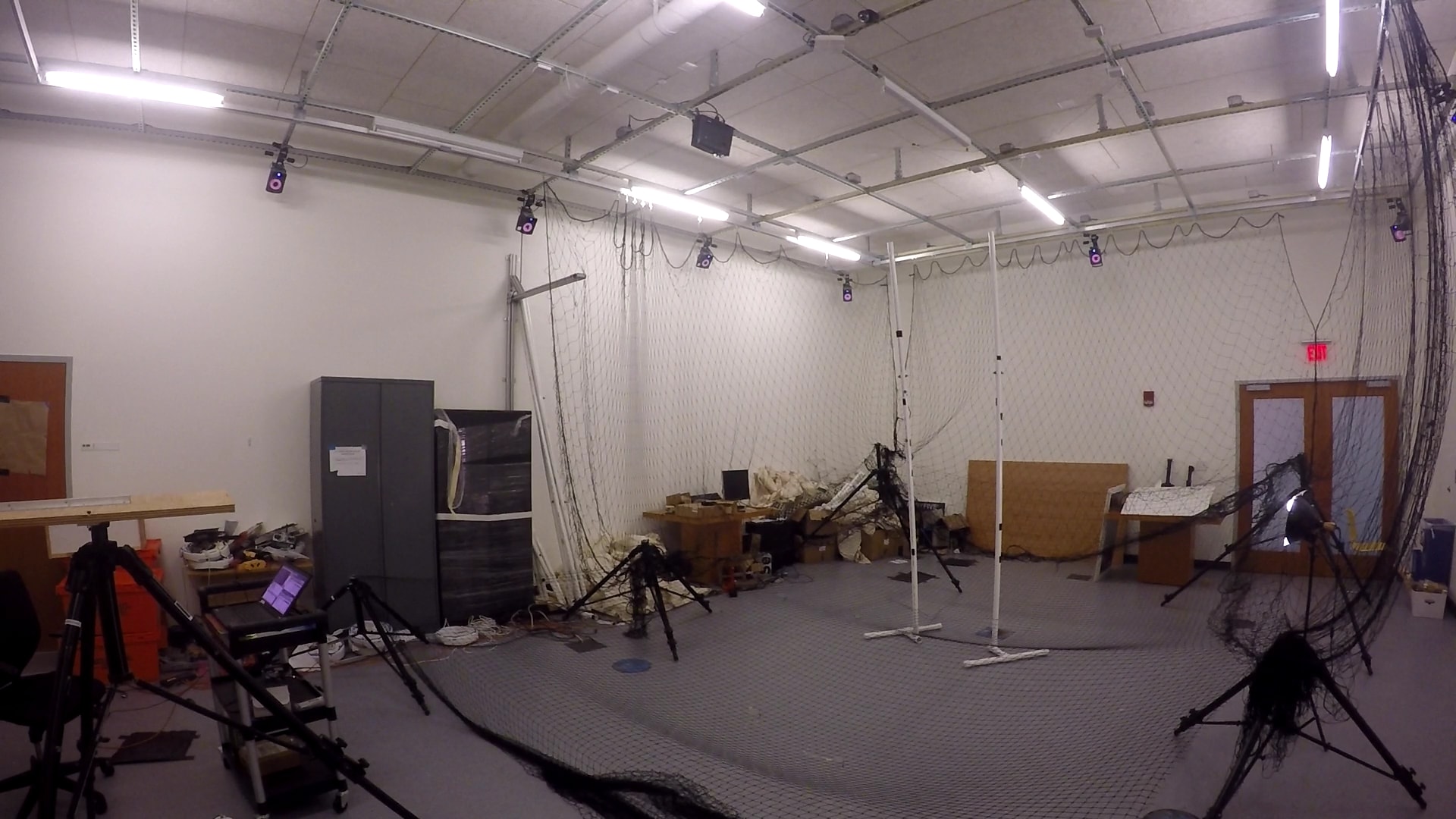}}    
\subfigure{\includegraphics[trim = 0mm 0mm 0mm 0mm, clip, width=0.49\columnwidth]{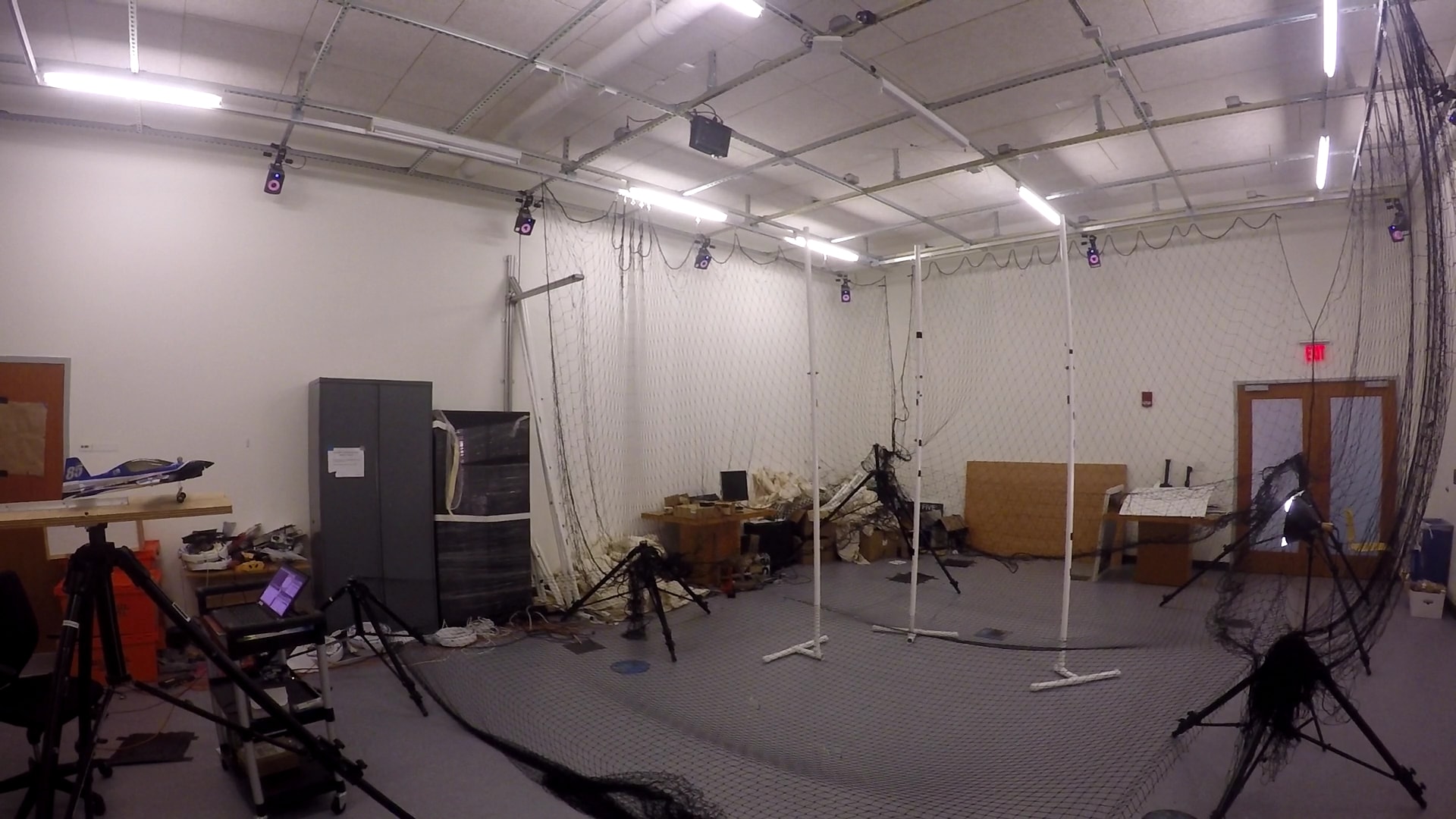}}    
\subfigure{\includegraphics[trim = 0mm 0mm 0mm 0mm, clip, width=0.49\columnwidth]{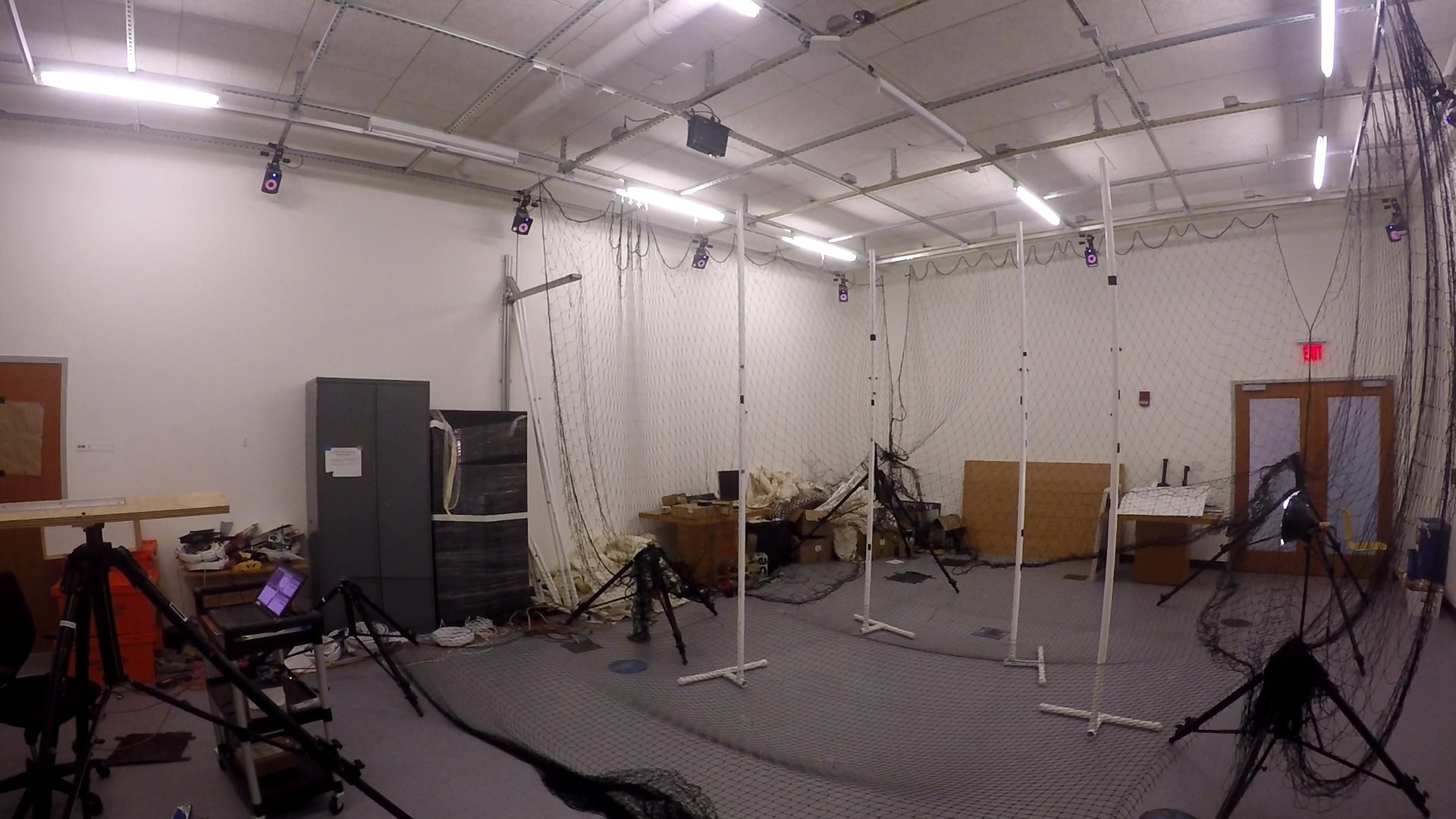}}    
\subfigure{\includegraphics[trim = 0mm 0mm 0mm 0mm, clip, width=0.49\columnwidth]{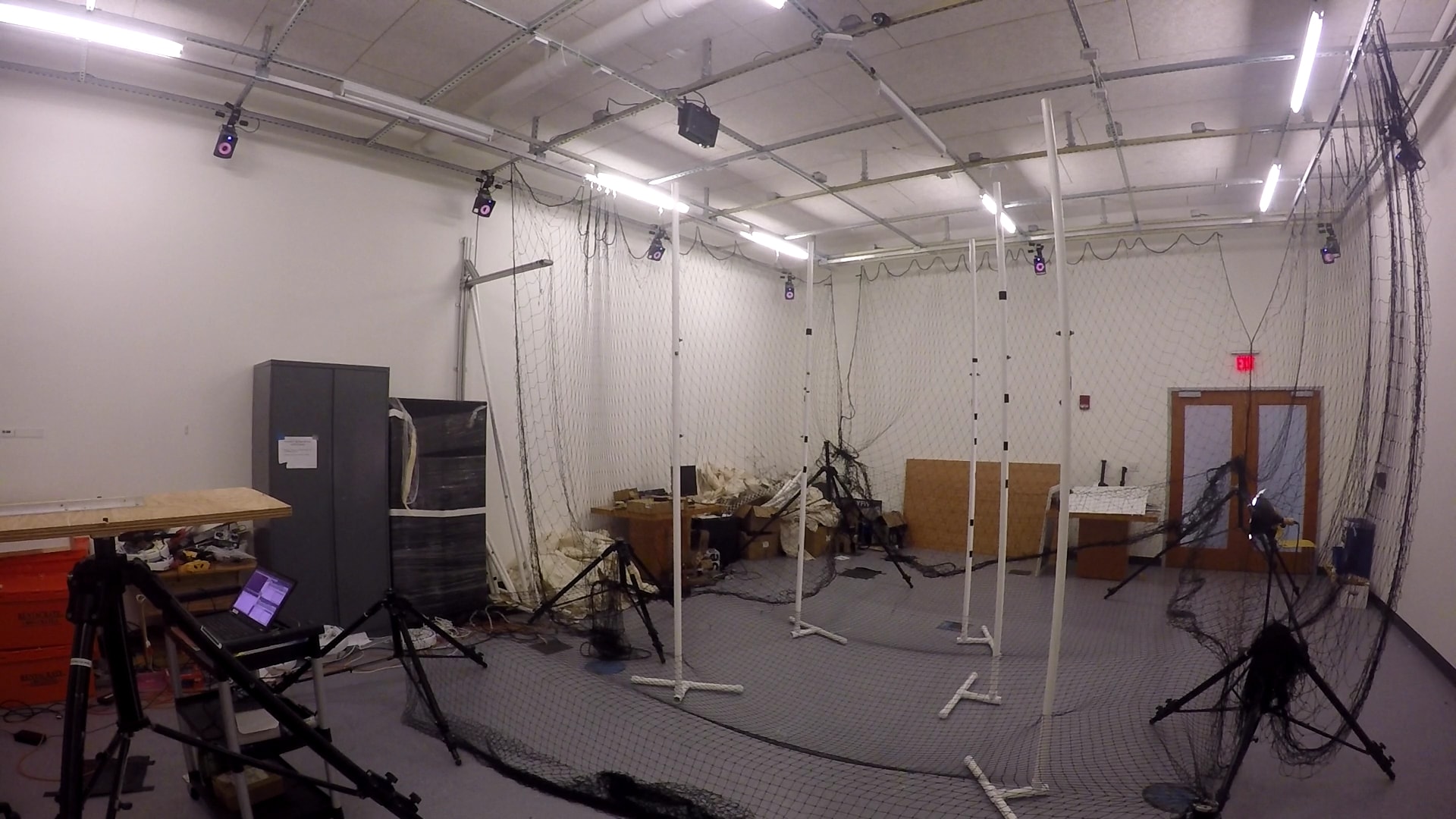}}    
   \subfigure{\includegraphics[trim = 0mm 0mm 0mm 0mm, clip, width=0.49\columnwidth]{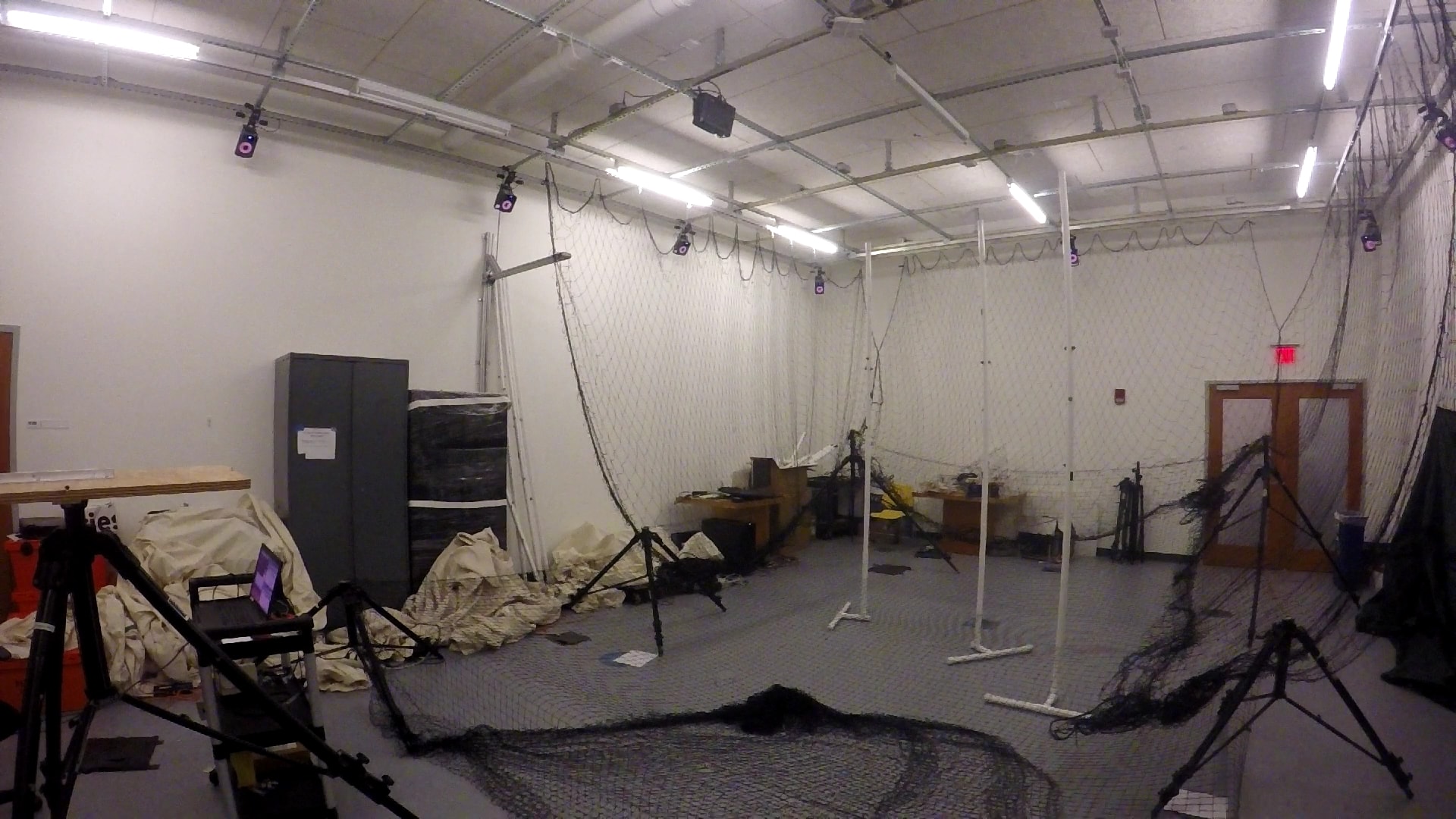}}
\subfigure{\includegraphics[trim = 0mm 0mm 0mm 0mm, clip, width=0.49\columnwidth]{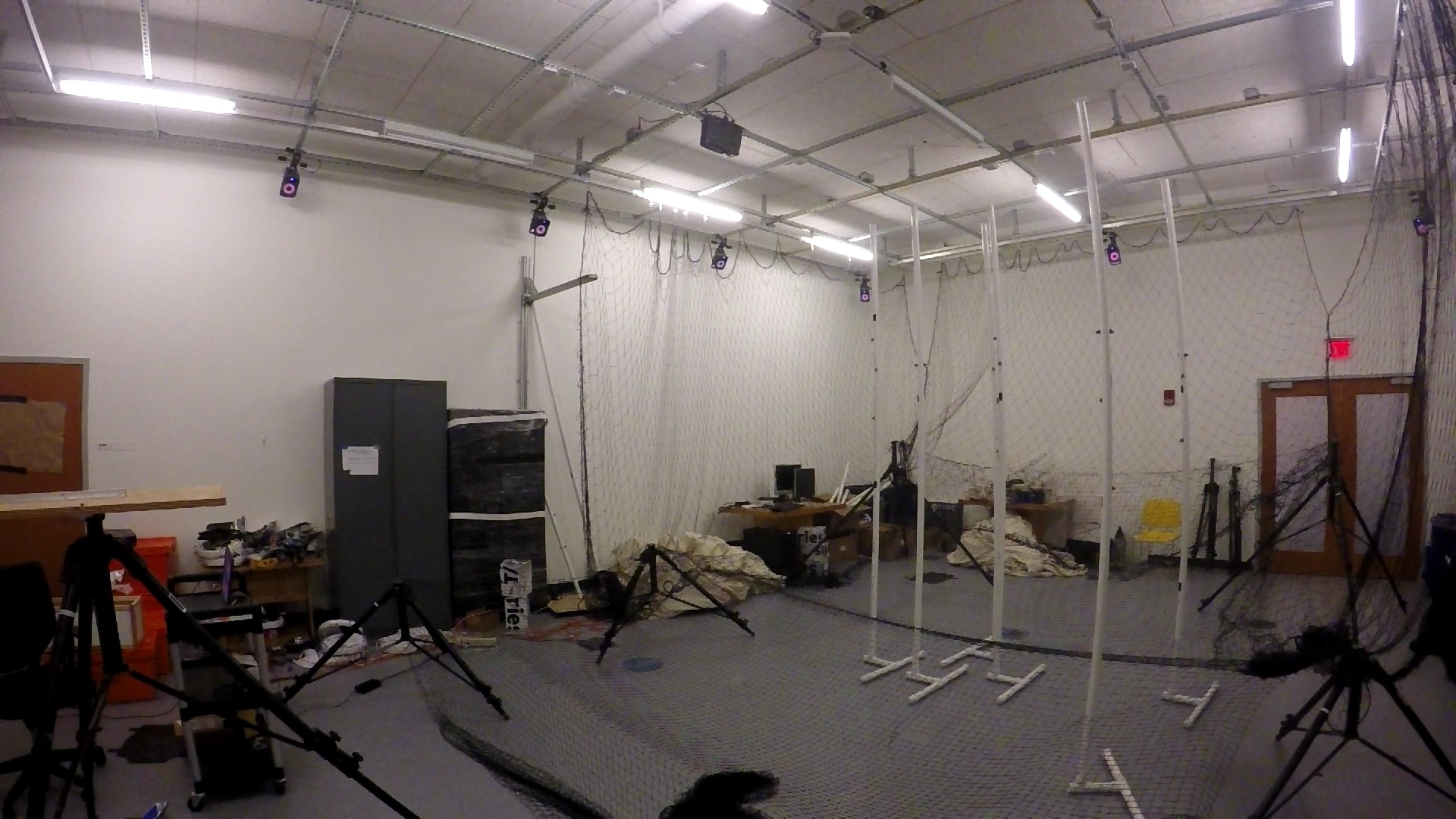}}  
\caption[Environments 9-15 on which the online planning algorithm was tested]{Environments 9-15 on which we tested our planning algorithm. The bottom most image shows the only failure case. Here the airplane brushed one of the obstacles on its way across the room. \label{fig:env_images_2}}
\end{figure}

\begin{figure}
\centering

\hfill
\subfigure{\includegraphics[trim = 0mm 15mm 0mm 15mm, clip, width=0.49\columnwidth]{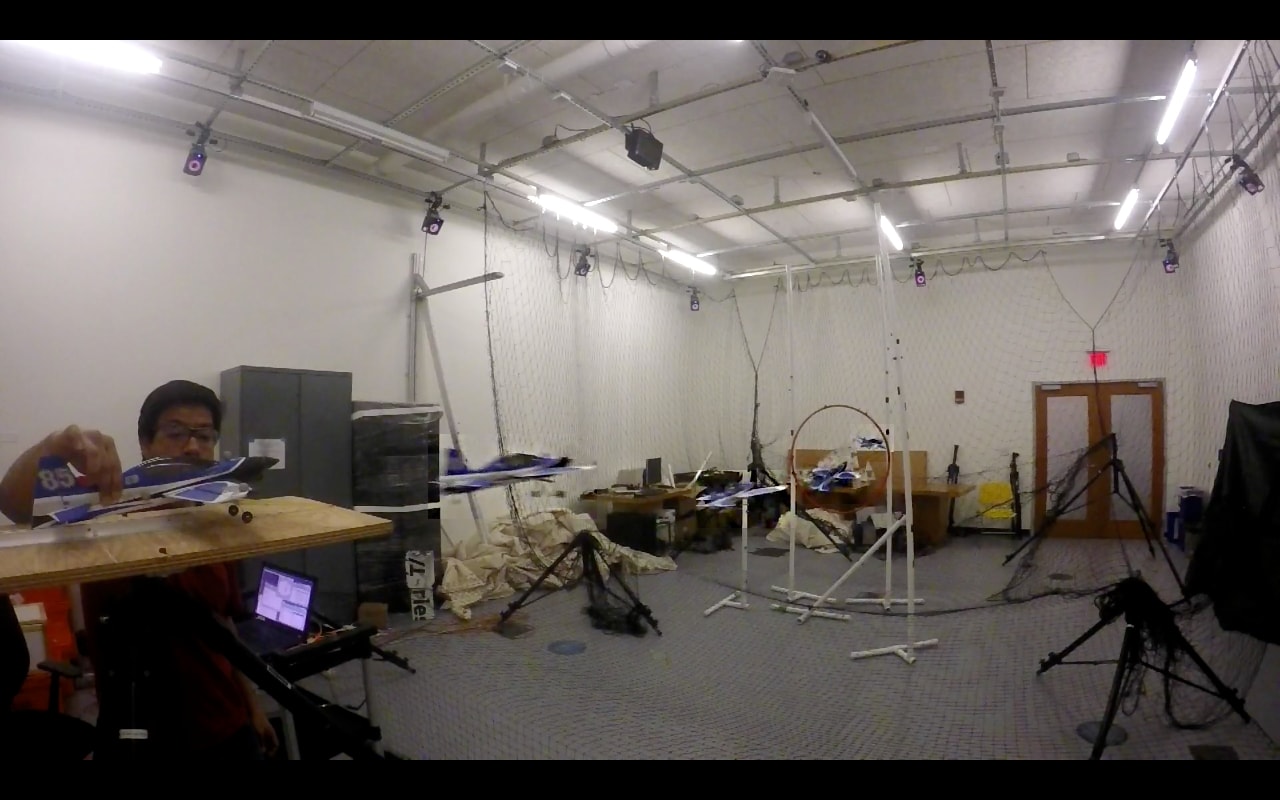}} 
\hfill
\subfigure{\includegraphics[trim = 0mm 0mm 0mm 0mm, clip, width=0.49\columnwidth]{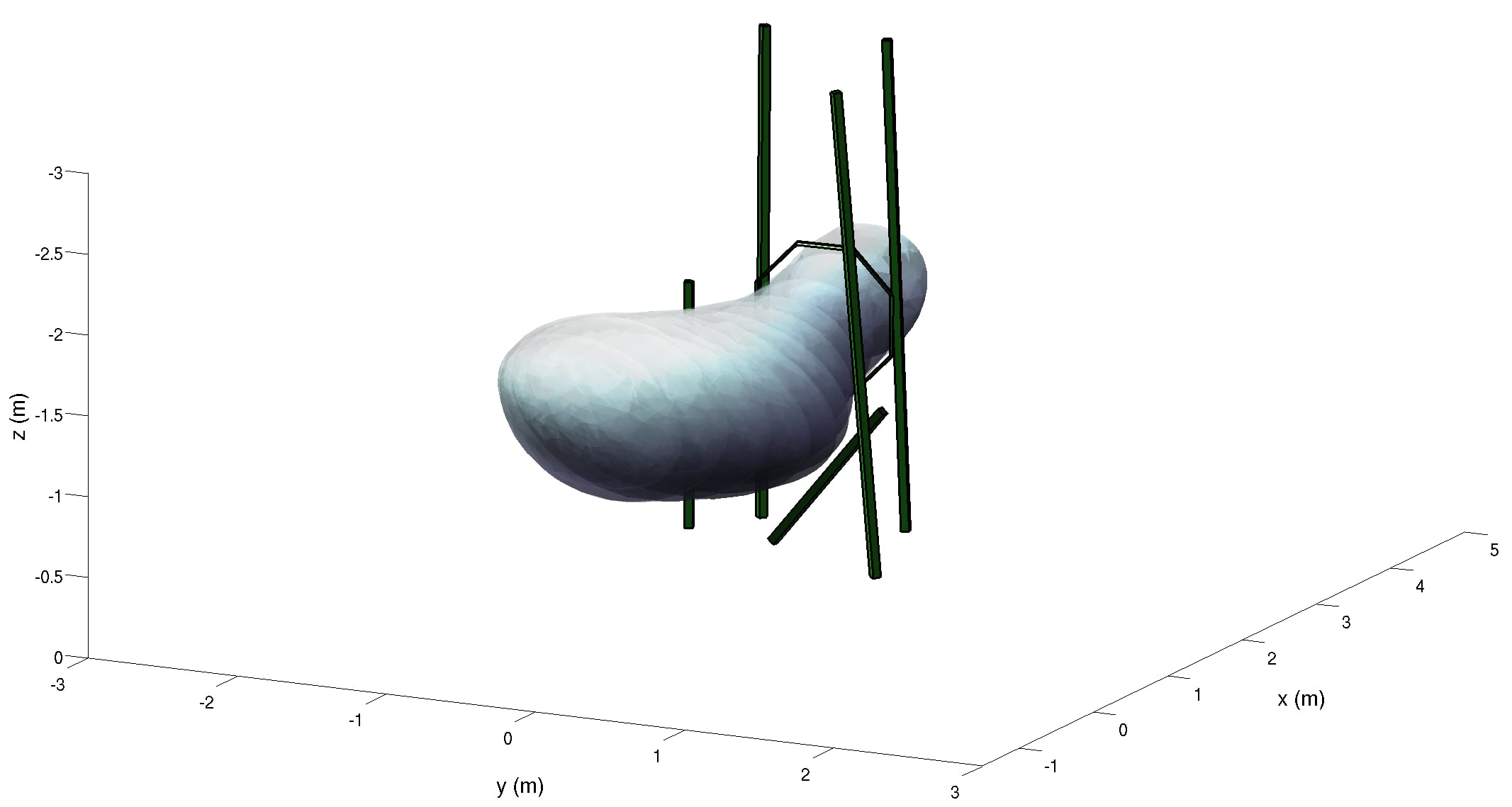}}    
\hfill\

\hfill
\subfigure{\includegraphics[trim = 0mm 15mm 0mm 15mm, clip, width=0.49\columnwidth]{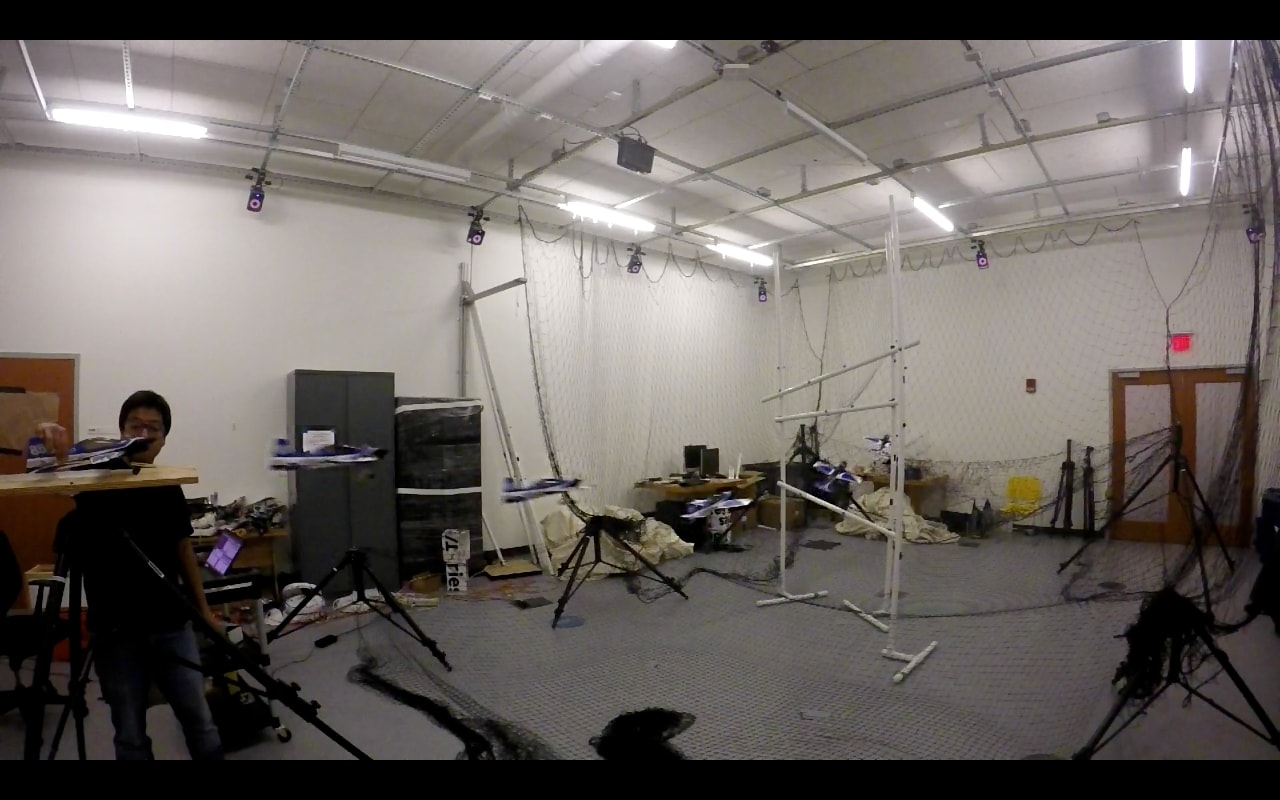}}  
\hfill  
\subfigure{\includegraphics[trim = 0mm 0mm 0mm 0mm, clip, width=0.49\columnwidth]{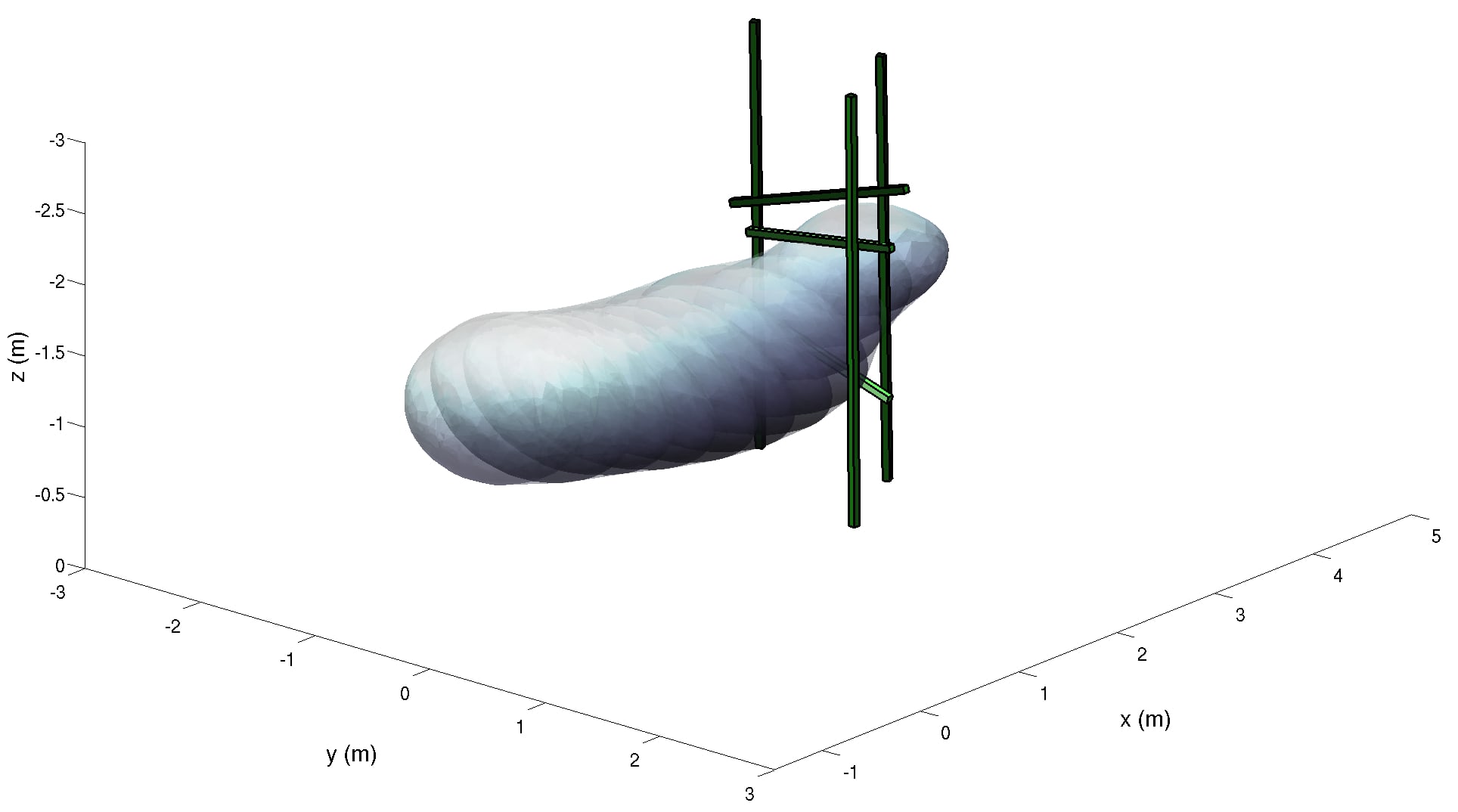}}    
\hfill\

\hfill
\subfigure{\includegraphics[trim = 0mm 15mm 0mm 15mm, clip, width=0.49\columnwidth]{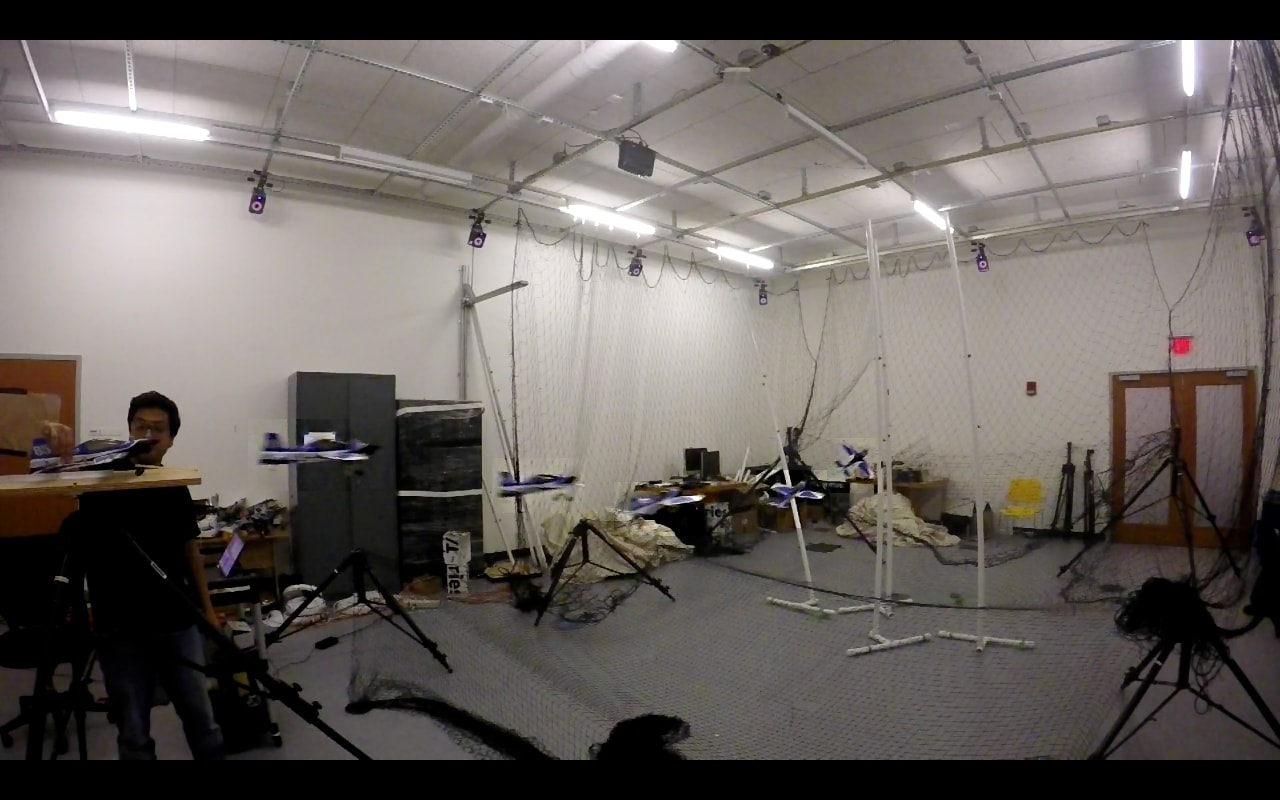}}  
\hfill
\subfigure{\includegraphics[trim = 0mm 0mm 0mm 0mm, clip, width=0.49\columnwidth]{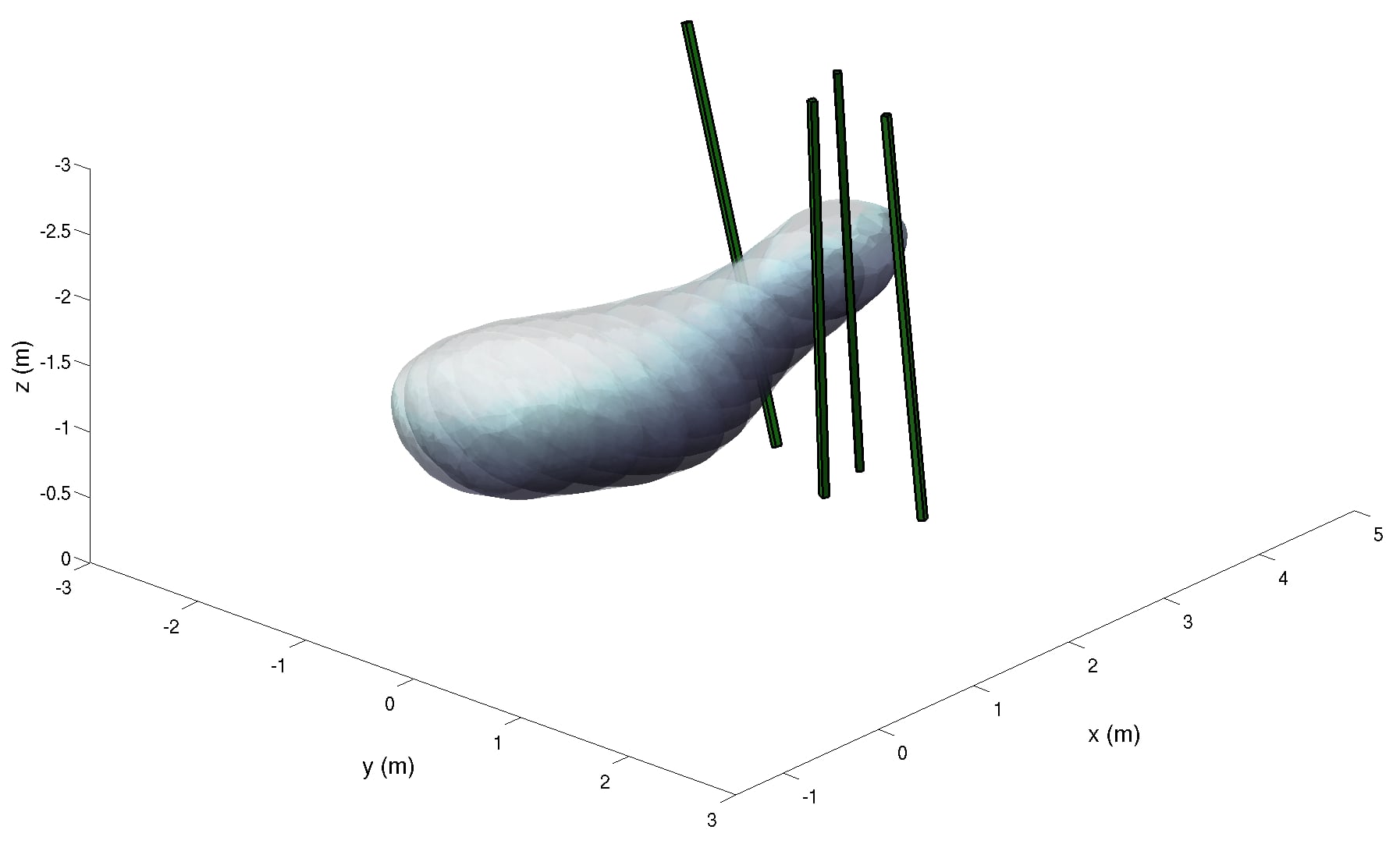}}   
\hfill\

\vfill
\subfigure{\includegraphics[trim = 0mm 15mm 0mm 15mm, clip, width=0.49\columnwidth]{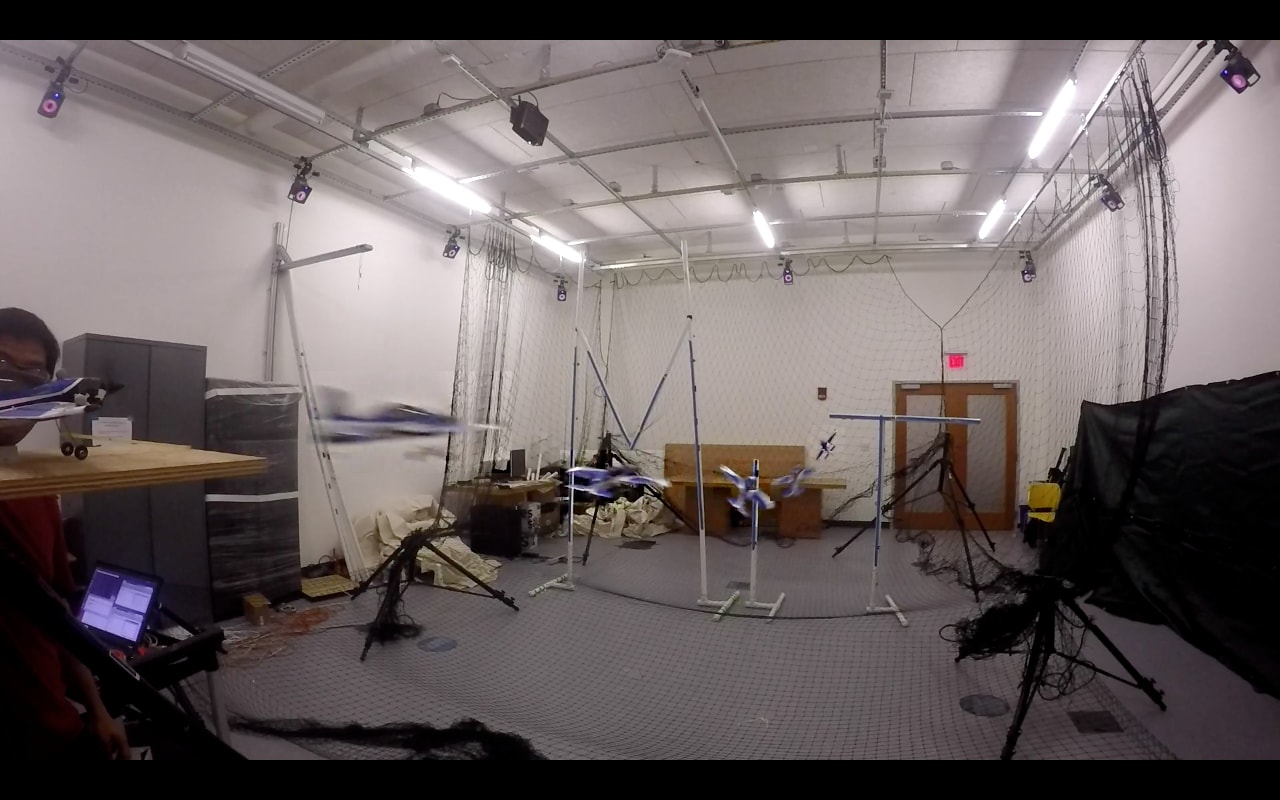}}   
\hfill
\subfigure{\includegraphics[trim = 0mm 0mm 0mm 0mm, clip, width=0.35\columnwidth]{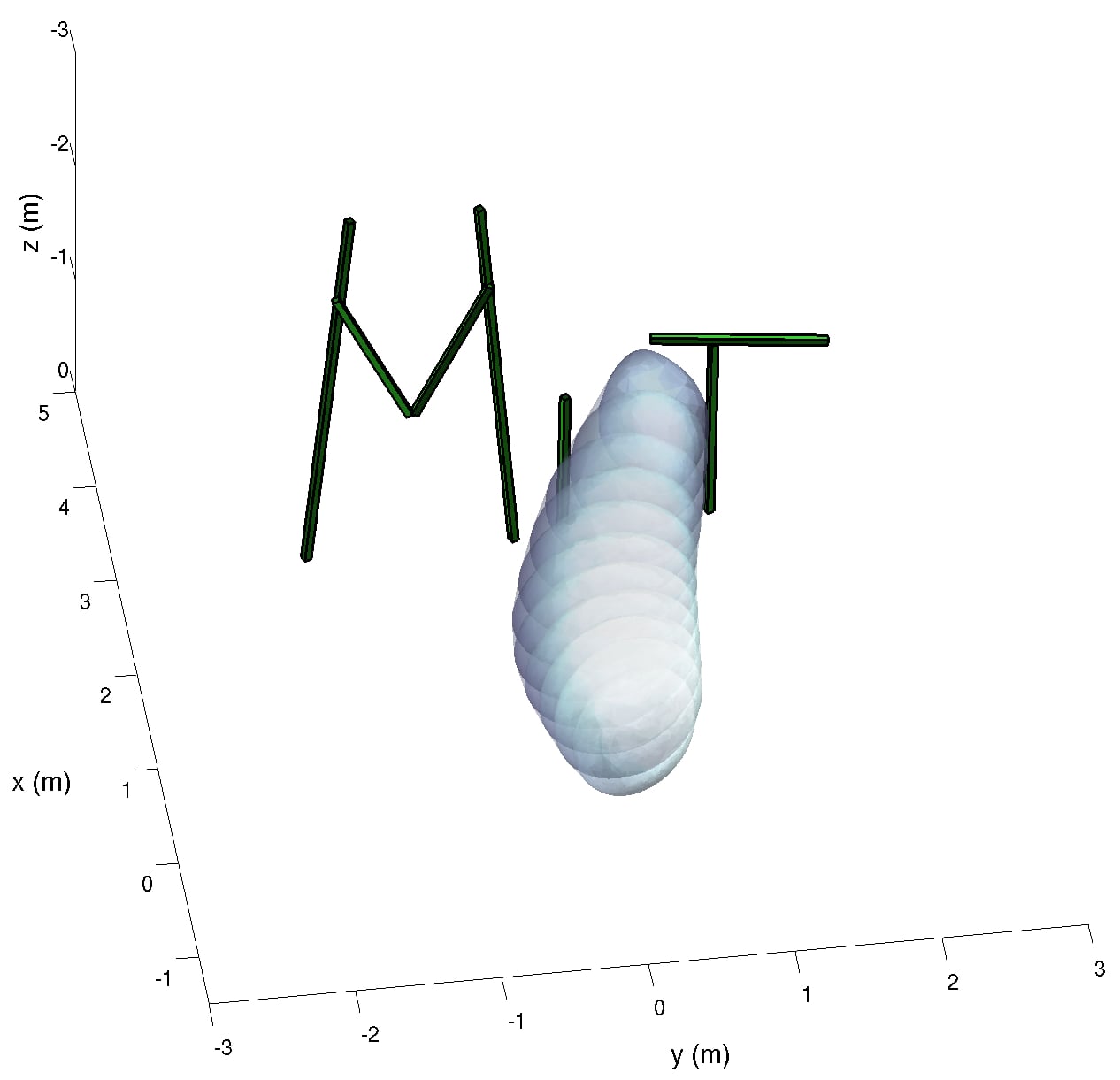}}    
\hfill\

\caption[Planned funnels for different environments traversed by airplane]{This figure depicts the planned funnel along with the polygonal obstacle representations for four of our fifteen test environments. Note that the funnels have been inflated to take into account the collision geometry of the airplane (modeled as a sphere of diameter equal to the wingspan).
\label{fig:sbach_planned_funnels}}
\end{figure}

We tested our approach on 15 different obstacle environments of varying difficulty. These environments are shown in Figures \ref{fig:env_images_1} and \ref{fig:env_images_2}. The obstacles include poles of different lengths in varying orientations and also a hoop of diameter equal to $0.9~m$ (as a reference the airplane's wingspan is $0.44~m$, thus only leaving $0.23~m$ of margin on either side of the airplane assuming it passes through the center of the hoop).  We model the poles as cuboids with heights equal to that of the poles and widths equal to the diameter. The hoop is approximated with eight polytopic segments (see Figure \ref{fig:sbach_planned_funnels}). 

\begin{figure*}[h!]
  \begin{center}
    \includegraphics[trim = 0mm 0mm 0mm 0mm, clip, width=0.7\textwidth]{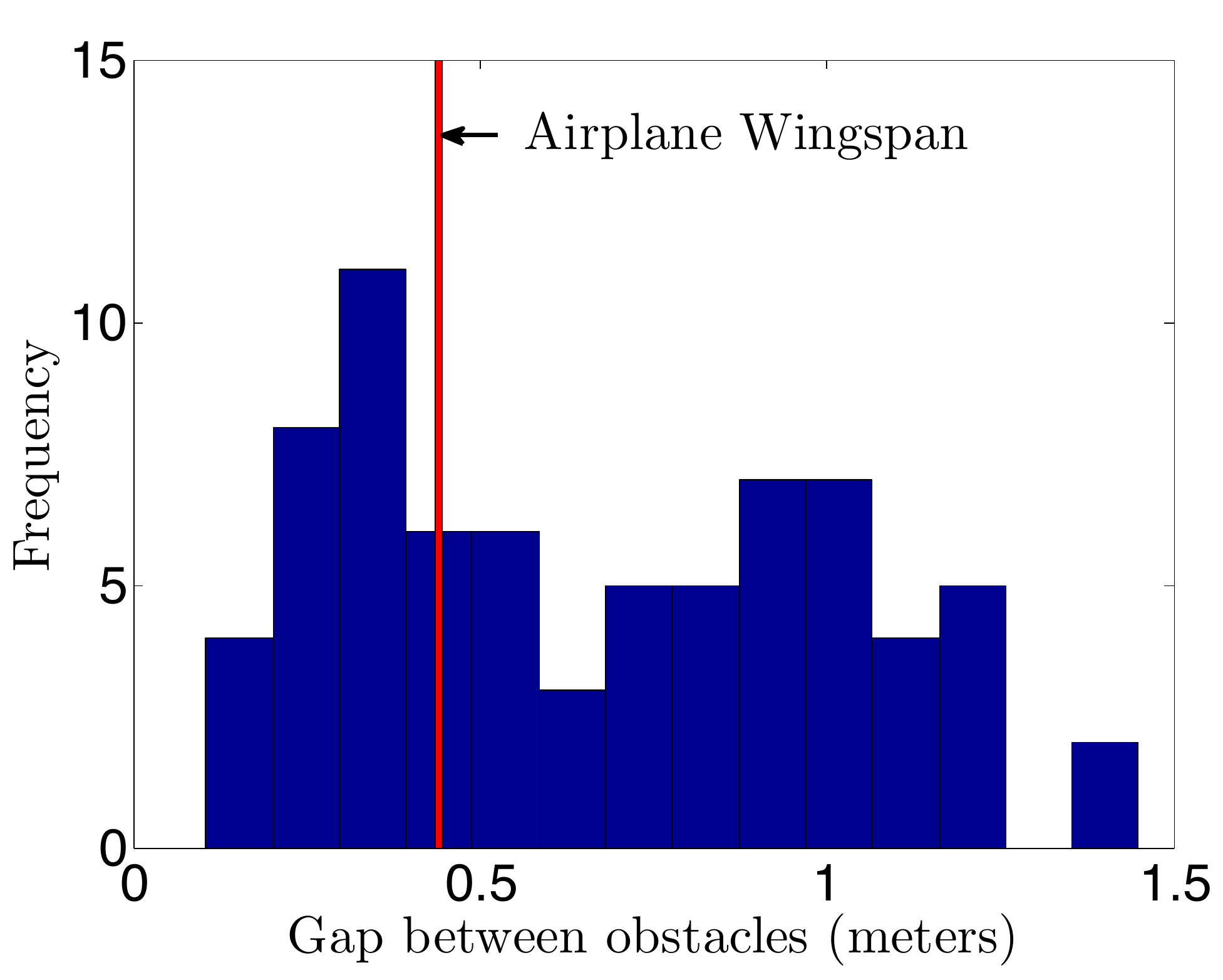}
  \end{center}
  \caption[Histogram of gaps between obstacles in our test environments]{Histogram of gaps between objects in our test environments compared with the airplane's wingspan ($0.44~m$). As the plot shows, a significant number of the gaps between obstacles are less than then wingspan.   \label{fig:histogram_object_distances}}
\end{figure*}

Figure \ref{fig:histogram_object_distances} presents a more quantitative perspective on the obstacle environments. Here we we have plotted a histogram of the gaps between obstacles in the environment and compared it to the wingspan of the airplane. In particular, for each obstacle in a given environment we consider the distance\footnote{Here, the distance between obstacles is defined in the usual way for sets. In particular, for any pair of obstacles $O_1$ and $O_2$, we define the distance between them as $\underset{x_1 \in O_1,x_2 \in O_2}{\textrm{min}} \ \|x_1 - x_2\|_2$.}  to the closest obstacle that is at least $5~cm$ away (the $5~cm$ threshold is chosen to prevent obstacles that are right next to each other being considered as having a small gap. For example, the obstacle closest to one of the horizontal poles in the first environment in Figure \ref{fig:env_images_1} should be the other horizontal pole and not the adjacent vertical pole). As the histogram illustrates, a significant fraction (approximately $35 \%$) of the gaps are less than the airplane wingspan and about $66\%$ of the gaps are less than two wingspans. Of course, it is worth noting that not all of the gaps greater than the wingspan are in fact negotiable (e.g., obstacles placed far apart along the $x$ direction).

A video of the airplane traversing a few representative environments is available online at \href{https://youtu.be/cESFpLgSb50}{https://youtu.be/cESFpLgSb50}. \revision{For our experiments, we flew the airplane in each of the 15 environments once\footnote{The only exception to this was a small number of occasions during our experiments where we observed the ``failsafe" option being employed by the planner (i.e., switching off the throttle and gliding to a halt). This was typically caused by failures in the launching mechanism such as the airplane making contact with the operator's hand on its way out of the launcher. Due to the altered initial conditions in these cases the airplane is not in the inlet of most or all of the funnels and thus has to resort to the failsafe. On these occasions we simply repeated the experiment to obtain a successful launch.}. Out of the 15 environments the airplane was able to successfully negotiate 14 of them, thus demonstrating the efficacy of our real-time planner on this challenging task.  We further performed five more flights on a single environment (the one in the bottom right corner of Figure \ref{fig:env_images_1}). The repeated flights on this environment were all successful, thus demonstrating the robustness of our approach to initial conditions.}

\begin{figure}
\centering

\subfigure{\includegraphics[trim = 135mm 40mm 130mm 90mm, clip, width=0.49\columnwidth]{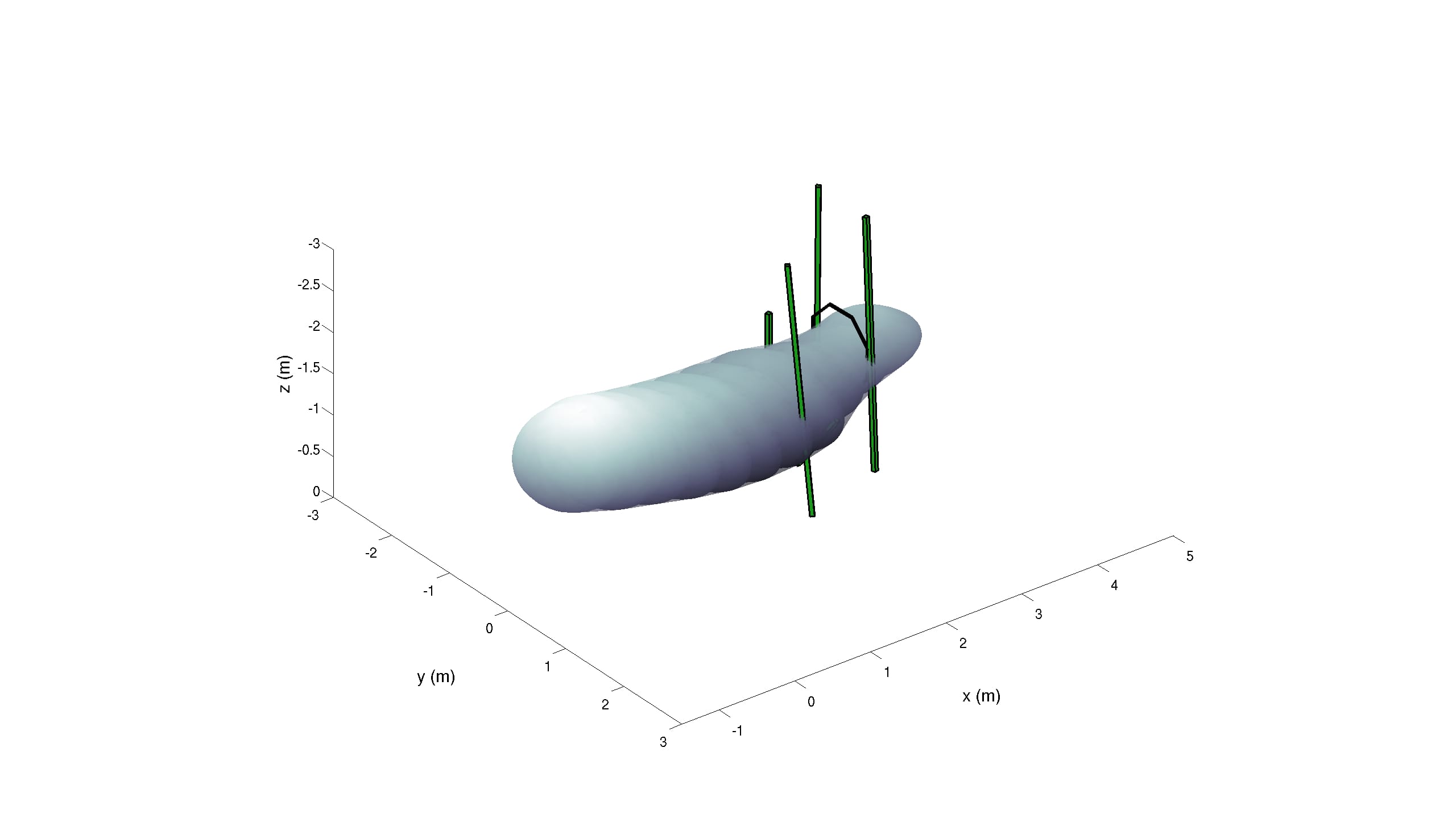}} 
\subfigure{\includegraphics[trim = 135mm 40mm 130mm 90mm, clip, width=0.49\columnwidth]{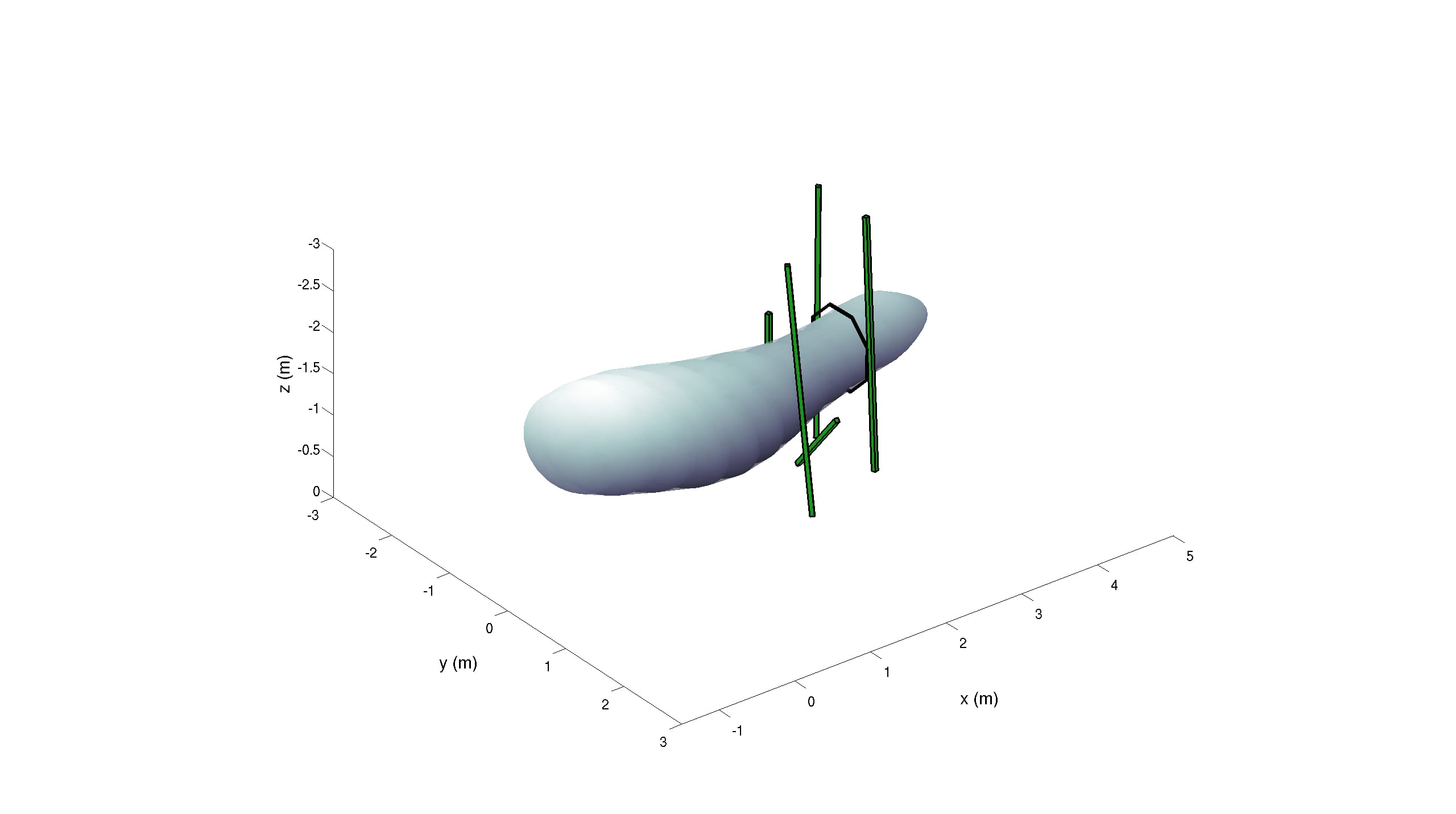}}    

\caption[The importance of shifting funnels by exploiting invariances in dynamics]{The QCQP-based algorithm for shifting funnels (Section \ref{sec:shifting funnels}) plays a crucial role in the planner's ability to find collision free funnels. Here we compare the output of the planner with and without applying the QCQP-based algorithm for shifting funnels. The best funnel found in the former case is well in collision with the obstacles while in the latter case the planner is able to find a collision free funnel. Note that the funnels chosen in the different cases correspond to different maneuvers.
\label{fig:sbach_hoop_shifting}}
\end{figure}

Figure \ref{fig:sbach_planned_funnels} presents the output of our real-time planner on four of the more challenging environments. In particular we plot the funnel chosen by the planner (which have been shifted using the QCQP-based algorithm presented in Section \ref{sec:shifting funnels}) alongside an image sequence showing the airplane executing the plan. We note that the increased expressivity afforded to us by the ability to shift funnels in the cyclic coordinates has a large impact on the planner being able to find collision free funnels. Perhaps the best example of this is the environment which contains the hoop (top row of Figure \ref{fig:sbach_planned_funnels}).  Without the ability to make small adjustments to the funnels, the chances of finding a collision free funnel are extremely low (we would require a funnel that passes almost exactly through the center of the hoop). Figure \ref{fig:sbach_hoop_shifting} compares the output of the planner with and without applying the QCQP-based algorithm for shifting funnels. As we can observe, the best funnel found in the former case is well in collision with the obstacles while in the latter case the planner is able to find a collision free funnel. This illustrates the crucial role that exploiting invariances plays in the success of the planner on this task.

As mentioned before, the airplane was able to successfully negotiate 14 out of our 15 environments. The single failure case occurred on the environment shown at the bottom of Figure \ref{fig:env_images_2}, where the airplane clipped one of the poles close to the end of its flight. 
This failure can be attributed to the fact that the planner chose to execute one of a small handful of maneuvers that are more aggressive than the funnel we validated in Section \ref{sec:sbach_funnel}. The controller saturates the control inputs significantly on this maneuver and hence violates our assumption about input limits not being reached. Hence this funnel is not entirely valid on the hardware system. While we could have taken actuator saturations into account while computing the funnels (see Section \ref{sec:funnels_saturations}), we chose not to do so in order to reduce the computation time. In hindsight, a more careful treatment of the system would have included saturations when computing funnels.

\subsection{Implementation details}

We end this section by mentioning a few implementation details that were important in achieving the results presented above. First, while the Vicon motion tracking system provides accurate position and orientation estimates at $120$ Hz, a finite difference of these measurements can lead to noisy estimates of the derivative states. For our experiments we filtered the finite differences using a simple Luenberger observer \cite{Luenberger71}. 

Second, we observed a delay of approximately $55~ms$ in our closed-loop hardware system. While there are several ways to explicitly take this into account by adding delay states to our airplane model, we accommodate for the delay during execution by simulating our model forwards by $55~ms$ from the estimated current state and using this simulated future state to compute the current control input. This simple strategy is a common one and has previously been found to be effective in a wide range of applications \cite{Horiuchi99, Moore14a, Sipahi12}.

%% file: conclusion.tex
\section{Discussion and Conclusion}
\label{sec:conclusion}

In this paper we have presented an approach for real-time motion planning in a priori unknown environments with dynamic uncertainty in the form of bounded parametric model uncertainty and external disturbances.  The method augments the traditional
trajectory library approach by constructing stabilizing controllers around the
nominal trajectories in a library and computing outer approximations of reachable sets
(ÒfunnelsÓ) for the resulting closed-loop controllers via sums-of-squares (SOS)
programming. The pre-computed funnel library is then used to plan online by sequentially
composing them together in a manner that ensures obstacles are avoided.

We have demonstrated our approach using extensive simulation experiments on a ground vehicle model. These experiments demonstrate that our approach can afford significant advantages over a trajectory-based approach. We also applied our approach to a quadrotor model and demonstrated how for certain classes of environments we can guarantee that the system will fly forever in a collision-free manner. We have also validated our approach using thorough hardware experiments on a small fixed-wing airplane flying through previously unseen cluttered environments at high speeds. To our knowledge, \revision{these simulation and hardware results are among the first examples} of provably safe and robust control for robotic systems with complex nonlinear dynamics that need to plan in real-time in environments with complex geometric constraints. It is also worth noting that while one often associates robustness with very conservative behavior, our hardware experiments demonstrate that this need not be the case. In particular, the airplane performs some very aggressive maneuvers while still being robust.

\subsection{Challenges and extensions}
\label{sec:challenges}

\subsubsection{Numerical difficulties}

There are a number of challenges associated with our approach. One of the main difficulties in implementing our method is the relative immaturity of solvers for semidefinite programs (SDPs). While recently released software such as the MOSEK solver \cite{Mosek10} have improved the speed with which solutions can be obtained, SDP solvers are still in their infancy as compared to solvers for Linear Programs (LPs). Thus, numerical issues (e.g., due to the scaling of problem data) inevitably arise in practice and must be dealt with in a relatively ad hoc manner (e.g., rescaling problem data or removing redundant decision variables). However, preprocessing of sums-of-squares (SOS) programs and SDPs is an active area of inquiry \cite{Permenter14,Lofberg09,Seiler13}  and solver technology is bound to improve as SDPs are more widely adopted in practice.

\subsubsection{Robots with complex geometries}

Another challenge associated with our method has to do with implementing the approach on robots with complex kinematics/geometry. One option (as described in Section \ref{sec:planning}) is to project the funnel onto the configuration space (C-space) of the system and perform collision checking against the C-space representation of obstacles. However, computing C-space representations of obstacles is typically challenging for non-trivial geometries, especially since it needs to be done as obstacles are reported in real-time. 

For the examples considered in this paper, the geometry of the robots were approximated relatively accurately by spheres. This allowed us to project the funnel onto the $x-y-z$ space (in contrast to the full configuration space of the system) and inflate this projection by the radius of the corresponding sphere. These inflated funnels were then used for collision-checking during real-time planning. For robots with complex geometries, inflating the funnel in $x-y-z$ space in this manner is not an entirely straightforward operation. However, this is potentially a more promising approach than performing collision-checking against C-space obstacles since the inflation can computed in the offline computation stage as it depends only on the geometry of the robot and not the obstacles.

\subsubsection{Designing funnel libraries}
\label{sec:designing libraries}

There is an inherent tradeoff in our approach between the richness of the funnel library and the amount of computation that needs to be performed in real-time in order to be able to search through it. For extremely large funnel libraries, it may be computationally difficult to search through all the funnels while planning online. It is thus important to explore ways in which one could speed up this search. For example, one could exploit the observation that funnels in close proximity to a funnel that is in collision are also likely to be in collision and use this to choose the sequence in which to evaluate funnels for collisions. Further, given certain features of the environment one may be able to predict which funnels are likely to be in collision (without actually performing collision-checking) and evaluate these funnels first. These intuitions have been formalized and exploited using the theory of submodular optimization in the context of trajectory libraries \cite{Dey12a, Dey15a}. The approach allows one to optimize the sequence in which trajectories are evaluated and should be generalizable to funnel libraries as well.

A closely related question is how to choose the funnels in the library in the first place. As we observed in Section \ref{sec:quadrotor} for the quadrotor system, one can derive relatively simple geometric conditions on the environment in order for us to be able to guarantee that the system will be able to navigate through it without collisions. If we know a priori that our environment will satisfy these geometric conditions, this provides a way to check if our funnel library is sufficiently rich. However, for real-world environments (e.g., forests) we may not be able to make such assumptions. Instead, we might have a \emph{generative} (probabilistic) model of our environments and could use this to evaluate/design our funnel library. For instance, it is known that the locations of
trees in a forest are modeled well by spatial Poisson processes \cite{Stoyan00}. In such settings, it may be possible to design a randomized
algorithm for generating the funnel library where one samples different realizations of the environment, searches through the existing library to find a collision-free funnel for the sampled environment, and adds a new funnel to the library when such a funnel doesn't exist.

\revision{In a similar vein, extending the results presented in \cite{Karaman12} on flight through ergodic forests to the funnel library setting is an interesting direction to pursue. In particular, \cite{Karaman12} uses results from percolation theory to provide bounds on the speed at which a robot can fly safely through a forest in which obstacles have been sampled from an ergodic spatial process. Extending the results from \cite{Karaman12} to reason about funnel libraries could allow us to check if a given funnel library is sufficiently rich under the assumption that the spatial process from which obstacles are drawn is ergodic and known.}

\subsubsection{Probabilistic guarantees}

Throughout this paper, we have assumed that all disturbances and uncertainty are
bounded with probability one. In practice, this assumption may either not be fully
valid or could lead to overly conservative performance. In such situations, it is more
natural to provide guarantees of a probabilistic nature.
Recently, results from classical super-martingale theory have been combined with
sums-of-squares programming in order to compute such probabilistic certificates
of finite time invariance \cite{Steinhardt11a}, i.e. provide upper bounds on the probability that a
stochastic nonlinear system will leave a given region of state space. Combining the techniques presented in \cite{Steinhardt11a} with the approach presented in this work
to perform robust online planning on stochastic systems will be the subject of
future work.

\subsubsection{Reasoning about perception systems}

In this paper, we have focused most of our attention on reasoning about the real-time planning and control systems. In particular, we assumed that the robot is equipped with a perception system that reports (at runtime) regions in which obstacles are guaranteed to lie within. This may not always be a valid assumption; in practice, perceptual systems can often have false negatives and fail to report an obstacle in the environment. In such scenarios, it is unreasonable to expect to guarantee with probability one that the system will remain collision-free and we can only hope for probabilistic guarantees. One could envision modeling this perceptual uncertainty by considering an occupancy map with probabilities of occupancy associated with each voxel. Modeling and reasoning about such perceptual uncertainty remains an important open problem.

\subsubsection{Real-time computation of funnels}
\label{sec:dsos}

Due to the computation time associated with funnels, the approach presented in this work had two phases: an offline phase for computing funnels and an online stage for real-time planning with funnels. Recently, more scalable alternatives to SOS programming have been introduced \cite{Ahmadi14,Ahmadi14a}.  These alternatives rely on \emph{linear} and \emph{second-order cone programming} instead of semidefinite programming. This makes it possible to obtain large computational gains in terms of scalability and running time using DSOS and SDSOS programming as compared to SOS programming (see \cite{Majumdar14a} for comparisons of DSOS/SDSOS with SOS on large-scale controls applications). In addition to the ability to handle higher dimensional robotic systems, this also raises the interesting possibility of being able to dispense with the offline computation stage and compute a funnel and feedback controller \emph{in real-time} based on the geometry of the environment. Currently, this would only be feasible for relatively low-dimensional systems but may be an exciting direction to pursue.

\quad

\noindent We believe that the work presented in this paper has the potential to be deployed on real robots to make them operate safely in real-world environments. Our hope is that by building upon this work and pursuing the directions for future research presented above we can make this a reality.

%% file: acknowledgements.tex
\section*{Acknowledgements}
The authors are grateful to Tomas Lozano-Perez, Emilio Frazzoli, Claire Tomlin, and the members of the Robot Locomotion Group at MIT for a number of helpful discussions on this work. 
The authors would also like to thank Michael Posa for help formulating the QCQP for shifting funnels presented in Section \ref{sec:shifting funnels}; Adam Bry and Charlie
Richter for help setting up the USB RC transmitter used in the experiments described in Section \ref{sec:hardware}; and Andy Barry for advice on choosing the fixed-wing airplane used in Section \ref{sec:hardware} and his help with the launching mechanism for the airplane.

\section*{Funding}
This work was supported by the Office of Naval Research [MURI grant number N00014-10-1-095].